\documentclass{article}

\PassOptionsToPackage{square,numbers}{natbib}
\usepackage[preprint]{neurips_2026} 
\usepackage[utf8]{inputenc} 
\usepackage[T1]{fontenc}    
\usepackage[colorlinks]{hyperref}       
\usepackage{url}            
\usepackage{booktabs}       
\usepackage{amsfonts}       
\usepackage{nicefrac}       
\usepackage{microtype}      
\usepackage{xcolor}         

\usepackage{amsmath, amsthm, mathtools, bbm}
\usepackage[most]{tcolorbox}
\usepackage{enumitem}
\usepackage{subcaption}
\usepackage{tikz}
\usetikzlibrary{arrows.meta}
\usepackage[capitalize,noabbrev]{cleveref}
\crefname{appendix}{Appendix}{Appendices}
\Crefname{appendix}{Appendix}{Appendices}
\usepackage{multirow}
\usepackage{wrapfig}
\usepackage{algorithm}
\usepackage{algpseudocode}
\crefname{algorithm}{Algorithm}{Algorithms}
\Crefname{algorithm}{Algorithm}{Algorithms}

\usepackage{macros}

\usepackage[widebox]{restatelinks}

\title{Inverting the Bellman Equation: \\ From $Q$-Values to World Models}

\author{}

\begin{document}

\maketitle


\vspace{-45pt}
\begin{center}
    \begin{tabular}{@{}c@{\hspace{3em}}c@{\hspace{3em}}c@{}}
        {\bfseries Alistair Letcher}$^\alpha$ & {\bfseries Mattie Fellows}$^\alpha$ & {\bfseries Alexander D. Goldie}$^\alpha$ \\[15pt]
        {\bfseries Jonathan Richens}$^\beta$ & {\bfseries Jakob N. Foerster}$^\alpha$ & {\bfseries Oliver Richardson}$^\gamma$ \\[15pt]
        $^\alpha$ FLAIR, University of Oxford & \hspace*{-1.0em} $^\beta$ Google DeepMind & \hspace*{-1.0em} $^\gamma$ Mila, University of Montreal
    \end{tabular}
\end{center}
\vspace{20pt}

\begin{abstract}
    Model-based and model-free reinforcement learning are traditionally viewed as separate paradigms: instead of learning a model of the transition kernel $P$, model-free agents typically estimate value functions tied to a specific policy and reward. In this paper, we challenge this dichotomy by proving that value-based agents trained on a sufficiently rich set of reward functions, e.g. using goal-conditioned RL, implicitly encode a unique and accurate world model. To extract this model in practice, we introduce \textit{$P$-learning}, an inverse analogue to $Q$-learning that samples from an agent's $Q$-values, policies and rewards to decode its internal model of the environment. We then provide sufficient conditions on the type and number of goals for which agents encode the true kernel $P$, covering both stochastic and deterministic MDPs over finite or continuous state spaces. Even when our assumptions are violated, we empirically demonstrate that agents trained on a handful of reward functions encode accurate dynamics in \texttt{Reacher}, \texttt{MountainCar} and stochastic variants of \texttt{FourRooms}. Surprisingly, we find that policies trained exclusively on a \texttt{Reacher} agent's implicit world model are quasi-optimal on out-of-distribution, velocity-based goals despite position-only training -- suggesting that agents contain hidden generalisation capabilities and providing a new lens into the connection between model-based, model-free, and goal-conditioned RL.
\end{abstract}

\begin{figure}[ht]
   \centering
    \vspace*{-5pt}
    \includegraphics[width=1\textwidth]{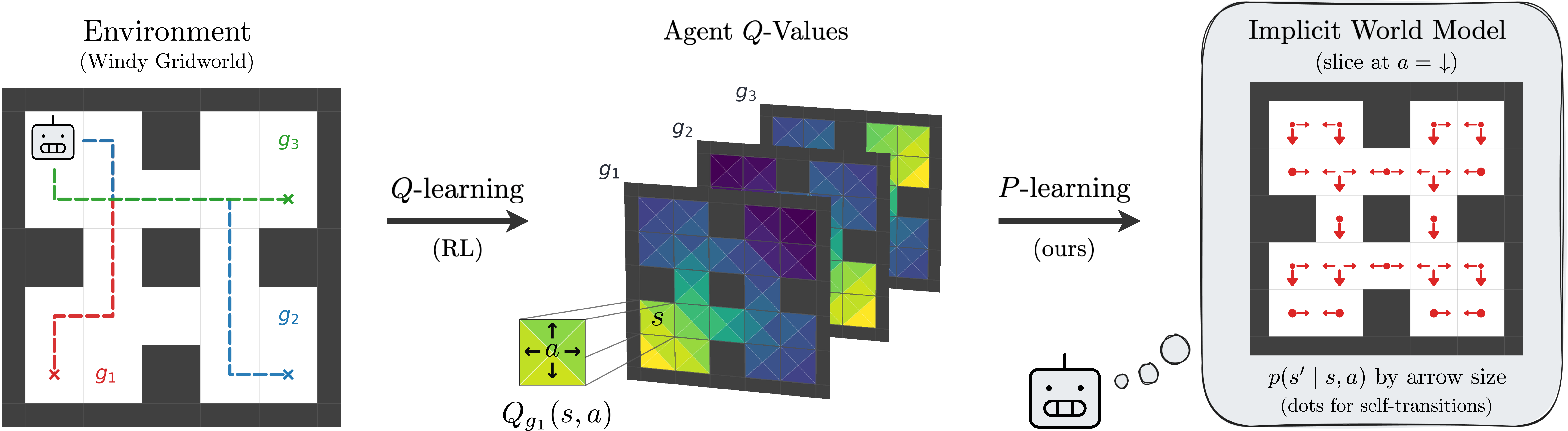}
    \vspace*{10pt}
    \caption{Illustration of $P$-learning: extracting the world model contained in an agent's $Q$-values. Code available at \href{https://github.com/aletcher/inverting-bellman}{\color{black}\nolinkurl{github.com/aletcher/inverting-bellman}}.}
    \label{fig:hero}
\end{figure}

\section{Introduction}\label{sec:intro}

The model-free paradigm of reinforcement learning (RL) has proven effective in training highly capable agents across a variety of domains \citep{SuttonBarto2018Reinforcement,Mnih2013,schulman_proximal_2017}, sidestepping the challenges associated with modelling complex world dynamics \citep{Lambert2020, Moerland2023}. However, the policies or value functions learnt by such agents are tied to reward functions, making it difficult to repurpose them for new tasks without additional data, or interpret what they \textit{understand} of the world -- as opposed to what they \textit{value}. In general, distinct world models (transition kernels) can give rise to identical value functions and optimal policies, an obstacle formalised as the \textit{value equivalence} problem \cite{Grimm2020, Grimm2021}.

For example, in a deterministic maze with a binary-reward goal, the optimal value of a state is $\gamma^d$, where $\gamma$ is the discount factor and $d$ is the minimal distance to the goal (measured by path-length). In particular, $Q$-values encode whether an action moves the agent closer to the goal, but do not pin down the state that an action leads to, i.e. the transition kernel, if there are equidistant states. This underdeterminacy implies that model-free agents do not uniquely encode the environment they inhabit, suggesting a fundamental divide between model-free and model-based RL.

However, agents are increasingly trained on a \textit{variety} of goals -- e.g. via successor features \cite{Barreto2017, Borsa2018}, forward-backward learning \cite{Touati2021} or goal-conditioned RL (GCRL) \cite{Schaul2015} -- which may provide additional constraints on the underlying dynamics. This raises the core question motivating this paper:
\begin{center}
    \textit{When do model-free agents implicitly encode an accurate model of their environment?}
\end{center}
We hypothesise that value-based agents trained on a sufficiently rich set of goals implicitly learn a \textit{unique} model of the world dynamics, even in stochastic and continuous MDPs, and that this model is \textit{accurate} insofar as $Q$-values are accurate. Taking GCRL as the broader framework for agents trained on goals parameterised by arbitrary reward functions, our work makes three contributions to test this hypothesis bridging model-free, model-based and goal-conditioned RL.

\textbf{Methodologically}, we introduce $P$-learning (\cref{sec:method}), an analogue to $Q$-learning illustrated in \cref{fig:hero}. Instead of iteratively updating value estimates for a fixed environment, $P$-learning updates a candidate world model (WM) to be progressively more consistent with fixed value functions, effectively inverting the Bellman equation. We make this precise by proving that tabular $P$-learning converges to a solution involving the Moore-Penrose pseudo-inverse (\cref{thm:inverse-bellman-tabular}). Our algorithm enables the efficient extraction of an implicit WM from a collection $(Q_g, \pi_g, r_g)_{g \in \cG}$ of $Q$-functions, policies and reward functions -- noting that policies are usually induced by $Q$-functions via $\text{argmax}$ or $\text{softmax}$, so $(Q, \pi, r)$ is often equivalent to $(Q, r)$ in practice.

\textbf{Theoretically}, we prove sufficient conditions on the goals and reward functions for which $P$ is uniquely identifiable from $Q$-values (\cref{sec:results}), breaking value equivalence. Four regimes emerge: in \emph{deterministic finite} MDPs, a single generic goal suffices (\cref{thm:deterministic}); in \emph{stochastic finite} MDPs, goals must span the state space (\cref{thm:finite}), with tight error bounds in both regimes when $Q$-values are $\ep$-approximate. In the \emph{deterministic continuous} case, a finite number of goals is, surprisingly, sometimes sufficient (\cref{thm:deterministic-gaussian,thm:deterministic-gaussian-generic}). In \emph{stochastic continuous} MDPs, identifiability is driven by \emph{coverage} instead of count: a goal set with non-empty interior for Gaussian rewards (\cref{thm:gaussian-stochastic}), or covering the state space for indicator rewards (\cref{thm:stochastic-sparse}). Taken together, our results show that methods like GCRL can bridge model-free and model-based RL, with the collection $(Q_g, \pi_g, r_g)_{g \in \cG}$ becoming informationally equivalent to the kernel $P$ when $\cG$ is sufficiently rich.

\textbf{Empirically}, we show that $P$-learning is highly effective in extracting accurate world models from agents, even with very few goals (\cref{sec:experiments}). We validate our finite-MDP theory across three increasingly stochastic variants of \texttt{FourRooms} (deterministic, windy and teleporting), with $|\cG| = 1$, $|\cG| = 4$ and $|\cG| = 20$ goals respectively sufficient for quasi-perfect recovery. In \texttt{Reacher}, with only $|\cG| = 4$ goals despite a continuous state space, we find that implicit WMs are even accurate over variables that rewards never directly depend on. Surprisingly, this accuracy transfers to near-optimal planning inside the WM for a range of out-of-distribution goals, including reaching specific angular velocities, despite position-only training. We study these ``implicit generalisation capabilities'' in \texttt{MountainCar} and show that inverting the setup, by training on velocity-based goals, yields a near-identical WM -- revealing that agents with widely different objectives can secretly encode similar models. Finally, we identify a strong correlation (Spearman $\rho = 0.98$) between an agent's performance and the accuracy of its internalised WM, suggesting that GCRL is an implicitly hybrid method linking model-free and model-based RL. Discussion of related work is deferred to \cref{sec:related}.
\section{Background}\label{sec:background}

\paragraph{Goal-Conditioned RL (GCRL).}
We consider goal-augmented MDPs $\cM = (\cS, \cA, \cG, P, \mu, r, \gamma)$ with finite or continuous state space $\cS$, arbitrary action space $\cA$ and goal set $\cG$, Markov transition kernel $P : \cS \times \cA \to \Delta(\cS)$, initial state distribution $\mu \in \Delta(\cS)$, discount factor $\gamma \in [0, 1)$, and goal-conditioned reward function $r : \cS \times \cG \to \R$. As in early GCRL literature \cite{Schaul2015}, goals $g \in \cG$ are specified by \textit{arbitrary} functions of the state, as opposed to exclusively \textit{indicator} goals defined by $r(s, g) = \delta_{gs}$, also called \textit{goal states} \cite{Eysenbach2022, Park2024}. Denoting the return of a trajectory $\tau = (s^t, a^t)_{t\geq 0}$ with respect to $g$ by $R(\tau, g) \eq \sum_{t \geq 0} \gamma^t r(s^{t+1}, g)$, the objective of GCRL is to learn a goal-conditioned policy $\pi : \cS \times \cG \to \Delta(\cA)$ such that each $\pi_g \eq \pi(\cdot,g)$ maximises the objective $J_g(\pi) \eq \E_{\tau \sim \pi_g} [ R(\tau, g) ]$, where $\tau \sim \pi_g$ is shorthand for independently sampling $s_0 \sim \mu$, $a_t \sim \pi(s_t, g)$, $s_{t+1} \sim P(s_t, a_t)$ for all $t$. The \emph{value} and \emph{action-value} functions are defined by
\begin{equation*}\label{eq:def-Q}
    V^\pi(s, g) \eq \E_{\tau \sim \pi_g} \left[ R(\tau, g) \mid s^0 = s \right]
    \ \text{and} \ \
    Q^\pi(s, a, g) \eq \E_{\tau \sim \pi_g} \left[ R(\tau, g) \mid s^0 = s, a^0 = a \right] \,.
\end{equation*}
Our finite-MDP results extend to goals with arbitrary \textit{termination indicators} \cite{White2017}, while some of our continuous-MDP results  include explicit termination conditions; see \cref{sec:proofs:termination} for details.

\paragraph{Q-Iteration.} For a fixed goal-augmented MDP, we define $\cQ \eq \{ Q : \cS \times \cA \times \cG \to \R \}$ as the space of action-value functions, and the goal-conditioned Bellman operators $\cT^\pi, \cT^\star : \cQ \to \cQ$ as
\begin{align*}
    \cT^\pi(Q)(s, a, g) &\eq \E_{s' \sim P(s,a), a'\sim \pi(s',g)} \big[ r(s', g) + \gamma Q(s', a', g) \big] \,,\\
    \cT^\star(Q)(s, a, g) &\eq \E_{s' \sim P(s,a)} \big[ r(s', g) + \gamma \sup\nolimits_{a' \in \cA} Q(s', a', g) \big] \,,
\end{align*}
with the corresponding Bellman and Bellman optimality equations $\cT^\pi(Q^\pi) = Q^\pi$, $\cT^\star(Q^\star) = Q^\star$. The aim of value-based methods is to approximate a solution to the Bellman optimality equation. A classical approach, known as \textit{$Q$-iteration} \cite{SuttonBarto2018Reinforcement,Riedmiller2005}, starts with an initial guess $Q_0$ and iteratively applies the Bellman operator $Q_{n+1} \eq \mathcal T^\star(Q_n)$ to get a sequence $(Q_n)_{n \in \mathbb N}$ converging to a fixed-point. In practice, with a parametric family $\T \mapsto Q_\T \in \cQ$, we fix a reference distribution $d \in \Delta(\cS \times \cA \times \cG)$, e.g. induced by an exploration policy, and measure fixed-point violation via the \emph{Bellman residual} $\mathcal{L}(\theta,\theta_n)\eq \| \mathcal{T}^\star(Q_{\theta_n}) - Q_\theta\|^2_d\,$, where $\|X\|^2_d \eq \frac12 \E_{d}[ X^2 ]$. One can then update $\T_{n}$ by minimising $\cL(\T, \T_n)$ with respect to $\T$. Writing $\delta_{\theta, {\theta_n}} \eq \cT^\star (Q_{\theta_n}) - Q_\theta$ for the \textit{temporal difference} error, the residual is usually minimised using gradient-based methods, with parameter updates driven by
\begin{align*}
    \nabla_\theta \mathcal{L}(\theta,\theta_n) = -\mathbb{E}_{(s,a,g)\sim d}\big[\delta_{\T, \T_n}(s,a,g)\nabla_\theta Q_\theta(s,a,g)\big] \,.
\end{align*}
Notably, this \emph{does not} correspond to gradient descent on a fixed objective; the target $\mathcal{T}^\star(Q_{\theta_n})$ depends on current parameters, so $\mathcal{L}(\theta,\theta_n)$ shifts with each iteration.
Estimating this expectation using environment samples yields a family of temporal difference (TD) algorithms \citep{Sutton1988} including $Q$-learning \citep{Watkins89,Watkins1992}, DQN \citep{Mnih2013,mnih2015humanlevel} and PQN \citep{Gallici2025}, adapted here with goal-conditioning. These differ in the iterative optimization method, choice of sampling distribution $d$, number of steps performed on $\mathcal{L}(\theta,\theta_n)$ before updating $\theta_n$, parametric family, and resulting convergence guarantees.
\section{\texorpdfstring{$P$-Learning}{P-Learning}}\label{sec:method}

While $Q$-iteration treats the kernel $P$ as fixed and searches for a function $Q^*$ that satisfies $\mathcal{T}^*(Q^*) = Q^*$, we consider the inverse problem of extracting an ``internal'' model $P^\star$ of the environment from a fixed agent with goal-conditioned $Q$-values, policy $\pi$, and known reward function $r$. Formally, let $\mathcal{P}\coloneqq\{ \mathcal{S}\times \mathcal{A}\rightarrow\Delta(\mathcal{S}) \}$ be the set of transition kernels, and consider a parametric family $\phi \mapsto P_\phi \in \cP$. We write $\cT^\pi_\phi = \cT^\pi_{P_\phi}$ for the corresponding Bellman operator, making explicit the dependence on WM parameters $\phi$. Our guiding observation is that for an agent ($Q, \pi)$, a model $P_\phi$ satisfying the Bellman equation $\cT^\pi_\phi(Q) = Q$ is compatible with the agent's behavior, and may thus be viewed as (one possible) subjective belief of the agent about the environment.

A natural objective is therefore to minimise the Bellman residual $\mathcal{L}(\phi) = \| \cT^\pi_\phi(Q) - Q \|_d^2 \qe \| \delta_\phi \|_d^2\,$, this time with respect to $\phi$. We call this \textbf{$P$-learning}. If $P_\phi$ is differentiable, we can minimise this loss with gradient-based methods, where the direction of parameter updates is driven by
\begin{equation*}\label{eq:p-gradient}
\nabla_\phi \cL(\phi) = \E_{(g,s,a)\sim d}\left[\delta_\phi(s,a,g) \nabla_\phi \cT^\pi_\phi(Q)(s,a,g) \right] \,. 
\end{equation*}
Unpacking this, we define the TD estimate $\hat\delta_\phi(s,a,g,s',a') \eq r(s',g) + \gamma Q(s',a',g) - Q(s,a,g)$, and prove in \cref{app:p_der} that under standard regularity assumptions,
\begin{equation*}
    \nabla_\phi \mathcal L(\phi) = \E_{g,s,a,s',a'}\bigl[\delta_\phi \hat\delta_\phi \nabla_\phi \log P_\phi(s'|s,a)\bigr] \,,
\end{equation*}
where $(s,a,g) \sim d$ and $s' \sim P_\phi(s, a), a' \sim \pi(s', g)$. For large or continuous state spaces, the gradient can instead be estimated via the reparameterisation trick \cite{Kingma2014, Rezende2014}.
Intuitively, while $Q$-learning finds a $Q$-function $Q_{\theta^\star}$ that is most consistent with the MDP (best satisfies the Bellman equation $\mathcal{T}^\star[Q_{\theta^\star}]=Q_{\theta^\star}$), $P$-learning finds a model $P_{\phi^\star}$ that is most consistent with a $Q$-function (best satisfies the Bellman equation $\mathcal{T}^\pi_{\phi^\star}[Q]=Q$). See \cref{alg:p-learning} for pseudocode. 


\textbf{What does $P$-learning converge to?}
Unlike $Q$-learning \citep{Baird1995}, $P$-learning solves a supervised learning problem using a \textit{stationary} objective $\mathcal{L}(\phi)$, inheriting the standard convergence guarantees of SGD algorithms.
\unskip\footnote{E.g. if the Robbins-Monro conditions are satisfied, stochastic gradient descent (SGD) converges to a stationary point \cite{Robbins51}.}
But what does the solution look like? Fixing the policy $\pi$ and omitting the superscript for brevity, let us first rewrite the goal-conditioned Bellman operator as
\begin{equation}\label{eq:bellman-general}
    \cT_{P_\phi}(Q)(s, a, g) = \E_{s' \sim P_\phi(s, a)} \left[ M(s', g) \right] \,, \quad \text{where} \quad M(s',g) \eq r(s', g) + \gamma V(s', g) \,.
\end{equation}
When $\cS, \cA, \cG$ are finite, we can assemble a matrix $M \in \R^{|\cG| \times |\cS|}$ with entries $M_{lk} = M(s'_k, g_l)$, and the Bellman equation becomes a linear system, viewing $Q(s,a) \in \R^{|\cG|}, P_\phi(s, a) \in \R^{|\cS|}$ as vectors:
\begin{equation}\label{eq:bellman-linear}
    Q(s,a,g_l) = \cT_\phi(Q)(s, a, g_l) = \sum\nolimits_k P_\phi(s,a,s'_k) M_{lk} \quad \implies \quad M P_\phi(s,a) = Q(s,a)
\end{equation}
for each $s, a$. Writing $J_\phi(s,a)$ for the Jacobian of $P_\phi(s,a)$ with respect to $\phi$, $P$-learning by gradient descent takes the explicit form $\phi_{n+1} = \phi_n - \alpha_n \sum_{s, a} J_\phi(s,a)^\top M^\top (M P_\phi(s,a) - Q(s,a))$. In the tabular setting where $\cS,\cA,\cG$ are finite and $P_\phi(s,a) = \phi(s,a)$ with $\phi : \cS \times \cA \to \R^{|\cS|}$, each Jacobian is the identity $J_\phi(s,a) = I$ and the update decouples, using a constant step size $\alpha$, as
\begin{equation}\label{eq:p-learning-tabular}
\textbf{Tabular P-learning:} \quad P_{n+1}(s,a) = P_n(s,a) - \alpha M^\top (M P_n(s,a) - Q(s,a)) \,.
\end{equation}
As shown below, tabular $P$-learning converges to a particularly friendly closed-form solution.

\begin{linked}[Convergence of $P$-learning]{theorem}{inverse-bellman-tabular}\label{thm:inverse-bellman-tabular}
    In the tabular setting, fix a policy $\pi$, a $Q$-function $Q \in \cQ$, initial $P_0(s,a) \in \Delta(\cS)$ for each $(s,a)$, and let $\sigma_{\max}$ denote the largest singular value of $M$ (Equation \ref{eq:bellman-general}). Then $P$-learning (Equation \ref{eq:p-learning-tabular}) with step size $0 < \alpha < 2/\sigma_{\max}^2$ converges to
    \[
        P_\infty(s,a) = M^+ Q(s,a) + \bigl(I - M^+ M\bigr) P_0(s,a) \,,
    \]
    where $M^+$ is the Moore--Penrose pseudo-inverse of $M$. Equivalently, $P_\infty$ is the orthogonal projection of $P_0$ onto the set of global minima $\mathcal{Z} = \arg\min_P \|\cT_P(Q) - Q\|_2^2$.
\end{linked}

\cref{thm:inverse-bellman-tabular} reveals key conditions for $P$-learning to converge to the true world model. The term $(I - M^+ M) P_0$ represents the underdetermination from value equivalence: when $M$ is rank-deficient ($M^+ M \neq I$), multiple WMs yield identical $Q$-values, and $P$-learning selects among them based on the initial guess $P_0$.
\footnote{Note that $P$-learning can desert the probability simplex if $M$ is rank-deficient; see \cref{sec:proof-simplex} for a projected variant.} 
When $M$ has \textbf{full rank}, the iteration converges to a \textit{unique} solution $M^+Q$, in which case $M^+$ can be viewed as an \textbf{inverse Bellman map}, noting that $M^+ = M^{-1}$ when $M$ is square. In this regime, $P$-learning can informally be viewed as a way to uniquely extract the agent's `beliefs' about the world (implicit model $P$) by sampling its `beliefs' about the value of states and actions (learnt function $Q$). If $Q$-values are moreover \textit{exact}, i.e. satisfy $\cT_P(Q) = Q$, then $P$-learning converges to the \textit{true} kernel $P$.
We provide intuition for how goals break value equivalence in \cref{sec:breaking-equivalence}, and formal guarantees in \cref{sec:results}.


\textbf{Deterministic MDPs.} If dynamics are deterministic, we can parameterise the model as a successor function $P_\phi : \cS \times \cA \to \cS$. The drawback is that backpropagation flows through $r(P_\phi(s,a), g)$ and $Q(P_\phi(s,a), a', g)$ via the chain rule, requiring these to be differentiable almost everywhere. This is usually satisfied when the $Q$-function is a neural network and $r$ is distance-based (including $r_g(s) = \mathbf{1}[||s-g|| < \sigma]$); otherwise, methods like evolutionary strategies \citep{Rechenberg78,sarkar2026} are appropriate. If $\cS$ is finite, $P$-learning can instead be performed via \textit{column-matching} (\cref{sec:finite}). The WM is then unique provided \textit{column-injectivity}, a much weaker requirement than full-rankness.


\section{Theoretical Results}\label{sec:results}

Our core theoretical results are sufficient conditions on the goals and reward functions $(\cG, r)$ that resolve the value equivalence problem via $Q$-values, meaning that a unique world model can be extracted from $(Q, \pi, r)$. When $Q$-values are \textit{exact}, we prove that the model coincides with the true transition kernel $P$. When $Q$-values are \textit{approximate}, we provide bounds on the accuracy of the extracted world model, with further results for \textit{arbitrary} $Q$-values in \cref{sec:proof-arbitrary-Q}.

For finite MDPs, our results characterise the number of goals $|\cG|$ needed to identify $P$ uniquely, for various families of reward functions $r$. For MDPs with continuous $\cS \subseteq \R^d$, uniqueness is typically driven not by count, but coverage of $\cS$ (excluding \cref{thm:deterministic-gaussian,thm:deterministic-gaussian-generic}). In this regime, our results therefore provide guarantees for goals sampled from a \textit{distribution} with sufficiently broad support, e.g. uniform over a bounded goal region, as is common in GCRL benchmarks like JaxGCRL \cite{Bortkiewicz2025}. However, the next section will demonstrate that even agents trained with a handful of goals, despite not covering $\cS$, often encode $P$ with high precision. We summarise our results in Table~\ref{tab:results}.

\textbf{Notation.} We write $Q$ for arbitrary goal-conditioned $Q$-values, with $M$ the corresponding function from \cref{eq:bellman-general}, and $Q^\pi$ for the exact $Q$-values of policy $\pi$, with $M = M^\pi$. We say $Q$ is \emph{$\ep$-approximate} if $\|Q(s, a, \cdot) - Q^\pi(s, a, \cdot)\|_1 \leq \ep$ for all $s, a$. The norm $\| \cdot \|$ without subscript refers to the standard Euclidean norm $\| \cdot \|_2$, and $B_\sigma(0)\coloneqq \{s \in \R^d \mid \lVert s \rVert<\sigma  \}$ is the open ball of radius $\sigma$, with $\cS \oplus B_\sigma(0)\coloneqq \{ s+x \mid s \in S, x \in B_\sigma(0)\}$. Finally, we write $\exists!P(Q, \pi, r, \gamma)$ or simply $\exists! P(Q)$ to assert that $P$ is the unique world model which satisfies the Bellman equation $\cT^\pi_P(Q) = Q$.

\begin{table}[ht]
    \centering
    \caption{\small Sufficient conditions on the goals $(\cG, r)$ for which $\exists! P(Q^\pi)$.}\vspace{3pt}
    \label{tab:results}
    \small
    \renewcommand{\arraystretch}{1.2}
    \newlength{\condw}\newlength{\refw}\newlength{\tmpw}
    \settowidth{\condw}{$\cG \supseteq \cS \oplus B_\sigma(0)$$^\star$}%
    \settowidth{\tmpw}{$\cG$ non-empty interior$^\star$}\ifdim\tmpw>\condw\setlength{\condw}{\tmpw}\fi
    \settowidth{\tmpw}{$|\cG|$ large, finite}\ifdim\tmpw>\condw\setlength{\condw}{\tmpw}\fi
    \settowidth{\refw}{\scriptsize\itshape Thms~00,00}%
    \newcommand{\refcell}[2]{\makebox[\condw][c]{#1}\hspace{0.0em}\makebox[\refw][c]{\scriptsize\itshape #2}}%
     \begin{tabular}{l@{\hskip 1.5em} | c@{\hskip 2.0em}c | @{\hskip 1.5em} r}
        \toprule
        State space $\cS$ & \textbf{Deterministic $P$} & \textbf{Stochastic $P$} & Reward $r$ \\
        \midrule
        \multirow{3}{*}{\textbf{Finite $\cS$}}
            & \refcell{$|\cG| \geq 1$}{Prop~\ref{prop:concrete-rewards}}
            & \refcell{$|\cG| \geq |\cS|$}{Prop~\ref{prop:concrete-rewards-stochastic}}
            & \textbf{Gaussian} \\
            & \refcell{$|\cG| \geq 1$}{Thm~\ref{thm:deterministic}a}
            & \refcell{$|\cG| \geq |\cS|$}{Thm~\ref{thm:finite}a}
            & \textbf{Generic} \\
            & \refcell{$|\cG| \geq |\cS|$}{Thm~\ref{thm:finite}b}
            & \refcell{$|\cG| \geq |\cS|$}{Thm~\ref{thm:finite}b}
            & \textbf{Spanning}$^\dagger$ \\
        \midrule
        \multirow{2}{*}{\shortstack[c]{\textbf{Continuous}\\\textbf{$\cS \subseteq \R^d$}}}
            & \refcell{$|\cG|$ large, finite}{Thms~\ref{thm:deterministic-gaussian},\ref{thm:deterministic-gaussian-generic}}
            & \refcell{$\cG$ non-empty interior$^\star$}{Thm~\ref{thm:gaussian-stochastic}}
            & \textbf{Gaussian} \\
            & \refcell{$\cG \supseteq \cS \oplus B_\sigma(0)$}{Thm~\ref{thm:deterministic-sparse}}
            & \refcell{$\cG \supseteq \cS \oplus B_\sigma(0)$$^\star$}{Thm~\ref{thm:stochastic-sparse}}
            & \textbf{Indicator} \\
        \bottomrule
    \end{tabular}
    \\[0.2em]
    {\footnotesize $^\star\,$For unconditional policies; see \cref{sec:proofs:unconditional}.}
    {\footnotesize $^\dagger\,$Including indicator goals, i.e. $r(s, g) = \delta_{sg}$.}
\end{table}

\subsection{Finite Space}\label{sec:finite}

Using our formulation of the Bellman equation \eqref{eq:bellman-general} as $\E_{s' \sim P(s,a)} [ M(s',g) ] = Q(s,a,g)$, we show that unique recovery of $P$ is guaranteed by different structural conditions on $M$ depending on kernel stochasticity, but not on the choice of parameterisation for $Q$-values or policies.

\paragraph{Deterministic MDPs.} For deterministic kernels $P : \cS \times \cA \to \cS$, the Bellman equation \eqref{eq:bellman-general} collapses to $M(P(s,a), g) = Q(s, a, g)$.
Writing $M(s') = M(s', \cdot)$ and $Q(s,a) = Q(s,a,\cdot)$, extracting $P$ reduces to identifying which column of $M$ matches $Q(s, a)$ via \textit{column-matching}, i.e.
\[ P(s,a) \eq \argmin_{s'} \norm{M(s') - Q(s,a)}_1 \,. \]
In particular, the $\argmin$ (and therefore the kernel $P$) is unique provided $M$ is \textbf{column-injective}: $M(s) \neq M(s')$ for all $s \neq s'$. To quantify injectivity, we define the \textit{column separation} of $M$ as $\gap(M) \eq \min_{s \neq s'} \norm{M(s) - M(s')}_1$. \cref{thm:deterministic} shows that even with a single goal, (a) $P$ is uniquely determined by exact $Q$-values $Q^\pi$ for generic (``almost all'') reward functions and (b) $P$ can also be recovered \textit{exactly} from $\ep$-approximate $Q$-values if $\ep$ is sufficiently small.

\begin{linked}[Deterministic, Finite MDPs]{theorem}{deterministic}\label{thm:deterministic}
    Fix any deterministic finite MDP and goal set $|\cG| \geq 1$.
    \begin{enumerate}[label=(\alph*)]
        \item For any policy $\pi$, the set of reward functions $r : \cS \times \cG \to \R$ for which $M^\pi$ is column-injective, and thus $\exists! P(Q^\pi)$, has full Lebesgue measure.
        \item If $Q$ is $\ep$-approximate and $M$ is column-injective with column separation $\gap$, the estimator $\hat{P}(s, a) = \operatorname{argmin}_{s'} \norm{Q(s,a) - M(s')}_1$ satisfies $\hat P = P$ for all $\ep < \gap(1 + \ga m)/2$, and any estimator fails to recover $P$ in the worst-case for $\ep \geq \gap/2$.
    \end{enumerate}
\end{linked}


As a corollary, for any base reward $r_0$ and any random vector $\xi \in \R^{|\cS|}$ whose distribution is absolutely continuous with respect to Lebesgue measure (e.g. Gaussian, or uniform on a bounded open region), the perturbed reward $r = r_0 + \xi$ satisfies $\exists!P(Q^\pi)$ almost surely (proof in \cref{sec:proof-deterministic}). This gives a recipe for ensuring identifiability by initialising rewards with noise, as demonstrated empirically in \texttt{FourRooms} (\cref{sec:exp-finite}). The same conclusion does not follow for general parametric families like indicator rewards; however, when the state space can be embedded in continuous space (e.g. in a 2D environment), we can guarantee $\exists!P(Q^\pi)$ for Gaussian rewards:

\begin{linked}[Gaussian Rewards]{proposition}{concrete-rewards}\label{prop:concrete-rewards}
    Fix any deterministic finite MDP, policy $\pi$, and variance $\sigma > 0$. For any injective map $\phi : \cS \hookrightarrow \R^d$, the single-goal set $\cG = \{g\}$ with Gaussian reward function $r_g(s) = \exp(-\|\phi(s) - g\|^2/2\sigma^2)$ satisfies $\exists! P(Q^\pi)$ for Lebesgue-almost every $g \in \R^d$.
\end{linked}

\paragraph{Stochastic MDPs.} For stochastic kernels, the Bellman equation \eqref{eq:bellman-linear} becomes a set of linear systems $M^\pi P(s,a) = Q^\pi(s,a)$, implying that $P$ is unique if $M$ is \textbf{full-rank}. In order to specify a concrete class of rewards for which $\exists!P(Q^\pi)$, we say that goals $(\cG,r)$ \textbf{span the state space} if $\rank(R) = |\cS|$, where $R_{lk} \eq r(s_k, g_l)$. This includes the common indicator goals \cite{Eysenbach2022, Park2024} given by $\cG = \cS$ and $r(s,g) = \delta_{sg}$, since $\rank(R) = \rank(I) = |\cS|$. Using this condition, we state our main result.

\begin{linked}[Finite MDPs]{theorem}{finite}\label{thm:finite}
    Fix any finite MDP and goal set $|\cG| \geq |\cS|$.
    \begin{enumerate}[label=(\alph*)]
        \item For any policy $\pi$, the set of reward functions $r : \cS \times \cG \to \R$ for which $M^\pi$ is full-rank, and thus $\exists!P(Q^\pi)$, has full Lebesgue measure.
        \item For any goals $(\cG, r)$ that span $\cS$ and any discount factor $\gamma \in [0, 1)$, the set of policies $\pi$ for which $M^\pi$ is full-rank has full Lebesgue measure. Similarly, for any $\pi$, the set of $\gamma \in [0, 1)$ for which $M^\pi$ is full-rank has full Lebesgue measure. In particular, there exist $\gamma^-, \gamma^+ \in (0, 1)$ such that $M^{\pi}$ is full-rank for all $\gamma < \gamma^-$ or $\gamma > \gamma^+$.
        \item If $Q$ is $\ep$-approximate and $M$ has full rank, the estimator $\hP_{ij} = M^{+} Q_{ij}$ satisfies
        \[
            \norm{\hP_{ij} - P_{ij}}_1 \leq \norm{M^+}_1 (1 + \ga m) \ep \,,
        \]
        and any estimator has worst-case error $(1+\ga)\ep/2$.
    \end{enumerate}
\end{linked}

While \cref{thm:finite}(a) is analogous to \cref{thm:deterministic}(a), showing that identifiability holds for \textit{generic} reward functions when $|\cG| \geq |\cS|$, part (b) shows that spanning goals are also sufficient if we shift genericity to policies or discount factors. We provide a stronger version of this result in \cref{sec:proof-finite-stoch} (\cref{prop:finite-exact}), including a \textit{finite} failure set over $\gamma$ and a treatment of optimal and entropy-regularised policies (functions of $Q$). Results for perturbed and Gaussian rewards are provided in \cref{sec:proof-concrete-rewards-stochastic}.

In some settings, such as single-goal contrastive RL \cite{Liu2025}, the actor learns a single \textit{unconditional} policy $\pi_g = \pi_{g'}$ for all $g,g'$, trained to maximise an environment-provided goal, but with a critic that encodes $Q$-values $Q^\pi_g$ for this policy with respect to all indicator goals $g \in \cS$. In this unconditional regime, Theorem \ref{thm:finite}(b) strengthens to $M^\pi$ \textit{always} being full-rank (unlike general policies, cf. \cref{sec:proof-singularity}). If we also assume indicator goals, the upper bound improves to $\| \hP_{ij} - P_{ij} \|_1 \leq 4\ep/(1 - 2\ep)$ for all $\ep < 1/2\,$ (\cref{prop:finite-approximate}), and $M^\pi$ reduces to the \textit{occupancy measure} \cite{Blier2021}.

\textit{What about arbitrary $Q$-values?} While our results are stated for exact or approximate $Q \approx Q^\pi$, note that a unique compatible world model can also be extracted for arbitrary $Q$-functions, e.g. during training, under similar conditions; see \cref{prop:duality-deterministic,prop:duality-stochastic} in \cref{sec:proof-arbitrary-Q}.

\textit{Are multiple goals necessary?} Could value equivalence instead be broken by $Q$-values for a single goal, but multiple horizons (discount factors $\gamma$) or behaviours (policies $\pi$), e.g. Multi-Rainbow \cite{Fedus2019} or Horde \cite{Sutton2011}? We prove in \cref{sec:proof-multi-horizon} that this is not generally possible: for any $|\cS| \geq 4$, there is a family of MDPs in which $P$ cannot be identified from the $Q$-values $\{Q^\pi_{r,\gamma} \mid \pi \in \Pi,\ \ga \in [0, 1)\}$ over \textit{all} policies $\pi$ and \textit{all} discount factors $\ga$, for \emph{any} single-goal reward function $r : \cS \to \R$.

\paragraph{Local MDPs.} Sitting between deterministic and fully-stochastic settings, we say that an MDP is \emph{$N$-local} if $\supp(s, a) \eq |\{s' : P(s' \mid s, a) > 0\}| \leq N$ for every $(s, a)$, e.g. a robot who can only move to a neighbourhood of its position in a single timestep. In \cref{sec:proof-local} we prove that $P$ can be identified uniquely in local MDPs for (a) $|\cG| \geq N-1$ generic goals when the support is known, and (b) $|\cG| \geq 2N-1$ otherwise. This generalises the stochastic case (where the support $N = |\cS|$ is trivially known) and the deterministic case (where $N = 1$ but the successor state is unknown).


\subsection{Continuous Space}\label{sec:continuous}

For continuous $\cS \subseteq \R^d$, the matrix $M$ is replaced by a possibly infinite family $\{m_g^\pi\}_{g \in \cG}$, and invertibility or column-injectivity generalises to whether this family is \textbf{measure determining} or \textbf{point separating} for stochastic or deterministic MDPs; see \cref{sec:proofs:continuous-family}. We prove results for two classes of goals $\cG \subseteq \R^d$: (sparse) indicator goals given by $r_g(s) = \mathbf{1}[||s-g|| < \sigma]$ with a threshold $\sigma > 0$, commonly used in GCRL, and Gaussian goals $r_g(s) = \exp(-||s-g||^2/2\sigma^2)$ with a variance $\sigma > 0$ controlling reward sparsity, which admit stronger guarantees. \cref{thm:deterministic-sparse}, stated below, is our most relevant result for MuJoCo environments like \texttt{Reacher}, which often use indicator goals and \textit{termination upon arrival}, i.e. a termination criterion $D_g(s) = \mathbf{1}[||s-g|| \geq \sigma]$ that prevents further accumulation of reward once the goal is reached; see formal definition in \cref{sec:proofs:termination}.

\begin{linked}{theorem}{deterministic-sparse}\label{thm:deterministic-sparse}
    Fix any continuous deterministic MDP, policy $\pi$, threshold $\sigma > 0$ and a goal set $\cG \supseteq \cS \oplus B_\sigma(0)$ with indicator rewards $r_g(s') = \mathbf{1}[\norm{s' - g} < \sigma]$ and termination upon arrival. Then $\{m_g^\pi : g \in \cG\}$ is point separating on $\cS$, hence $\exists! P(Q^\pi)$.
\end{linked}

Our full set of results, including those for stochastic kernels, are stated and proven in \cref{sec:proofs:continuous}. In the deterministic setting, we show that a sufficiently large but finite number of terminating Gaussian goals is sufficient to identify $P$ (\cref{thm:deterministic-gaussian}), while only $2d+1$ non-terminating Gaussian goals are enough when $V^\pi$ is real-analytic (\cref{thm:deterministic-gaussian-generic}). For stochastic kernels, \cref{thm:gaussian-stochastic,thm:stochastic-sparse} show that a goal set with non-empty interior suffices for Gaussian goals, and $\cG \supseteq \cS \oplus B_\sigma(0)$ for indicator goals, but only for unconditional policies. The challenging intersection of stochastic kernels and general (goal-conditioned) policies is beyond the scope of this paper.
\section{Experiments}\label{sec:experiments}

We verify that our findings hold empirically in \texttt{MountainCar} \cite{Towers2023} and MuJoCo \texttt{Reacher} \cite{Todorov2012}, both of which are continuous and deterministic. We defer the finite and stochastic setting to \cref{sec:appendix-experiments-gridworld}, where agents are shown to contain highly accurate world models using only $|\cG|=1,4,20$ goals for deterministic, local (windy) and fully stochastic (teleporting) variants of \texttt{FourRooms} \cite{Sutton1999} respectively. We train agents using PQN \cite{Gallici2025} with Hindsight Experience Replay (HER) \cite{Andrychowicz2017}, and extract each agent's implicit world model (WM) at the end of training using $P$-learning. Hyperparameters and environment details are in \cref{sec:appendix-experiments}. All results are reported over 10 seeds.

\begin{figure}[ht]
    \centering
    \vspace*{10pt}
    \raisebox{0.5em}{\includegraphics[width=0.49\textwidth]{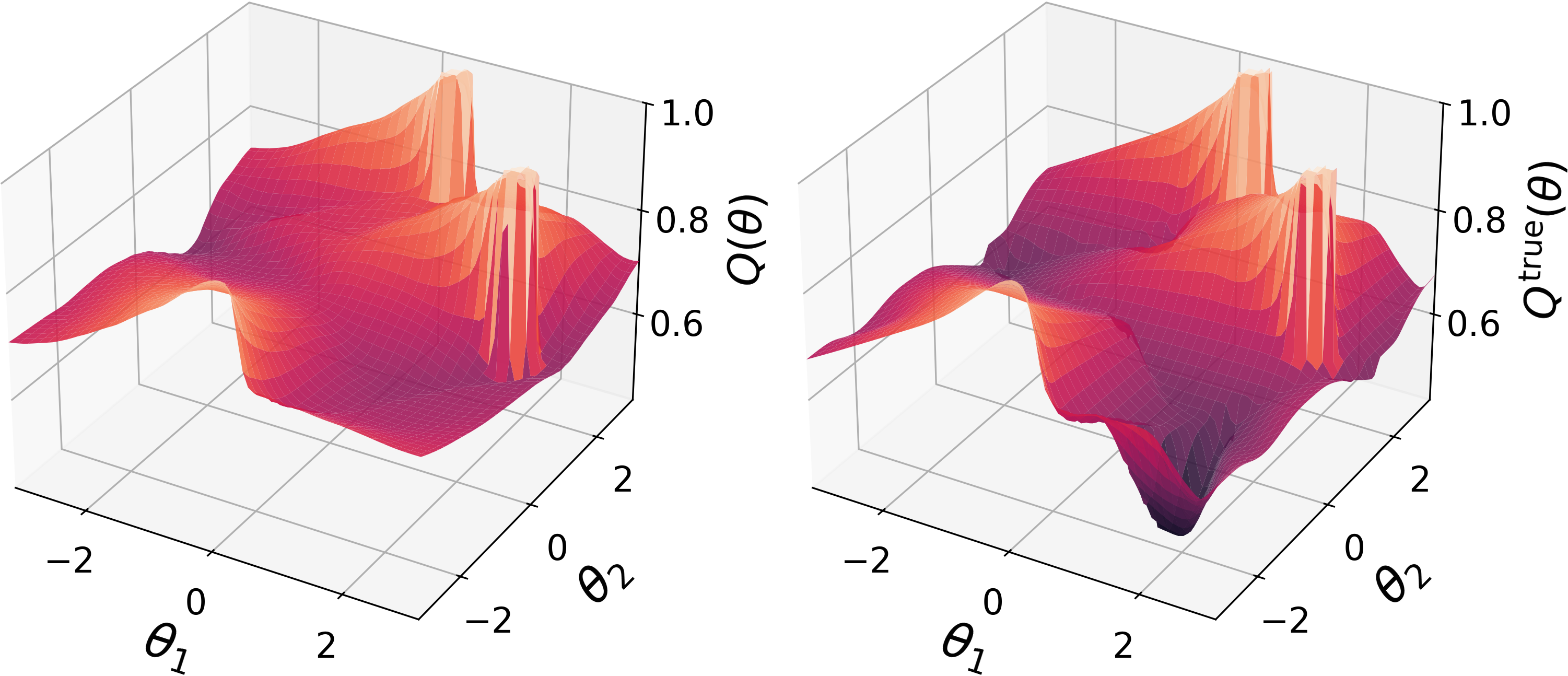}}
    \hfill
    \raisebox{0.5em}{\includegraphics[width=0.49\textwidth]{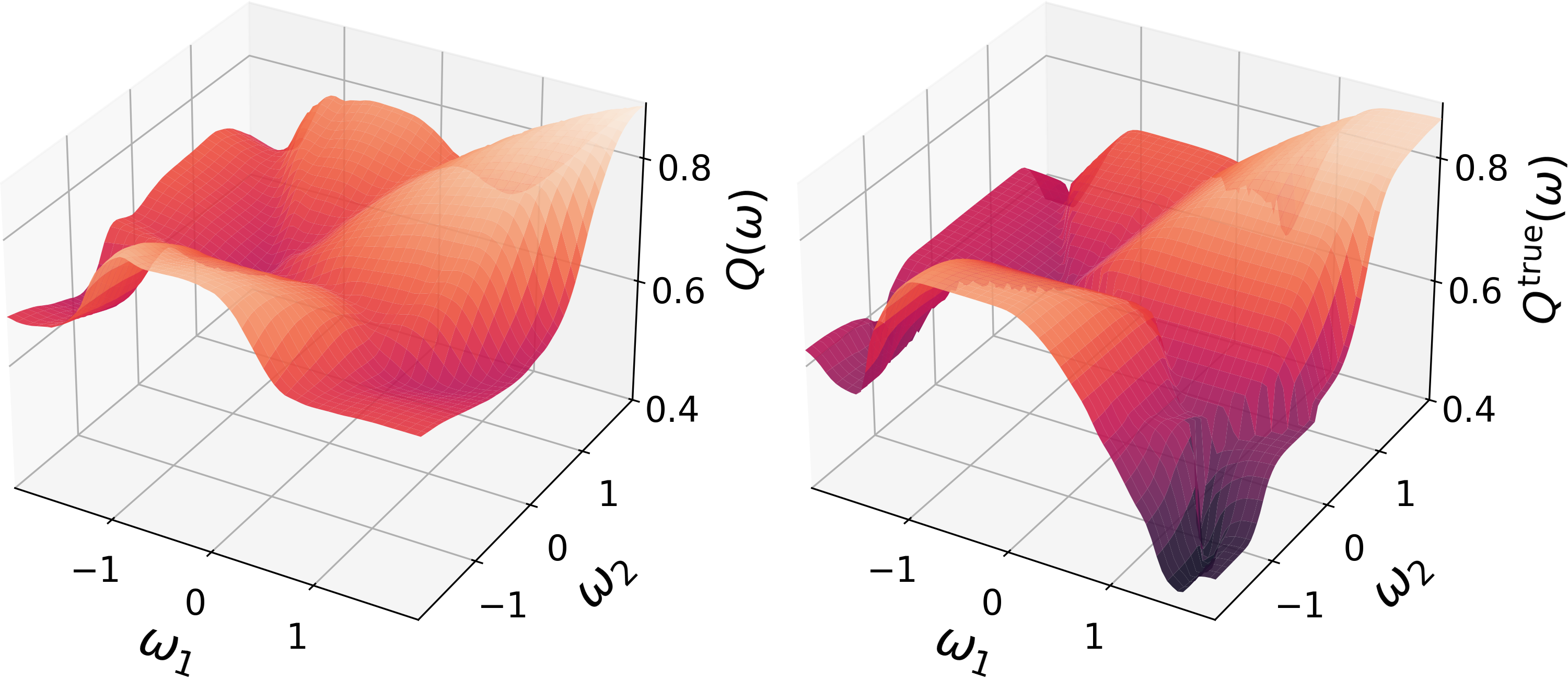}}
    \\[-0.6em]
    \makebox[0pt][c]{%
        \begin{tikzpicture}[overlay, xshift=-1.2em]
            \draw[-{Stealth[length=2.4mm,width=2.2mm]}, line width=1pt]
                (-0.5\textwidth-0.3em,  1.9em)
                  .. controls (-0.5\textwidth-1.6em, 1.0em) and (-0.5\textwidth-1.6em, -1.0em) ..
                (-0.5\textwidth-0.3em, -1.9em);
            \node[anchor=east, font=\small, inner sep=1pt]
                at (-0.5\textwidth-1.8em, 0) {$P$-learning};
        \end{tikzpicture}}%
    \\[-0.6em]
    \hspace*{-0.4em}
    \raisebox{1.2em}{\includegraphics[width=0.49\textwidth]{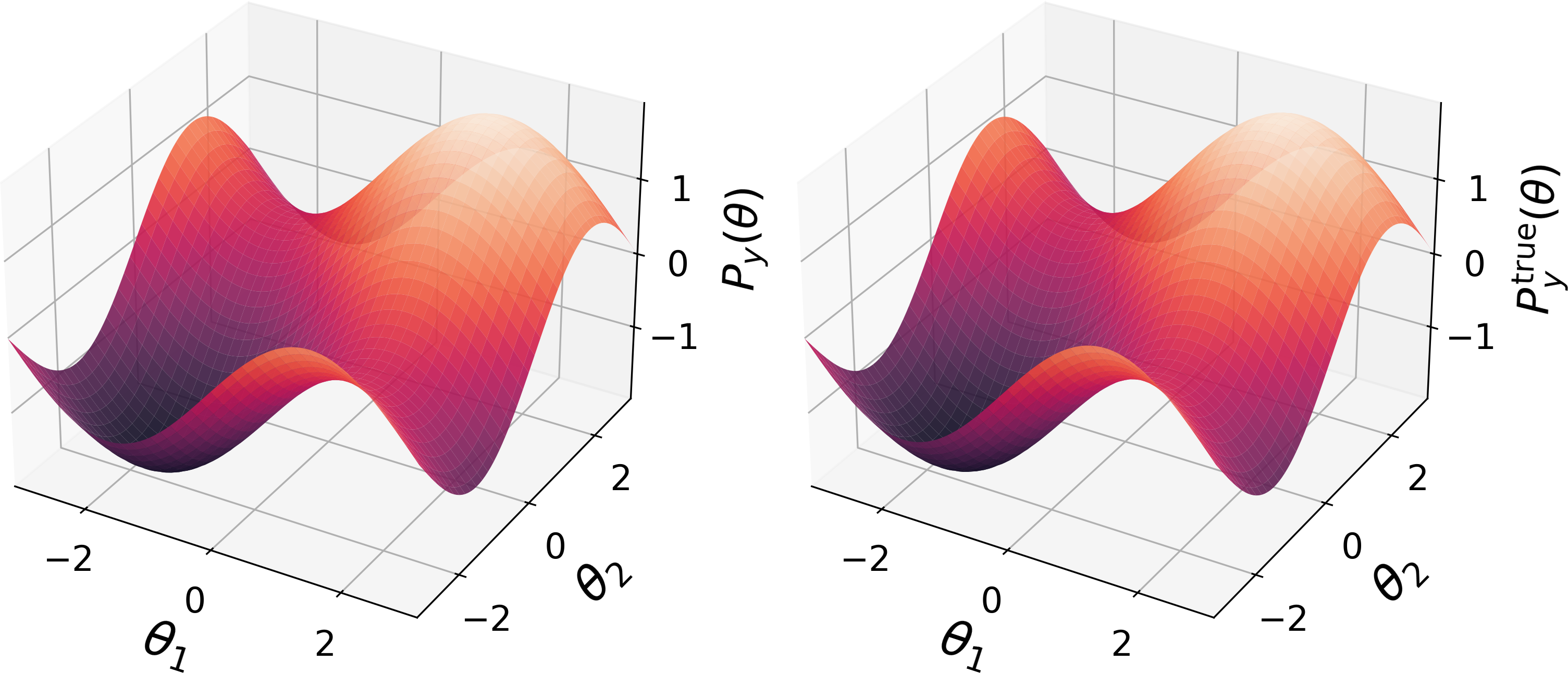}}
    \hspace*{0.2em}
    \includegraphics[width=0.49\textwidth]{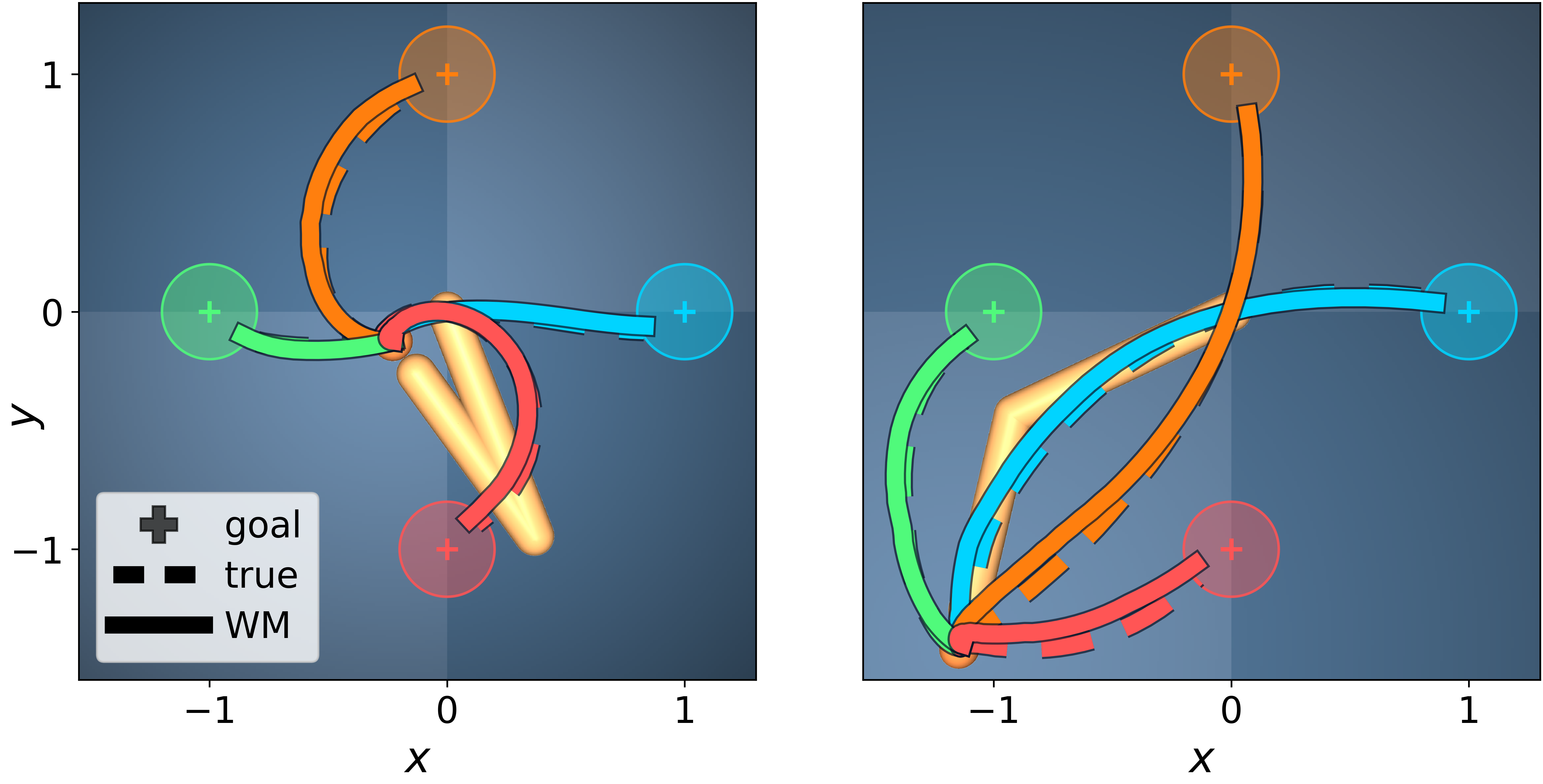}
    \caption{$Q$-values (top) and implicit WM (bottom) of a \texttt{Reacher} agent trained on $|\cG| = 4$ goals. Learnt values $Q_g(\theta,\om,a)$ are coarsely similar ($\texttt{NMSE} = 5.7 \times 10^{-1}$) to ground truth $Q^\text{true}$, shown for $g = (0, 1), a = (1,1)$ and two slices $\om = (1, -1), \theta = (1, -1)$. Nevertheless, the extracted WM $P(\T, \om, a)$ is highly accurate ($\texttt{NMSE} = 1.2 \times 10^{-4}$), shown for prediction of $y$-position $P_y$ vs true $P_y^\text{true}$ (same slice). Bottom right: WM/true rollouts of goal-conditioned policy for two start states.}
    \label{fig:reacher-results}
\end{figure}
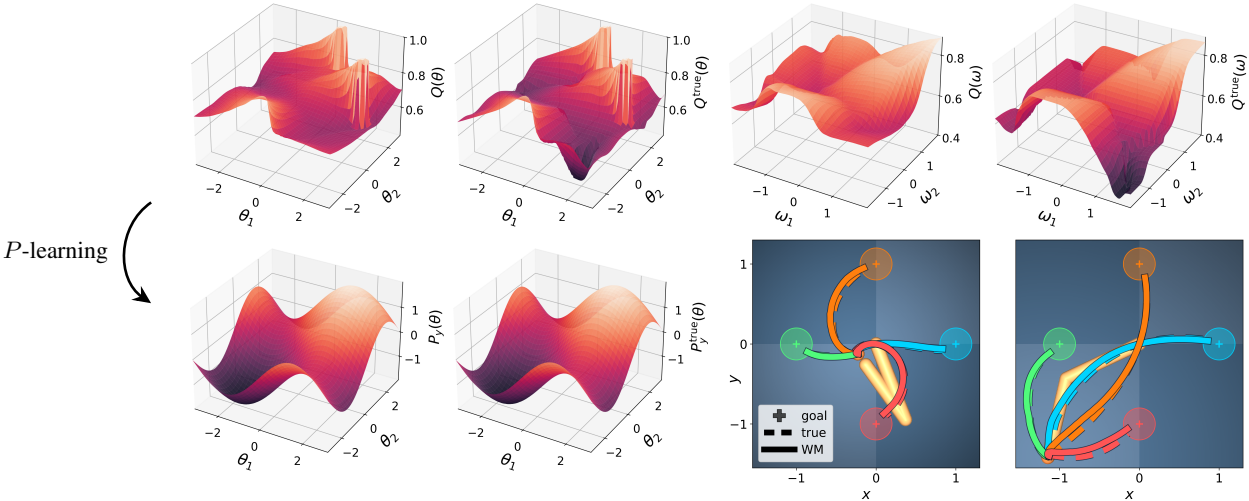

\subsection{Reacher}\label{sec:exp-reacher}

Our theoretical guarantees for continuous $\cS \subseteq \R^d$ require an infinite number of indicator goals covering all dimensions of $\cS$. However, we find that accurate world model extraction in Reacher is feasible even with a small number $|\cG| = 4$ of goals defined only with respect to fingertip position $(x,y) = (\cos\theta_1 + \cos\theta_2, \sin\theta_1 + \sin\theta_2)$. We consider the goals with unit distance in the cardinal directions, $\cG = \{(\pm 1,0), (0, \pm 1)\}$, covering a limited portion of the fingertip space $[-2, 2]$. The indicator (sparse) reward function is $r_g(\T,\om,x) = \mathbf{1}\{\|x-g\| < \sigma\}$, with $\sigma = 0.2$ matching ReacherHard in MuJoCo Playground \cite{MujocoPlayground2025}, and episodes terminate upon reaching the goal region.

\begin{wrapfigure}{r}{0.46\textwidth}
    \centering
    \vspace{-1em}
    \includegraphics[width=0.42\textwidth]{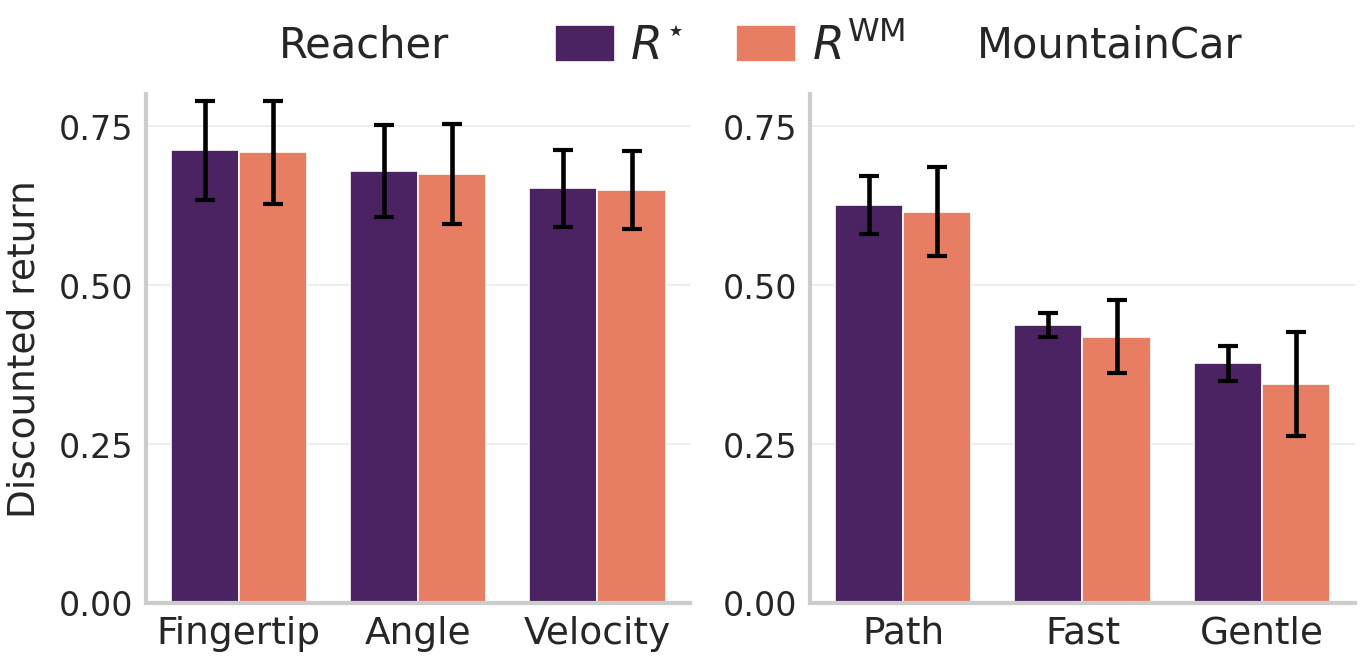}
    \captionsetup{margin={0.2em,0.0em}}
    \caption{Mean return $\pm$ SE (10 training seeds and 512 env. resets), for optimal $(R^\star)$ vs WM-trained $(R^\text{\tiny WM})$ policies on 3 unseen goals.}
    \vspace{-0.4em}
    \label{tab:unseen-tasks}
\end{wrapfigure}

We visualise slices of the learnt $Q$-values, extracted WM and trajectory rollouts in \cref{fig:reacher-results}.
Despite imperfect $Q$-values (normalised mean squared error $\texttt{NMSE} = 5.7 \times 10^{-1}$), the WM matches ground truth with extremely high fidelity ($\texttt{NMSE} = 1.2 \times 10^{-4}$).
To further validate accuracy, and determine whether $P$-learning reveals hidden generalisation capabilities, we train policies exclusively inside the extracted WM using value iteration, for three unseen goals that are increasingly far from the training distribution: \textbf{(i) far fingertip:} reach fingertip position $(x, y) = (\sqrt{2}, \sqrt{2})$, which is twice as far as any training goal; \textbf{(ii) target angle:} reach joint configuration $\T = (0, \pi)$, noting that distinct angle configurations can induce identical positions; and \textbf{(iii) target velocity:} reach velocity $\om = (2, 2)$, where $\omega$ covers a different dimension as the training goals. Despite the agent only training on position-based goals, the WM-learnt policy is quasi-optimal on all three tasks, as shown in \cref{tab:unseen-tasks}. Notably, the second and third goals would be impossible to achieve using position-based successor features, since the optimal $Q$-functions for these goals is not a linear combination of training $Q$-functions.

\begin{figure}[ht]
    \centering
    \includegraphics[width=\textwidth]{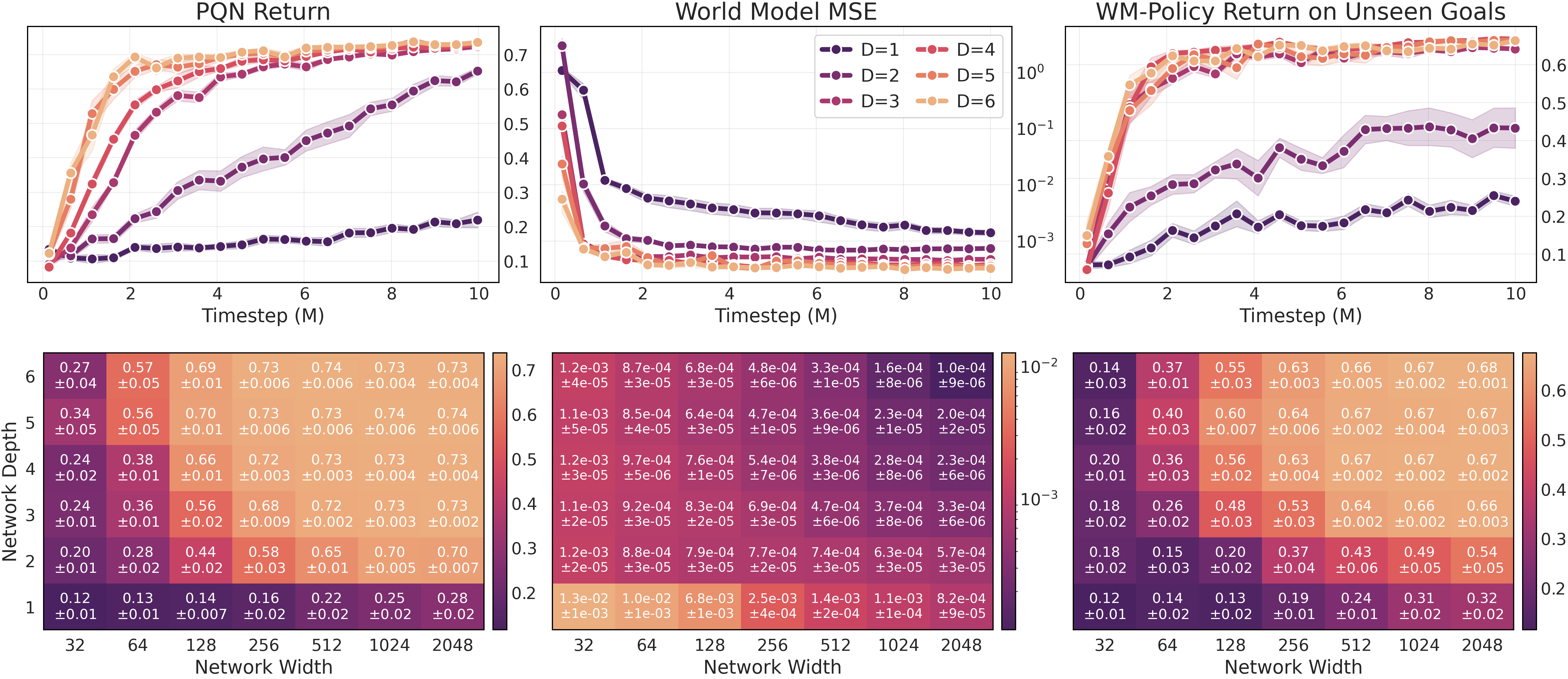}
    \caption{Agent return on training goals (left), implicit world model MSE (middle) and return of WM-extracted policies on unseen goals (right), for a \texttt{Reacher} agent trained with just $|\cG|=4$ goals. Top: metrics as a function of training time for fixed width $W = 512$ and increasing depth $D$. Bottom: Heatmap over all widths/depths at the end of training, with mean $\pm$ SE over 10 seeds. The Spearman correlation coefficient between agent return and WM error is $\rho = -0.98 \, [-0.99, -0.95]$.\vspace{-17pt}}
    \label{fig:reacher-correlation}
\end{figure}

To further study the relationship between model-free training and implicit world models, we analyse whether more performant agents tend to encode more accurate WMs. We follow recent work suggesting that network capacity improves goal-conditioned performance \cite{wang2026} and plot agent return, MSE of the extracted WM, and return of a WM-trained policy on unseen goals, against architecture capacity ($Q$-network depth/width) in \cref{fig:reacher-correlation}. We run each configuration over ten seeds, and fix all other hyperparameters. The resulting Spearman rank correlation coefficient between agent return and WM error is $\rho = -0.98$ with 95\% confidence interval $[-0.99, -0.95]$, and similarly $\rho = +0.95$ $[+0.89, +0.97]$ for agent return and WM-policy return on unseen goals. These strong correlations suggest that performant goal-conditioned agents tend to have better world models and better implicit generalisation capabilities, and vice-versa. We expand this analysis in \cref{sec:appendix-reacher-arch}.

\subsection{MountainCar}\label{sec:exp-mountaincar}\vspace{-3pt}

In \texttt{MountainCar}, the environment goal is to reach position $x = 0.5$ under 200 timesteps. Unlike \texttt{Reacher}, which has goal-conditioning baked in, we augment the environment to support arbitrary goals. We train a \texttt{MountainCar} agent using $4$ linearly spaced, position-based goals, with a sparse reward function $r_g(x, v) = \mathbf{1}\{|x-g| < \sigma\}$, where $g \in [-1.2, 0.6]$, with termination upon arrival and threshold $\sigma = 0.1$.
The original environment goal corresponds exactly to $g = 0.6$. Again, the dynamics extracted from a trained PQN agent match the true transition kernel with high fidelity ($\texttt{NMSE} = 6.7 \times 10^{-3}$). To further validate accuracy, we consider three out-of-distribution goals: \textbf{(i) Fast car:} reach velocity $v = 0.06$; \textbf{(ii) Gentle car:} reach the top of the hill $x = 0.6$ without exceeding $|v| = 0.05$ (the unconstrained optimal policy uses $|v| \geq 0.06$); \textbf{(iii) Shortest path:} find the shortest path to one of two goals $x = -1.0$ and $x = 0.4$, which, as noted by \citet{Touati2021}, could not be achieved using successor features based on the indicator-goal family.

\begin{wrapfigure}{r}{0.54\textwidth}
    \centering
    \vspace{-0.8em}
    \includegraphics[width=0.48\textwidth]{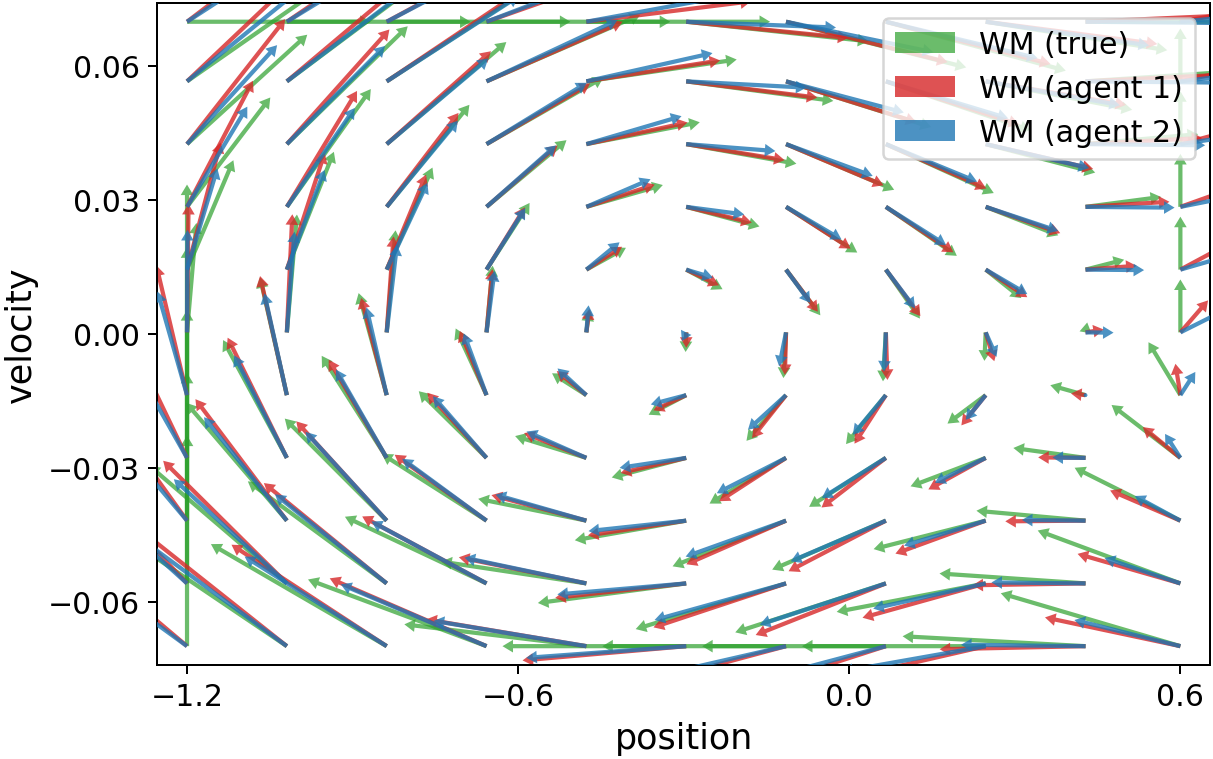}
    \captionsetup{margin={2em,0pt}}
    \caption{Quiver plots of WMs extracted from two agents trained with position- and velocity-based goals, vs ground truth, for action $a = \text{right}$. Each arrow starts from a state $s$ and points to the predicted or true next-state $s' = P(s,a)$.}
    \vspace{-\intextsep}
    \label{fig:mountaincar-platonic}
\end{wrapfigure}

Again, planning inside the implicit WM reveals implicit generalisation capabilities (\cref{tab:unseen-tasks}). Investigating further, we train a second \texttt{MountainCar} agent with four \mbox{\textit{velocity}-based goals}. The resulting implicit WM is near-identical to that of the position-trained agent, namely, they are \textasciitilde 42 times closer to each other ($\texttt{NMSE} = 1.7 \times 10^{-4}$) than to ground truth ($\texttt{NMSE} = 7.2 \times 10^{-3}$), as illustrated in \cref{fig:mountaincar-platonic}. This leads us to an informal ``local-global'' hypothesis, left for future exploration: \textit{value-based agents, trained to predict returns on a small number of local goals (i.e. with reward functions defined on a subset of variables), tend to encode an accurate world model over all dependent environment variables.}
\vspace{-3pt}
\section{Related Work}\label{sec:related}

\textbf{Value equivalence.}
\citet{Grimm2020} introduced the value equivalence (VE) principle, and showed that the equivalence class collapses to a single point for value functions that form a basis of $\R^{|\cS|}$ combined with $|\cA|$ pointwise-linearly-independent policies, neither of which we assume. Their work focuses on learning a surrogate model to enhance planning using environment samples, while ours addresses the problem of extracting an agent's internal world model with access only to $(Q, \pi, r)$. Algorithms such as MuZero \citep{Schrittwieser2020}, VAML \citep{Farahmand2018,Lambert2020} and unified frameworks \citep{Freed2024} also learn value-equivalent models without addressing identifiability. \citet{Fellows2024} demonstrate that Bayesian model-free methods can learn Bayes-optimal policies, implying that model-free approaches implicitly encode a posterior over environment dynamics, and hence a space of world models. Relatedly, delayed $Q$-learning \citep{Strehl2006} achieves PAC sample complexity comparable to model-based RL despite never representing $P$ explicitly. Conversely, \citet{Zhu2024} exhibit MDP families where $P$ admits exponentially smaller circuit representations than $Q^\star$, which motivates extracting $P$ as a compact summary of an agent.

\textbf{GCRL.} GCRL has emerged as a popular field wherein agents are trained on a number of tasks \citep{Kaelbling1993LearningTA, Andrychowicz2017, Blier2021}, with the bulk of recent research focused on enhancing the performance of goal-conditioned agents, e.g. by scaling compute usage \citep{ghugare_role_2026}, goal set \cite{Matthews2026} or network size \citep{wang2026}. Our work takes a different tack by analysing what agents encode during and after training, and showing that their implicit representation of the world improves with the number of goals, network size and performance.

\textbf{World models from agents.}
\citet{Richens2025} proved that $\ep$-optimal policies for sufficiently diverse \emph{multi-step} goals of depth $d$ recover approximate dynamics, but require competence on $dn^2m$ goals; they do not address the more realistic setting of \emph{single-step} goals, nor continuous MDPs. \citet{Adamczyk2025} showed that converged $Q$-values encode dynamics under a ``reverse Lipschitz'' assumption on $V^\pi$ which rarely holds in practice, and is limited to deterministic MDPs and a single goal. \citet[Theorem~F.117]{Turner2022} proved that $V^\star$ recovers deterministic MDPs up to isomorphism, but not in stochastic environments. \citet{Bush2025} provide mechanistic-interpretability evidence that model-free agents develop implicit forward models, but do not address uniqueness or extraction.

\textbf{Decoupling dynamics from reward.}
The successor representation \citep{Dayan1993} and successor features \citep{Barreto2017,Borsa2018} decompose value functions into transition- and reward-dependent components, but only address families of reward functions that can be decomposed as linear functions of shared features. This excludes e.g. the indicator functions in continuous environments \cite{Towers2023}, which would require infinitely many features to represent. Forward-backward learning \citep{Touati2021} relaxes this constraint significantly, and the successor measure of \citet[p.~7]{Lehnert2020} and \citet[Prop.~1]{Blier2021} was shown to be invertible in the special case of unconditional $\pi$, indicator goals and exact $Q$, but does not address identifiability or extraction. Reward-free RL \citep{Jin2020} and inverse RL \citep{Cao2021} solve orthogonal problems to ours: \textit{reward-free} exploration produces a dataset sufficient to plan for any downstream reward, whereas we show the dual phenomenon: diverse \emph{reward-rich} training implicitly encodes $P$. Inverse RL recovers $r$ from $(P,\pi)$, while we recover $P$ from $(r,\pi,Q)$. Indeed, RL, Q-learning, and P-learning can all be thought of as different resolutions to the same inconsistency (the Bellman residual) between a world model $P$, action-values $Q$, and rewards $r$ -- an instance of a general pattern that also has analogues in decision theory and supervised learning \cite{richardson_lir_26}.

\section{Conclusion}\label{sec:conclusion}

In this work, we provided sufficient conditions under which it is possible to recover environment dynamics from the $Q$-values of a goal-conditioned model-free agent, for a variety of MDP classes (deterministic or stochastic, discrete or continuous). We introduced a novel method, $P$-learning, which enables the extraction of this implicit world model. We subsequently showed, empirically, the accuracy and effectiveness of $P$-learning in a number of RL environments, despite goal sets that significantly breach our theoretical guarantees. We demonstrated that this WM leads to implicit generalisation capabilities hidden in the agent's value functions, with zero-shot transfer to reward functions defined on orthogonal variables. Finally, we provided evidence for a relationship between the performance of an agent, and the fidelity of its underlying world model.

\textbf{Limitations.} Our results come with a number of caveats. First, our theoretical results for stochastic MDPs with continuous state spaces only apply to unconditional policies, and while we expect them to generalise, we leave them as open conjectures for future work. Secondly, our work does not encompass partially observable MDPs, or model-free agents that do not learn $Q$-values. Third, we have not run experiments to test whether $P$-learning scales to very large, finite environments. Finally, $P$-learning is not a substitute for model-based RL; instead, we focus on the conceptual and empirical connection between model-based, model-free and goal-conditioned RL.

\textbf{Future Work.} We hope our results will inspire a deeper understanding of the implicit generalisation capabilities lurking in model-free agents; hybrid methods that leverage them; and interpretability or auditing tools based on internal world models, to ensure that agents can efficiently be repurposed when they are trained using reward functions that are, perhaps inevitably, misspecified.

\section*{Acknowledgements}
We are grateful to David Abel for his timely suggestions and references to existing work, Michael Matthews for his JAX implementation of PQN augmented with HER, Maria Kostylew for her comments and support, Uljad Berdica for his feedback and world model expertise, Shimon Whiteson for feedback on early drafts, and Matthew Farrugia-Roberts for helpful early conversations.

\textbf{AL} conducted this work as part of the MATS Program (ML Alignment \& Theory Scholars), and is generously supported by the Cooperative AI PhD Fellowship. \textbf{MF} is funded by the UKRI Engineering and Physical Sciences Research Council EP/Y028481/1. \textbf{AG} is funded by the EPSRC Centre for Doctoral Training in Autonomous Intelligent Machines and Systems EP/S024050/1.

\section*{Author Contributions}

\textbf{AL} led the project, designed the method, proved the theoretical results, ran the experiments and wrote the paper. \textbf{MF} and \textbf{AG} helped with writing, editing and experiment design. \textbf{JR} and \textbf{JF} provided guidance and helped frame the paper throughout. \textbf{OR} contributed to the writing, editing, mathematical precision, and provided guidance throughout.

{\small
\bibliography{refs}

@inproceedings{Richens2025,
  title = {General Agents Need World Models},
  booktitle = {Forty-Second {{International Conference}} on {{Machine Learning}}},
  author = {Richens, Jonathan and Everitt, Tom and Abel, David},
  year = {2025}
}

@PhdThesis{Watkins89,
	author =       "Watkins, Christopher John Cornish Hellaby",
	title =        "Learning from Delayed Rewards",
	school =       "King's College, University of Cambridge",
	year =         "1989",
	address =   "Cambridge, UK",
	month =     "May",
	bib2html_rescat = "Parameter",
}

@inproceedings{Eysenbach2022,
  title={Contrastive Learning as Goal-Conditioned Reinforcement Learning},
  author={Eysenbach, Benjamin and Zhang, Tianjun and Levine, Sergey and Salakhutdinov, Ruslan},
  booktitle={Advances in Neural Information Processing Systems},
  year={2022},
}

@article{Moerland2023,
author = {Moerland, Thomas M. and Broekens, Joost and Plaat, Aske and Jonker, Catholijn M.},
title = {Model-based Reinforcement Learning: A Survey},
year = {2023},
journal = {Foundations and Trends in Machine Learning},
}

@inproceedings{Todorov2012,
  title={MuJoCo: A physics engine for model-based control},
  author={Todorov, Emanuel and Erez, Tom and Tassa, Yuval},
  booktitle={2012 IEEE/RSJ International Conference on Intelligent Robots and Systems},
  year={2012},
  organization={IEEE},
}

@misc{MujocoPlayground2025,
  title = {MuJoCo Playground: An open-source framework for GPU-accelerated robot learning and sim-to-real transfer.},
  author = {Zakka, Kevin and Tabanpour, Baruch and Liao, Qiayuan and Haiderbhai, Mustafa and Holt, Samuel and Luo, Jing Yuan and Allshire, Arthur and Frey, Erik and Sreenath, Koushil and Kahrs, Lueder A. and Sferrazza, Carlo and Tassa, Yuval and Abbeel, Pieter},
  year = {2025},
  url = {https://github.com/google-deepmind/mujoco_playground}
}

@inproceedings{Strehl2006,
author = {Strehl, Alexander L. and Li, Lihong and Wiewiora, Eric and Langford, John and Littman, Michael L.},
title = {PAC model-free reinforcement learning},
year = {2006},
booktitle = {Proceedings of the 23rd International Conference on Machine Learning},
}

@inproceedings{Zhu2024,
title={On Representation Complexity of Model-based and Model-free Reinforcement Learning},
author={Hanlin Zhu and Baihe Huang and Stuart Russell},
booktitle={The Twelfth International Conference on Learning Representations},
year={2024},
}

@InProceedings{Jin2020,
  title = 	 {Reward-Free Exploration for Reinforcement Learning},
  author =       {Jin, Chi and Krishnamurthy, Akshay and Simchowitz, Max and Yu, Tiancheng},
  booktitle = 	 {Proceedings of the 37th International Conference on Machine Learning},
  year = 	 {2020},
  series = 	 {Proceedings of Machine Learning Research},
}

@article{mnih2015humanlevel,
	author = {Mnih, Volodymyr and Kavukcuoglu, Koray and Silver, David and Rusu, Andrei A. and Veness, Joel and Bellemare, Marc G. and Graves, Alex and Riedmiller, Martin and Fidjeland, Andreas K. and Ostrovski, Georg and Petersen, Stig and Beattie, Charles and Sadik, Amir and Antonoglou, Ioannis and King, Helen and Kumaran, Dharshan and Wierstra, Daan and Legg, Shane and Hassabis, Demis},
	description = {Human-level control through deep reinforcement learning - nature14236.pdf},
	issn = {00280836},
	journal = {Nature},
	month = feb,
	publisher = {Nature Publishing Group, a division of Macmillan Publishers Limited. All Rights Reserved.},
	title = {Human-level control through deep reinforcement learning},
	year = 2015
}

@inproceedings{Baird1995,
author = {Baird, Leemon C.},
title = {Residual algorithms: reinforcement learning with function approximation},
year = {1995},
booktitle = {Proceedings of the International Conference on Machine Learning},
}

@misc{Matthews2026,
      title={Goal-Conditioned Agents that Learn Everything All at Once}, 
      author={Michael Matthews and Matthew Jackson and Michael Beukman and Thomas Foster and Alistair Letcher and Scott Fujimoto and Cédric Colas and Jakob Foerster},
      year={2026},
      eprint={2605.23551}
}

@inproceedings{White2017,
author = {White, Martha},
title = {Unifying task specification in reinforcement learning},
year = {2017},
booktitle = {Proceedings of the 34th International Conference on Machine Learning - Volume 70},
series = {ICML'17}
}

@article{Mohamed2020,
author = {Mohamed, Shakir and Rosca, Mihaela and Figurnov, Michael and Mnih, Andriy},
title = {Monte Carlo gradient estimation in machine learning},
year = {2020},
journal = {J. Mach. Learn. Res.},
}

@book{Bochner1955,
  author    = {Bochner, Salomon},
  title     = {Harmonic Analysis and the Theory of Probability},
  publisher = {University of California Press},
  year      = {1955},
}

@InProceedings{Rechenberg78,
author="Rechenberg, I.",
title="Evolutionsstrategien",
booktitle="Simulationsmethoden in der Medizin und Biologie",
year="1978",
publisher="Springer Berlin Heidelberg",
}

@misc{sarkar2026,
      title={Evolution Strategies at the Hyperscale}, 
      author={Bidipta Sarkar and Mattie Fellows and Juan Agustin Duque and Alistair Letcher and Antonio León Villares and Anya Sims and Clarisse Wibault and Dmitry Samsonov and Dylan Cope and Jarek Liesen and Kang Li and Lukas Seier and Theo Wolf and Uljad Berdica and Valentin Mohl and Alexander David Goldie and Aaron Courville and Karin Sevegnani and Shimon Whiteson and Jakob Nicolaus Foerster},
      year={2026},
      eprint={2511.16652},
      archivePrefix={arXiv},
      primaryClass={cs.LG},
}

@article{Robbins51,
	Author = {Herbert Robbins and Sutton Monro},
	Journal = {The Annals of Mathematical Statistics},
	Publisher = {Institute of Mathematical Statistics},
	Title = {{A Stochastic Approximation Method}},
	Year = {1951}}

@misc{Mnih2013,
      title={Playing Atari with Deep Reinforcement Learning}, 
      author={Volodymyr Mnih and Koray Kavukcuoglu and David Silver and Alex Graves and Ioannis Antonoglou and Daan Wierstra and Martin Riedmiller},
      year={2013},
      eprint={1312.5602},
      archivePrefix={arXiv},
}

@article{Watkins1992,
author = {Watkins, Christopher J. C. H. and Dayan, Peter},
title = {Q-learning},
year = {1992},
journal = {Machine learning},
}

@article{Sutton1988,
    author = {Sutton, Richard},
    year = {1988},
    month = {08},
    title = {Learning to Predict by the Method of Temporal Differences},
    journal = {Machine Learning},
}

@InProceedings{Fellows2024,
  title = 	 {{B}ayesian Exploration Networks},
  author =       {Fellows, Mattie and Kaplowitz, Brandon Gary and Schroeder De Witt, Christian and Whiteson, Shimon},
  booktitle = 	 {Proceedings of the 41st International Conference on Machine Learning},
  year = 	 {2024},
  series = 	 {Proceedings of Machine Learning Research},
}

@misc{Fedus2019,
  title = {Hyperbolic Discounting and Learning over Multiple Horizons},
  author = {Fedus, William and Gelada, Carles and Bengio, Yoshua and Bellemare, Marc G. and Larochelle, Hugo},
  year = {2019},
  eprint = {1902.06865},
  archivePrefix = {arXiv},
}

@inproceedings{Sutton2011,
author = {Sutton, Richard S. and Modayil, Joseph and Delp, Michael and Degris, Thomas and Pilarski, Patrick M. and White, Adam and Precup, Doina},
title = {Horde: a scalable real-time architecture for learning knowledge from unsupervised sensorimotor interaction},
year = {2011},
booktitle = {The 10th International Conference on Autonomous Agents and Multiagent Systems}
}

@article{Sutton1999,
  title = {Between MDPs and semi-MDPs: A framework for temporal abstraction in reinforcement learning},
  author = {Sutton, Richard S. and Precup, Doina and Singh, Satinder},
  journal = {Artificial Intelligence},
  year = {1999},
}

@article{Gallici2025,
    title={Simplifying Deep Temporal Difference Learning},
    author={Matteo Gallici and Mattie Fellows and Benjamin Ellis
     and Bartomeu Pou and Ivan Masmitja and Jakob Nicolaus Foerster
      and Mario Martin},
    year={2025}, 
    journal={The International Conference on Learning Representations},
}

@misc{Gymnax2022,
  author = {Robert Tjarko Lange},
  title = {{gymnax}: A {JAX}-based Reinforcement Learning Environment Library},
  url = {http://github.com/RobertTLange/gymnax},
  version = {0.0.4},
  year = {2022},
}

@misc{Towers2023,
  title        = {Gymnasium: A Standard Interface for Reinforcement Learning Environments},
  author       = {Towers, Mark and Kwiatkowski, Ariel and Terry, Jordan},
  year         = {2023},
  note         = {Farama Foundation},
  howpublished = {\url{https://gymnasium.farama.org/}}
}

@inproceedings{Adamczyk2025,
  title = {Inferring Transition Dynamics from Value Functions},
  author = {Adamczyk, Jacob},
  booktitle = {The {{Thirteenth International Conference}} on {{Learning Representations}}},
  year = {2025},
  eprint = {2501.09081},
  archiveprefix = {arXiv}
}

@article{Wilkie1996,
  author = {Wilkie, A. J.},
  title = {Model completeness results for expansions of the ordered field of real numbers by restricted Pfaffian functions and the exponential function.},
  journal = {Journal of the American Mathematical Society},
  year = {1996}
}

@misc{Turner2022,
  title = {On Avoiding Power-Seeking by Artificial Intelligence},
  author = {Turner, Alexander Matt},
  year = {2022},
  eprint = {2206.11831},
  archiveprefix = {arXiv}
}

@inproceedings{Liu2025,
  title={A Single Goal is All You Need: Skills and Exploration Emerge from Contrastive {RL} without Rewards, Demonstrations, or Subgoals},
  author={Liu, Grace and Tang, Michael and Eysenbach, Benjamin},
  booktitle={The Thirteenth International Conference on Learning Representations},
  year={2025},
}

@inproceedings{Schaul2015,
  title = {Universal {{Value Function Approximators}}},
  booktitle = {Proceedings of the 32nd {{International Conference}} on {{Machine Learning}}},
  author = {Schaul, Tom and Horgan, Daniel and Gregor, Karol and Silver, David},
  year = 2015,
  publisher = {PMLR}
}

@inproceedings{Park2024,
  title = {{{OGBench}}: {{Benchmarking Offline Goal-Conditioned RL}}},
  shorttitle = {{{OGBench}}},
  booktitle = {The {{Thirteenth International Conference}} on {{Learning Representations}}},
  author = {Park, Seohong and Frans, Kevin and Eysenbach, Benjamin and Levine, Sergey},
  year = 2024
}

@inproceedings{Bush2025,
  title={Interpreting Emergent Planning in Model-Free Reinforcement Learning},
  author={Thomas Bush and Stephen Chung and Usman Anwar and Adri{\`a} Garriga-Alonso and David Krueger},
  booktitle={The Thirteenth International Conference on Learning Representations},
  year={2025},
}

@inproceedings{Touati2021,
author = {Touati, Ahmed and Ollivier, Yann},
title = {Learning one representation to optimize all rewards},
year = {2021},
booktitle = {Proceedings of the 35th International Conference on Neural Information Processing Systems},
}

@inproceedings{Bortkiewicz2025,
    author    = {Bortkiewicz, Micha\l{} and Pa\l{}ucki, W\l{}adek and Myers, Vivek and
                 Dziarmaga, Tadeusz and Arczewski, Tomasz and Kuci\'{n}ski, \L{}ukasz and
                 Eysenbach, Benjamin},
    booktitle = {{International Conference} on {Learning Representations}},
    title     = {{Accelerating Goal-Conditioned RL Algorithms} and {Research}},
    year      = {2025},
}

@article{Dayan1993,
  title = {Improving Generalization for Temporal Difference Learning: {{The}} Successor Representation},
  author = {Dayan, Peter},
  year = {1993},
  journal = {Neural Computation},
}

@inproceedings{Barreto2017,
  title = {Successor Features for Transfer in Reinforcement Learning},
  booktitle = {Advances in {{Neural Information Processing Systems}}},
  author = {Barreto, Andr{\'e} and Dabney, Will and Munos, R{\'e}mi and Hunt, Jonathan J. and Schaul, Tom and {van Hasselt}, Hado and Silver, David},
  year = {2017},
}

@misc{Borsa2018,
  title = {Universal {{Successor Features Approximators}}},
  author = {Borsa, Diana and Barreto, Andr{\'e} and Quan, John and Mankowitz, Daniel and Munos, R{\'e}mi and van Hasselt, Hado and Silver, David and Schaul, Tom},
  year = {2018},
  eprint = {1812.07626},
  archiveprefix = {arXiv}
}

@misc{Blier2021,
  title = {Learning {{Successor States}} and {{Goal-Dependent Values}}: {{A Mathematical Viewpoint}}},
  author = {Blier, L{\'e}onard and Tallec, Corentin and Ollivier, Yann},
  year = {2021},
  eprint = {2101.07123},
  archiveprefix = {arXiv}
}

@article{Lehnert2020,
  title = {Successor Features Combine Elements of Model-Free and Model-Based Reinforcement Learning},
  author = {Lehnert, Lucas and Littman, Michael L.},
  year = {2020},
  journal = {Journal of Machine Learning Research},
}

@inproceedings{Andrychowicz2017,
  title = {Hindsight Experience Replay},
  author = {Andrychowicz, Marcin and Wolski, Filip and Ray, Alex and Schneider, Jonas and Fong, Rachel and Welinder, Peter and McGrew, Bob and Tobin, Josh and Abbeel, Pieter and Zaremba, Wojciech},
  booktitle = {Advances in {{Neural Information Processing Systems}}},
  year = {2017}
}

@inproceedings{Grimm2020,
  title = {The Value Equivalence Principle for Model-Based Reinforcement Learning},
  author = {Grimm, Christopher and Barreto, Andr{\'e} and Singh, Satinder and Silver, David},
  booktitle = {Advances in {{Neural Information Processing Systems}}},
  year = {2020}
}

@inproceedings{Grimm2021,
  title = {Proper Value Equivalence},
  author = {Grimm, Christopher and Barreto, Andr{\'e} and Farquhar, Gregory and Silver, David and Singh, Satinder},
  booktitle = {Advances in {{Neural Information Processing Systems}}},
  year = {2021}
}

@article{Schrittwieser2020,
  title = {Mastering {Atari}, {Go}, Chess and Shogi by Planning with a Learned Model},
  author = {Schrittwieser, Julian and Antonoglou, Ioannis and Hubert, Thomas and Simonyan, Karen and Sifre, Laurent and Schmitt, Simon and Guez, Arthur and Lockhart, Edward and Hassabis, Demis and Graepel, Thore and Lillicrap, Timothy and Silver, David},
  journal = {Nature},
  year = {2020}
}

@InProceedings{Riedmiller2005,
  title = {Neural Fitted Q Iteration -- First Experiences with a Data Efficient Neural Reinforcement Learning Method},
  author = {Riedmiller, Martin},
  booktitle = {Machine Learning: ECML 2005},
  year = {2005}
}

@inproceedings{Farahmand2018,
  title = {Iterative Value-Aware Model Learning},
  author = {Farahmand, Amir-massoud},
  booktitle = {Advances in {{Neural Information Processing Systems}}},
  year = {2018}
}

@inproceedings{Lambert2020,
  title = {Objective Mismatch in Model-based Reinforcement Learning},
  author = {Lambert, Nathan and Amos, Brandon and Yadan, Omry and Calandra, Roberto},
  booktitle = {Proceedings of the 2nd {{Conference}} on {{Learning}} for {{Dynamics}} and {{Control}}},
  series = {Proceedings of Machine Learning Research},
  year = {2020},
  publisher = {PMLR}
}

@inproceedings{Freed2024,
  title = {Unifying Model-Based and Model-Free Reinforcement Learning with Equivalent Policy Sets},
  author = {Freed, Benjamin and Wei, Thomas and Calandra, Roberto and Schneider, Jeff and Choset, Howie},
  booktitle = {Reinforcement {{Learning Conference}}},
  year = {2024}
}

@inproceedings{Cao2021,
  title = {Identifiability in Inverse Reinforcement Learning},
  author = {Cao, Haoyang and Cohen, Samuel N.},
  booktitle = {Advances in {{Neural Information Processing Systems}}},
  year = {2021},
  eprint = {2106.03498},
  archiveprefix = {arXiv}
}

@inproceedings{Kingma2014,
  title = {Auto-Encoding Variational Bayes},
  author = {Kingma, Diederik P. and Welling, Max},
  booktitle = {International Conference on Learning Representations},
  year = {2014}
}

@inproceedings{Rezende2014,
author = {Rezende, Danilo Jimenez and Mohamed, Shakir and Wierstra, Daan},
title = {Stochastic backpropagation and approximate inference in deep generative models},
year = {2014},
booktitle = {Proceedings of the 31st International Conference on International Conference on Machine Learning - Volume 32},
}

@misc{bradbury_jax_2018,
    title = {{JAX}: composable transformations of {Python}+{NumPy} programs},
    url = {http://github.com/jax-ml/jax},
    author = {Bradbury, James and Frostig, Roy and Hawkins, Peter and Johnson, Matthew James and Leary, Chris and Maclaurin, Dougal and Necula, George and Paszke, Adam and VanderPlas, Jake and Wanderman-Milne, Skye and Zhang, Qiao},
    year = {2018},
}

@inproceedings{Kaelbling1993LearningTA,
  title={Learning to Achieve Goals},
  author={Leslie Pack Kaelbling},
  booktitle={International Joint Conference on Artificial Intelligence},
  year={1993},
}

@inproceedings{
wang2026,
title={1000 Layer Networks for Self-Supervised {RL}: Scaling Depth Can Enable New Goal-Reaching Capabilities},
author={Kevin Wang and Ishaan Javali and Micha{\l} Bortkiewicz and Tomasz Trzcinski and Benjamin Eysenbach},
booktitle={The Thirty-ninth Annual Conference on Neural Information Processing Systems},
year={2026},
}

@misc{ghugare_role_2026,
    title = {On the {Role} of {Iterative} {Computation} in {Reinforcement} {Learning}},
    publisher = {arXiv},
    author = {Ghugare, Raj and Bortkiewicz, Michał and Ziarko, Alicja and Eysenbach, Benjamin},
    month = feb,
    year = {2026},
    note = {arXiv:2602.05999 [cs]},
}

@article{richardson_lir_26,
  title={Local Inconsistency Resolution: The Interplay between Attention and Control in Probabilistic Models},
  author={Richardson, Oliver E and Samiei, Mandana and Shakerinava, Mehran and Viviano, Joseph D and Kabid, Abdessamad El and Parviz, Ali and Bengio, Yoshua},
  journal={Proceedings of the 29th International Conference on Artificial Intelligence and Statistics},
  year={2026}
}

@book{Folland1999,
  author    = {Folland, Gerald B.},
  title     = {Real Analysis: Modern Techniques and Their Applications},
  year      = {1999},
  publisher = {John Wiley \& Sons},
  series    = {Pure and Applied Mathematics: A Wiley-Interscience Series of Texts, Monographs and Tracts},
}

@book{SteinWeiss1971,
  author    = {Stein, Elias M. and Weiss, Guido},
  title     = {Introduction to {F}ourier Analysis on {E}uclidean Spaces},
  year      = {1971},
  publisher = {Princeton University Press},
  series    = {Princeton Mathematical Series},
}

@book{KrantzParks2002,
  author    = {Krantz, Steven G. and Parks, Harold R.},
  title     = {A Primer of Real Analytic Functions},
  year      = {2002},
  edition   = {2nd},
  publisher = {Birkh\"auser Boston},
}

@book{Lee2013,
  author    = {Lee, John M.},
  title     = {Introduction to Smooth Manifolds},
  edition   = {2nd},
  publisher = {Springer},
  year      = {2013},
}

@misc{schulman_proximal_2017,
    title = {Proximal {Policy} {Optimization} {Algorithms}},
    language = {en},
    publisher = {arXiv},
    author = {Schulman, John and Wolski, Filip and Dhariwal, Prafulla and Radford, Alec and Klimov, Oleg},
    month = aug,
    year = {2017},
}

@book{SuttonBarto2018Reinforcement,
    title = {Reinforcement learning: {An} introduction},
    publisher = {A Bradford Book},
    author = {Sutton, Richard and Barto, Andrew},
    year = {2018},
}
}

\appendix
\crefalias{section}{appendix}
\crefalias{subsection}{appendix}
\crefalias{subsubsection}{appendix}
\clearpage

\section{\texorpdfstring{Proofs for $P$-Learning}{Proofs for P-Learning}}
\label{app:p_der}

\subsection{\texorpdfstring{Derivation of $P$-Learning}{Derivation of P-Learning}}

We provide pseudocode for $P$-learning in \cref{alg:p-learning} below, with batch size $1$ for simplicity. To derive the update, recall the definition of the Bellman residual
\[ \cL(\phi) \eq \| \cT^\pi_\phi(Q) - Q \|_d^2 = \frac{1}{2} \E_{(g,s,a)\sim d} \left[ \left( \cT^\pi_\phi(Q)(s,a,g) - Q(s,a,g) \right)^2 \right] \, = \frac12 \E_d[ \delta_\phi^2 ] \,, \]
where $\delta_\phi(s,a,g) = \cT^\pi_\phi(Q)(s,a,g) - Q(s,a,g)$ is the TD error. 
By the chain rule, and since the expectation does not depend on $\phi$,  we have
\[ \nabla_\phi \cL(\phi) = \E_{(g,s,a)\sim d}\left[\delta_\phi(s,a,g) \nabla_\phi \cT^\pi_\phi(Q)(s,a,g) \right] \,. \]
We now assume three standard regularity conditions for the score-function gradient estimator \citep{Mohamed2020}, namely: (i) $P_\phi$ is continuously differentiable in $\phi$, (ii) $P_\phi$ has full support, and (iii) there is an integrable function $g : \cS \to \R$ with $\sup_\phi \|\nabla_\phi P_\phi(s' \mid s, a)\|_1 \leq g(s')$ for all $s, a, s'$. Then the gradient can be estimated via the log-derivative trick \citep{Mohamed2020}:
\begin{align*}
    \nabla_\phi \cL(\phi) &=\E_{(g,s,a)\sim d}\left[\delta_\phi(s,a,g) \nabla_\phi \E_{s' \sim P_\phi(s,a), a' \sim \pi(\cdot \mid s',g)}\left[r(s',g) + \gamma Q(s',a',g)\right] \right] \\
    &= \E_{(g,s,a)\sim d,s' \sim P_\phi( s,a), a' \sim \pi(s',g)}\left[\delta_\phi(s,a,g) (r(s',g) + \gamma Q(s',a',g)) \nabla_\phi \log P_\phi(s' \mid s, a)\right] \,.
\end{align*}
To reduce variance and make the per-sample direction of this update explicit, we subtract the baseline $Q(s,a,g)$ to get a per-sample TD error 
\[ \hat\delta_\phi(s, a, g, s', a') \eq r(s',g) + \gamma Q(s',a',g) - Q(s,a,g) \,, \]
and the update, as claimed in the main text, simplifies to
\begin{equation}
    \nabla_\phi \cL(\phi) = \E_{(g,s,a)\sim d, s' \sim P_\phi(s,a), a' \sim \pi(s',g)}\bigl[\delta_\phi(s, a, g) \hat\delta_\phi(s, a, g, s', a') \nabla_\phi \log P_\phi(s' \mid s,a)\bigr] \,.
       \label{eqn:gradexpr}
\end{equation}
Note that the baseline preserves unbiasedness: $Q(s, a, g)$ is constant under the inner sampling of $(s', a')$, so it factors out and multiplies the expected score $\E_{s' \sim P_\phi(s,a)}[\nabla_\phi \log P_\phi(s' \mid s, a)]$, which is zero because $\int_{\cS} P_\phi(s' \mid s, a)ds' = 1$. Now the TD error $\delta_\phi(s,a,g)$ does not depend on the variables $(s',a')$ in the outer expectation, so we can factor it out, yielding:
\[
    \nabla_\phi \cL(\phi) = \E_{(g,s,a)\sim d} \bigg[ \delta_\phi(s,a,g) \E_{s' \sim P_\phi(s,a),\, a' \sim \pi(s',g)} \Big[ \hat\delta_\phi \nabla_\phi \log P_\phi(s' \mid s, a) \Big] \bigg] \,.
\]
The gradient is a product of two expectations over $(s', a')$, so an unbiased estimator requires two independent samples: $(s'_1, a'_1)$ to form an unbiased estimate $\hat\delta^{1}$ of the outer factor $\delta_\phi$, and an independent $(s'_2, a'_2)$ to estimate the inner expectation. Independence ensures unbiased estimation:
\begin{align*}
    \nabla_\phi \cL(\phi) &= \E_{(g,s,a)\sim d}\bigg[ \delta_\phi(s,a,g) \E_{s' \sim P_\phi(s,a), a' \sim \pi(s',g)}\left[\hat\delta_\phi \nabla_\phi \log P_\phi(s' \mid s,a)\right] \bigg]  \\
    &= \E_{(g,s,a)\sim d, s_{1,2} \sim P_\phi(s,a), a_{1,2} \sim \pi(s_{1,2}, g)}\Big[\hat\delta^{1} \hat\delta^{2} \nabla_\phi \log P_\phi(s_2 \mid s,a)\Big] \,.
\end{align*}
This derivation leads to the pseudocode for $P$-learning in \cref{alg:p-learning} below, with batch size $1$ for simplicity. Crucially, both samples come from the model $P_\phi$ rather than the environment, so double sampling here is essentially free, costing only an extra forward pass through $P_\phi$ and $Q$, unlike the double-sampling obstruction in TD learning which requires double-sampling in the environment \citep{Baird1995}.

\begin{algorithm}[ht]
    \caption{$P$-Learning.}
    \label{alg:p-learning}
    {\linespread{1.2}\selectfont
    \begin{algorithmic}[1]
    \Require Parametric family $P_\phi$, $Q$-function $Q$, policy $\pi$, reward $r$, discount $\gamma$, sampling distribution $d$ over $(g, s, a)$, step sizes $\{\alpha_n\}$, initial $\phi_0$.
    \For{$n = 0, 1, \ldots,$}
        \State Sample $(g, s, a) \sim d$
        \State Sample $s'_1, s'_2 \sim P_{\phi_n}(\cdot \mid s, a)$, $a'_1 \sim \pi(\cdot \mid s'_1, g)$, $a'_2 \sim \pi(\cdot \mid s'_2, g)$
        \State $\delta^{i} \gets r(s'_i, g) + \gamma\, Q(s'_i, a'_i, g) - Q(s, a, g)$ for $i = 1, 2$
        \State $\hat g_n \gets \delta^{1} \delta^{2} \nabla_\phi \log P_\phi(s'_2 \mid s, a)\big|_{\phi=\phi_n}$
        \State $\phi_{n+1} \gets \phi_n - \alpha_n \hat g_n$
    \EndFor
    \State \Return Extracted world model $P_{\phi_n}$.
    \end{algorithmic}
    \par}
\end{algorithm}

\paragraph{Deterministic case.}
When dynamics are known to be deterministic and the model is parameterised as a successor function $P_\phi : \cS \times \cA \to \cS$, the inner expectation over $s'$ in $\cT^\pi_\phi(Q)$ collapses to
\[
    \cT^\pi_\phi(Q)(s,a,g) = r(P_\phi(s,a), g) + \gamma V_g(P_\phi(s,a)) \,, \qquad V_g(s') = \E_{a' \sim \pi(s', g)}[Q(s', a', g)] \,.
\]
Hence $\delta_\phi(s,a,g)$ is computable from a single forward pass, with no Monte-Carlo estimate over $s'$; the only remaining randomness is over $a' \sim \pi(P_\phi(s,a), g)$. The gradient
\[
    \nabla_\phi \cL(\phi) = \delta_\phi(s,a,g) \nabla_\phi \big[ r(P_\phi(s,a), g) + \gamma\, V_g(P_\phi(s,a)) \big]
\]
obtained by chain rule (reparameterisation) through $P_\phi$ is then unbiased from a single sample. This requires $r$ and $Q$ to be differentiable almost everywhere in their first argument, as discussed in \cref{sec:method}; otherwise, zero-shot methods like evolutionary strategies \citep{Rechenberg78,sarkar2026} are appropriate.

\subsection{\texorpdfstring{Proof of \cref{thm:inverse-bellman-tabular}}{Proof of Theorem (inverse Bellman, tabular)}}\label{sec:proofs:inverse-bellman}

\label{proof:inverse-bellman-tabular}
\recall{theorem:inverse-bellman-tabular}

\begin{lproof}
    Fix $s, a$ and write $M \eq M^\pi$ and $Q \eq Q(s,a)$, $P_t \eq P_t(s,a)$ for brevity. The iteration is
    \[
        P_{t+1} = P_t - \alpha M^\top (M P_t - Q) \,.
    \]
    Now let $r = \rank(M)$ and write the SVD $M = U \Sigma V^\top$ where $U \in \R^{L \times r}$, $V \in \R^{n \times r}$ have orthonormal columns, and $\Sigma = \diag(\sigma_1, \ldots, \sigma_r) \succ 0$. We extend $V$ to an orthonormal basis $[V\ V_\perp]$ of $\R^n$ with $V_\perp \in \R^{n \times (n-r)}$, so that $V_\perp^\top V_\perp = I_{n-r}$ and $V^\top V_\perp = 0$. In particular, $[V ; V_\perp]$ is an orthogonal $n \times n$ matrix with $V V^\top + V_\perp V_\perp^\top = I_n$, and we can decompose each iterate as
    \[ P_t = (VV^\top + V_\perp V_\perp^\top)P_t = V a_t + V_\perp b_t \,, \]
    where $a_t \eq V^\top P_t \in \R^r$ and $b_t \eq V_\perp^\top P_t \in \R^{n-r}$. Using orthogonality $(V_\perp^\top V) = 0$,
    \[
        b_{t+1} = V_\perp^\top P_{t+1} = b_t - \alpha V_\perp^\top M^\top (M P_t - Q) = b_t - \alpha (V_\perp^\top V) \Sigma U^\top (M P_t - Q) = b_t \,,
    \]
    so $b_t = b_0 = V_\perp^\top P_0$ for all $t$. We now prove by induction that
    \[ a_t = (I - \alpha \Sigma^2)^t a_0 + \alpha \sum_{k=0}^{t-1} (I - \alpha \Sigma^2)^k \Sigma U^T Q \]
    for all $t \geq 0$. The base case $t=0$ holds trivially. Assuming the claim holds for fixed $t \geq 0$,
    \begin{align*}
        a_{t+1} &= a_t - \alpha \Sigma U^\top (U \Sigma V^\top P_t - Q)
        = (I - \alpha \Sigma^2) a_t + \alpha \Sigma U^\top Q \\
        &= (I - \alpha \Sigma^2) \big[ (I - \alpha \Sigma^2)^t a_0 + \alpha \sum_{k=0}^{t-1} (I - \alpha \Sigma^2)^k \Sigma U^T Q \big] + \alpha \Sigma U^\top Q \\
        &= (I - \alpha \Sigma^2)^{t+1} a_0 + \alpha \sum_{k=0}^{t} (I - \alpha \Sigma^2)^k \Sigma U^T Q \,,
    \end{align*}
    so the induction is complete. In particular, since $\Sigma = \diag(\sigma_1, \ldots, \sigma_r) \succ 0$, we have $\abs{1-\alpha \sigma_i^2} < 1$ for all $0 < \alpha < 2/\sigma_{\max}^2$ and hence $(I - \alpha \Sigma^2)^t \to 0$ as $t \to \infty$, thus
    \begin{align*}
        \lim_{t \to \infty} a_t &= \lim_{t \to \infty} \alpha \sum_{k=0}^{t-1} (I - \alpha \Sigma^2)^k \Sigma U^T Q \\
        &= \alpha \big[I - (I - \alpha \Sigma^2)\big]^{-1} \Sigma U^T Q
        = \Sigma^{-1} U^\top Q \,.
    \end{align*}
    Combining the expressions for $a_t$ and $b_t$, we conclude that
    \[
        P_\infty \coloneqq \lim_{k \to \infty} P_t = V \Sigma^{-1} U^\top Q + V_\perp V_\perp^\top P_0
    \]
    for all $0 < \alpha < 2/\sigma_{\max}^2$. Now $M^+ = V \Sigma^{-1} U^\top$ is the Moore--Penrose pseudo-inverse of $M$, and
    \[
        M^+ M = V \Sigma^{-1} U^\top U \Sigma V^\top = V V^\top
        \quad \implies \quad
        I - M^+ M = V_\perp V_\perp^\top \,,
    \]
    so we obtain $P_\infty = M^+ Q + (I - M^+ M) P_0$ as claimed.
    

    \textbf{Full column-rank.} If $\rank(M) = n$, we have $I - M^+ M = 0$, and $P_\infty = M^+ Q$ independent of $P_0$. If in addition the system $M P = Q$ is consistent, e.g. when $Q = M P$ for exact tabular $Q$-values, then the true transition kernel $P$ is the unique solution and $P_\infty = M^+ Q = P$.
\end{lproof}

\section{Value Equivalence and Pseudo-World Models}\label{sec:duality}

\subsection{Intuition: Breaking Value Equivalence}\label{sec:breaking-equivalence}

We provide intuition for how exactly spanning goals break value equivalence using the notion of ``test functions'' to determine unknown distributions. Recalling the Bellman operator
\begin{equation*}
    \cT_P(Q)(s, a, g) = \E_{s' \sim P(s, a)} \left[ M_g(s') \right] \,, \quad \text{where} \quad M_g(s') \eq r_g(s') + \gamma V_g(s') \,,
\end{equation*}
notice that the Bellman equation reduces to $Q(s,a,g) = \E_{s' \sim P(s, a)} \left[ M_g(s') \right]$, which can be viewed as imposing $|\cG|$ constraints on the expectations of \textit{known} ``test functions'' or ``Bellman probes'' $M_g$ under the \textit{unknown} distribution $P(s, a)$. If the agent is trained on a single goal $g$, it is natural to expect that there will be many world models $\hat{P}$ that have the same expectation under the single test function, i.e. such that $\E_{s' \sim \hat{P}(s, a)} \left[ M_g(s') \right] = Q(s,a,g)$, inducing value equivalence.

However, probability distributions can be determined by expectations over multiple test functions: beginning with the finite setting, a distribution $p \in \Delta^n$ is uniquely determined by the first $n-1$ moments of any injective random variable (`test function') $X : [n] \to \R$. For example, in the 1-dimensional case $p = (p_1, 1-p_1)$, if we know the first moment $\E[X] = p_1 X(1) + (1-p_1) X(2)$, we can invert the system to determine $p_1 = (\E[X] - X(2))/(X(1) - X(2))$ uniquely. For general $p \in \Delta^n$, the first $n$ moments induce a linear system of equations known as a \textit{Vandermonde system}, which is invertible if and only if $X$ is injective -- meaning that $p$ can be recovered uniquely.

However, the Bellman probes are not simple functions like monomials, and it is not clear that they will similarly induce an invertible system. For continuous MDPs \cref{sec:continuous}, the picture is even more involved, and invertibility generalises to a set of test functions being \textit{measure determining}, meaning that distinct Borel measures have distinct expectations.

\subsection{World Models and Pseudo-World Models}\label{sec:proof-simplex}

We introduce two spaces of \textbf{stochastic} world models:
\[ \cP \eq \{ P : \cS \times \cA \to \Sigma(\cS) \} \quad \text{and} \quad \cP_\Delta \eq \{ P : \cS \times \cA \to \Delta(\cS) \} \subset \cP \,,\]
where $\Sigma(\cS)$ denotes the space of finite \textit{signed} measures over $\cS$. Elements of $\cP_\Delta$ are valid \textit{world models}, while elements of $\cP \setminus \cP_\Delta$ are \textit{pseudo-world models}: signed measures that may still satisfy the Bellman equation $\cT_P(Q) = Q$ but do not correspond to valid probabilistic dynamics. In particular, the $P$-learning iteration $P_{t+1}(s,a) = P_t(s,a) - \alpha M^\top (M P_t(s,a) - Q(s,a))$ does not preserve positivity or normalisation, so the limit $P_\infty = M^+ Q + (I - M^+M) P_0$ generally lies in $\cP$.

However, if the conditions of \cref{thm:finite}(a) are satisfied, then there exists a true kernel $P \in \cP_\Delta$ with $Q = M P$, and $M$ has full rank, so $P_\infty = M^+ M P = P \in \cP_\Delta$, so the iteration converges to a true world model. Moreover, in realistic settings with function approximation, the world model is typically parameterised as a neural network $P_\phi : \cS \times \cA \to \cP$ with a softmax output layer, so $P$-learning is constrained to the simplex by construction. Finally, in \textbf{deterministic} MDPs, a world model is defined as a function $P : \cS \times \cA \to \cS$, so the distinction collapses, and either column-matching or approximate methods always return a valid world model. If none of these conditions hold, the pseudo-world caveat can be mitigated in full generality by a projected variant of $P$-learning, given by
\[
    P_{t+1}(s,a) = \Pi_{\Delta}\bigl(P_t(s,a) - \alpha M^\top (M P_t(s,a) - Q(s,a))\bigr) \,,
\]
where $\Pi_{\Delta}$ is the Euclidean projection onto the simplex. Since projection onto a closed convex set is non-expansive, the iteration still converges, and the limit is $P_\infty^{\Delta} = \argmin_{P \in \cP_\Delta} \|MP - Q\|^2$. This is not generally equal to post-hoc projection $\Pi_{\Delta}(P_\infty)$ of the unconstrained limit, which ignores the geometry of the loss landscape and can return a strictly worse fit than $P_\infty^{\Delta}$ when $M$ is poorly conditioned. In the approximate setting of \cref{thm:finite}(b), where $M$ is full-rank, both post-hoc projection and projected $P$-learning inherit the same asymptotic upper bound.
\section{Proofs for Finite MDPs}\label{sec:proofs:finite}

In this appendix, we formalise and prove our results for MDPs with finite state spaces: \cref{sec:proof-deterministic,sec:proof-concrete-rewards} for deterministic MDPs, \cref{sec:proof-finite-stoch} for stochastic MDPs, and \cref{sec:proof-local} for local MDPs. We first state and prove a simple lemma that underpins many of our results.

\begin{lemma}\label{lem:resolvent}
    Fix any finite (goal-augmented) MDP $\cM = (\cS, \cA, \cG, P, \mu, r, \gamma)$, and recall the definition
    \[ M(s', g) \eq r(s', g) + \gamma V^\pi(s', g) \]
    from \cref{eq:bellman-general}. Then the resolvent identity
    \begin{equation}\label{eq:resolvent}
        M_l = R_l\big(I - \gamma (P^{\pi_l})^\tr\big)^{-1}
    \end{equation}
    holds for all $g_l \in \cG$, where $M_{lk} \eq M(s_k, g_l)$, $R_{lk} \eq r(s_k, g_l)$ and the policy-induced kernel $P^{\pi_l}(s_k, s_{k'}) \eq \sum_j \pi(a_j \mid s_k, g_l) P(s_{k'} \mid s_k, a_j)$ are viewed as matrices.
\end{lemma}

\begin{lproof}
    We use $n \eq |\cS|$, $m \eq |\cA|$, $L \eq |\cG|$ throughout, viewing reward functions $R_{lk} \eq r(s_k, g_l) \in \R^{|\cG| \times |\cS|}$ and similarly $M, V$ as matrices. Rewriting Equation \ref{eq:bellman-linear} as
    \[
        M^\pi_{lk} = R_{lk} + \gamma V_{lk}^{\pi_l} \,,
    \]
    and dropping the superscript $\pi$ for simplicity, we define the policy-induced kernel $P^{\pi_l}_{kk'} \eq \sum_{j} \pi_{lkj} P_{kjk'}$ for each $l$, and expand the Bellman equation
    \begin{align}
        M_{lk} &= R_{lk} + \gamma \sum_{j, k'} \pi_{lkj} P_{kjk'} M_{lk'} \label{eq:M_expansion} \\
        &= R_{lk} + \gamma \sum_{k'} P^{\pi_l}_{kk'} M_{lk'} \notag \,.
    \end{align}
    Defining $\bP^{\pi_l} = (P^{\pi_l})^\tr$, we rewrite this equation as
    \[
        \sum_{k'} M_{lk'} \left( \delta_{k'k} - \gamma \bP^{\pi_l}_{k'k} \right) = R_{lk} \,.
    \]
    The quantity in the brackets is not independent from $l$, so this is not a matrix product. However, the $l$th row of $M$ can still be written as a matrix product:
    \[
        M_l = R_l\left(I-\gamma \bP^{\pi_l}\right)^{-1} \,,
    \]
    noting that $\bP^{\pi_l}$ is column-stochastic matrix for each $l$, so $(I - \gamma \bP^{\pi_l})$ is invertible for each $l$. 
\end{lproof}

\subsection{Proof of \texorpdfstring{\cref{thm:deterministic} (Deterministic MDPs)}{Theorem (Deterministic)}}\label{sec:proof-deterministic}

\label{proof:deterministic}
\recall{theorem:deterministic}

We split this result into three parts: \cref{prop:finite-deterministic-a} for part (a), which includes the corollary mentioned below \cref{thm:deterministic}, and \cref{prop:finite-deterministic-upper,prop:finite-deterministic-lower} for part (b). 

\begin{linked}{proposition}{finite-deterministic-a}\label{prop:finite-deterministic-a}
    Fix any deterministic finite MDP, goal set $|\cG| \geq 1$ and policy $\pi$. Then the set of reward functions $r : \cS \times \cG \to \R$ for which $M^\pi$ is column-injective, and thus $\exists! P(Q^\pi)$, has full Lebesgue measure. In particular, $|\cG| = 1$, fix any $r_0 : \cS \to \R$ and any random vector $\xi \in \R^{|\cS|}$ whose distribution is absolutely continuous with respect to Lebesgue measure. Then the perturbed reward $r = r_0 + \xi$ satisfies $\exists!P(Q^\pi)$ almost surely, viewing $r \in \R^{|\cS|}$ as a vector.
\end{linked}

\begin{lproof}
    First recall from \cref{sec:finite} that for deterministic finite MDPs, the Bellman equation reduces to $M(P(s,a), g) = Q(s,a,g)$ for all $g \in \cG$. Writing $M(s') = M(s', \cdot)$, this implies that the true transition kernel satisfies
    \[ P(s,a) = \argmin_{s'} \norm{M(s') - Q(s,a)}_1 \,. \]
    In particular, a sufficient condition for this argmin to be unique, and thus $\exists! P(Q^\pi)$, is for $M$ to be column-injective, i.e. $M(s_k) \neq M(s_{k'})$ for all $k \neq k'$. Now the resolvent identity (\cref{eq:resolvent})
    \[
        M_l = R_l\left(I-\gamma \bP^{\pi_l}\right)^{-1}
    \]
    implies that the map $F : \R^{L \times n} \to \R^{L \times n}$ given by $F(R) = M$ is linear and invertible. Moreover, for any $k \neq k'$, the set $\{M \in \R^{L \times n} : M_{\cdot, k} = M_{\cdot, k'}\}$ is a proper subspace of $\R^{L \times n}$, so it has Lebesgue measure zero. Taking a finite union over the $\binom{n}{2}$ pairs $k \neq k'$, the set
    \[
        \{ R \in \R^{L\times n} \mid \exists k \neq k' \text{ with } F(R)_{\cdot, k} = F(R)_{\cdot, k'} \} = F^{-1}\Bigl( \textstyle\bigcup_{k \neq k'} \{ M \in \R^{L \times n} \mid M_{\cdot, k} = M_{\cdot, k'} \} \Bigr)
    \]
    has Lebesgue measure zero since it is the preimage of a Lebesgue-measure-zero set under an invertible linear map. Therefore, $M$ has pairwise distinct columns for almost every $R \in \R^{L \times n}$, i.e. almost every $r : \cS \times \cG \to \R$. The first half of the proof is complete. As a corollary, taking $|\cG| = 1$ and viewing the single reward function as a vector $r \in \R^{|\cS|}$, the set
    \[ B = \{ r \in \R^{|\cS|} \mid M^\pi \ \text{not column-injective} \} \]
    has zero Lebesgue measure. Now fix any base reward $r_0 \in \R^{|\cS|}$ and any random vector $\xi \in \R^{|\cS|}$ whose distributions $\mu$ is absolutely continuous with respect to $\lambda$. By translation-invariance of Lebesgue measure, $\lambda(B - r_0) = \lambda(B) = 0$, and absolute continuity ($\mu \ll \lambda$) implies $\mu(B - r_0) = 0$, i.e. $\P_{\xi \sim \mu}(r_0 + \xi \in B) = 0$. By definition of $B$, we conclude that $M^\pi(r_0 + \xi)$ is almost surely column-injective, hence $\exists! P(Q^\pi, \pi, r_0 + \xi, \gamma)$ almost surely.
\end{lproof}

\begin{linked}{proposition}{finite-deterministic-upper}\label{prop:finite-deterministic-upper}
    Fix any deterministic finite MDP and goal set $|\cG| \geq 1$. If $Q$ is $\ep$-approximate and $M$ is column-injective with column separation $\gap$, the estimator $\hat{P}(s, a) = \operatorname{argmin}_{s'} \norm{Q(s,a) - M(s')}_1$ satisfies $\hat P = P$ for all $\ep < \gap(1 + \ga m)/2$.
\end{linked}

\begin{lproof}
    Fix a state-action pair $(s_i, a_j)$ and write $k^\star \eq P(s_i, a_j)$ for the true successor. Throughout, indices $l \in [L], k \in [n], j \in [m]$ correspond to goals, states, and actions respectively, so $\pi_{lkj} = \pi(a_j \mid s_k, g_l)$ and $Q_{lkj} = Q(s_k, a_j, g_l)$; we moreover write $Q_{ij} = Q(s_i, a_j, \cdot) \in \R^L$. It suffices to show that $Q_{ij}$ is closer to the true column $e_{k^\star}$ than to any other column $e_k$ for $k \ne k^\star$, where $e_k$ is the standard basis vector of $\R^{n}$ corresponding to state $s_k \in \cS$. The approximate matrix $M$ satisfies $M_{lk} - M^\pi_{lk} = \ga \sum_{j'} \pi_{lkj'} (Q_{lkj'} - Q^\pi_{lkj'})$, so for each column $k$:
    \[
        \norm{M_{\cdot, k} - M^\pi_{\cdot, k}}_1 \leq \ga \sum_l \sum_{j'} \pi_{lkj'} \abs{Q_{lkj'} - Q^\pi_{lkj'}} \leq \ga \sum_{j'} \norm{Q_{kj'} - Q^\pi_{kj'}}_1 \leq \ga m \ep \,.
    \]
    By the triangle inequality, the distance from $Q_{ij}$ to the true column satisfies
    \[
        \norm{Q_{ij} - M^\pi_{\cdot, k^\star}}_1 \leq \norm{Q_{ij} - Q^\pi_{ij}}_1 + \norm{M_{\cdot, k^\star} - M^\pi_{\cdot, k^\star}}_1 \leq \ep + \ga m \ep = \ep(1 + \ga m) \,.
    \]
    For any $k \neq k^\star$, using the Bellman equation $Q^\pi_{ij} = M^\pi_{\cdot, k^\star}$, we apply both standard and reverse triangle inequalities as follows:
    \begin{align*}
        \norm{Q_{ij} - M_{\cdot, k}}_1 &= \norm{Q_{ij} - Q^\pi_{ij} + M^\pi_{\cdot, k^\star} - M^\pi_{\cdot, k} + M^\pi_{\cdot, k} - M_{\cdot, k}}_1 \\
        &\geq \norm{M^\pi_{\cdot, k^\star} - M^\pi_{\cdot, k}}_1 - \big( \norm{Q_{ij} - Q^\pi_{ij}}_1 + \norm{M^\pi_{\cdot, k} - M_{\cdot, k}}_1 \big) \\
        &\geq \Delta_1 - \ep(1 + \ga m) \,,
    \end{align*}
    where $\norm{M_{\cdot, k^\star} - M_{\cdot, k}}_1 \geq \Delta_1$ by definition of column separation. Recovery succeeds when the true column is strictly closer than any wrong column, i.e. $\ep(1 + \ga m) < \Delta_1 - \ep(1 + \ga m)$, giving $\ep < \Delta_1 / 2(1 + \ga m)$. Since the choice of $(s_i, a_j)$ was arbitrary, the bound holds uniformly. When $\pi$ is unconditional, factoring out $\pi_{lkj'} = \pi_{kj'}$ sharpens the column error to $\ga\ep$, giving an improved bound $\ep < \Delta_1 / 2(1+\ga)$.
\end{lproof}

\begin{linked}{proposition}{finite-deterministic-lower}\label{prop:finite-deterministic-lower}
    For any $n \geq 3$, $m \geq 1$, $L \geq 1$, $\ga \in [0, 1)$, and $\ep > 0$, there exist two distinct deterministic transition kernels $P, P'$, a reward function $r$, and a policy $\pi$ such that the induced matrix $M^\pi$ has column separation $\Delta_1 = \ep$, and
    \begin{enumerate}[label=(\roman*)]
        \item $\norm{Q^\pi_{ij}(P) - Q^\pi_{ij}(P')}_1 \leq \ep$ for all $(i, j)$;
        \item $P(s_i, a_j) \neq P'(s_i, a_j)$ for some $(i, j)$.
    \end{enumerate}
    Consequently, any estimator mapping $\ep$-approximate $Q$-values to a deterministic world model fails to recover the true world for $\ep \geq \Delta_1/2$.
\end{linked}

\begin{lproof}
    Take $n = 3$, $m = 1$, $L = 1$. The construction below extends to any $n \geq 3, m \geq 1, L \geq 1$ by adding identity-transition states $s_4, \ldots, s_n$, self-loop actions $a_2, \ldots, a_m$, and extra goals $g_2, \ldots, g_L$ with $r_{l, k}$ identical at $s_2, s_3$ for $l \geq 2$ (contributing zero to both column separation and $Q$-differences). We define two deterministic transition kernels as follows: $P$ and $P'$ agree on the absorbing states $P(s_2, \cdot) = P'(s_2, \cdot) = s_2$ and $P(s_3, \cdot) = P'(s_3, \cdot) = s_3$, but disagree at $(s_1, a_1)$ where $P(s_1, a_1) = s_2$ and $P'(s_1, a_1) = s_3$. Take the deterministic policy $\pi(a_1 \mid s) = 1$, and for any $c \in \R$, set rewards $r(s_2) \eq (c + \ep/2)(1 - \ga)$, $r(s_3) \eq (c - \ep/2)(1 - \ga)$, with $r(s_1)$ arbitrary. Since $s_2, s_3$ self-loop under all actions, the Bellman equation gives $V^\pi(s_k) = r_k/(1 - \ga)$ at these states for both $P,P'$, so
    \[
        M^\pi_{2} = c + \ep/2 \quad \,, \quad M^\pi_{3} = c - \ep/2
    \]
    and thus $\Delta_1 \coloneqq \norm{M^\pi_{2} - M^\pi_{3}}_1 = \ep$, as claimed. Now the $Q$-values at $s_2$ or $s_3$ depend only on $M^\pi_{2}$ and $M^\pi_{3}$ respectively, which are invariant across $P, P'$, so the $Q$-differences are zero. At $(s_1, a_1)$,
    \[
        Q^\pi_{11}(P) = M^\pi_{2} \quad \,, \quad Q^\pi_{11}(P') = M^\pi_{3} \,,
    \]
    so $\norm{Q^\pi_{11}(P) - Q^\pi_{11}(P')}_1 = \ep$. At $(s_1, a_j)$ for $j \geq 2$, $\pi(a_1 \mid s_1) = 1$ gives $V^\pi_1(P) = Q^\pi_{11}(P) = M^\pi_{2}$ and similarly $V^\pi_1(P') = M^\pi_{3}$, so
    \[ Q^\pi_{j}(P) = r_{1} + \ga M^\pi_{2} \quad \,, \quad Q^\pi_{1j}(P') = r_{1} + \ga M^\pi_{3} \,, \]
    giving $\norm{Q^\pi_{j}(P) - Q^\pi_{j}(P')}_1 = \ga \ep \leq \ep$. Hence (i) holds with all $Q$-differences bounded by $\ep$, and (ii) follows from $P(s_1, a_1) = s_2 \neq s_3 = P'(s_1, a_1)$. Now let $\hQ \eq (Q^\pi(P) + Q^\pi(P'))/2$, which satisfies $\norm{\hQ_{ij} - Q^\pi_{ij}(P)}_1 \leq \ep/2$ at every $(i, j)$, with equality at $(s_1, a_1)$, and similarly for $P'$. Hence at $\ep = \Delta_1/2$, both $(P, \hQ)$ and $(P', \hQ)$ satisfy the $\ep$-feasibility constraint $\norm{\hQ - Q^\pi}_1 \leq \ep$. Any estimator $\hat P$ returns a world model that necessarily disagrees with either $P$ or $P'$ at $(s_1, a_1)$, since $P(s_1, a_1) \neq P'(s_1, a_1)$, so every estimator fails to recover the true world for $\ep \geq \Delta_1/2$.
\end{lproof}

\subsection{Proof of \texorpdfstring{\cref{prop:concrete-rewards}}{Proposition (Gaussian Rewards)}}\label{sec:proof-concrete-rewards}

While \cref{thm:deterministic} proves that ``almost every'' single-goal reward function satisfies $\exists!P(Q^\pi)$, this does not necessarily hold for general parametric families. Below we prove that the Gaussian family is particularly well-behaved. Note that $\pi : \cS \to \Delta(\cA)$ is treated as fixed (not dependent on $g$), but the same conclusion extends to the optimal policy $\pi_g : \cS \to \cA$ with respect to the varying goal $g$ (cf. proof of the stochastic generalisation, \cref{prop:concrete-rewards-stochastic}).

\label{proof:concrete-rewards}
\recall{proposition:concrete-rewards}

\begin{lproof}
    Viewing single-goal reward functions $r : \cS \to \R$ as vectors $r \in \R^{|\cS|}$, the Gaussian family $\{r_{g} \in \R^{|\cS|} \mid g \in \R^d\}$ is a $d$-dimensional submanifold of $\R^{|\cS|}$ which can have zero Lebesgue measure, so we cannot apply \cref{thm:deterministic}(a) directly. We instead exploit the explicit form of the matrix induced by Gaussian rewards. Since $|\cG| = 1$, \cref{eq:resolvent} reduces to
    \[
        M^\pi = r_g^\top H \,, \quad \text{where} \quad H \eq (I - \ga \bP^\pi)^{-1} \,.
    \]
    Recall that $\exists!P(Q^\pi)$ holds if $M^\pi$ is column-injective, and define $v_{ss'} \eq H_{\cdot, s} - H_{\cdot, s'}$. Then column-injectivity of $M^\pi$ fails for a goal set $\cG = \{g\}$ iff $r_g^\top v_{ss'} = 0$ for some $s \neq s'$, which we can rewrite as $h_{ss'}(g) = 0$ for some $s \neq s'$, where
    \[
        h_{ss'}(g) \eq r_{g}^\top v_{ss'} = \sum_{k=1}^n e^{-\norm{\phi(s_k) - g}^2 / 2\sig^2} (H_{ks} - H_{ks'}) \,.
    \]
    Since $\norm{\phi(s_k) - g}^2$ is a polynomial in $g$, each summand is a composition of real-analytic functions, so $h_{ss'}$ is real-analytic in $g \in \R^d$. Now assume for contradiction that $h_{ss'} \equiv 0$. Then
    \[
        \sum_{k=1}^n c_k e^{-\norm{\phi(s_k) - g}^2/2\sig^2} = 0 \quad \forall g \in \R^d \,,
    \]
    where $c_k \eq H_{ks} - H_{ks'}$. Since $\phi$ is injective, the centres $x_k = \phi(s_k)$ are distinct, and the Gaussians $\{g \mapsto e^{-\norm{x_k - g}^2/2\sig^2}\}_{k = 1}^n$ centred at any distinct points $x_1, \ldots, x_n \in \R^d$ are linearly independent as functions on $\R^d$, so $c_k = 0$ for all $k$, hence $H_{\cdot, s} = H_{\cdot, s'}$. But $H$ is invertible, so it has pairwise distinct columns, which is a contradiction, so we conclude $h_{ss'} \not\equiv 0$. The zero set of a non-zero real-analytic function on $\R^d$ has Lebesgue measure zero, so $\lambda(h_{ss'}^{-1}(0)) = 0$ for each pair $s \neq s'$. Column-injectivity fails only if $g$ lies in $\bigcup_{s \neq s'} h_{ss'}^{-1}(0)$, a finite union of measure-zero sets, hence
    \[
        \lambda\bigg(\{g \in \R^d \mid \neg \exists! P(Q^\pi, \pi, r_{g})\}\bigg) \leq \sum_{s \neq s'} \lambda\big(h_{ss'}^{-1}(0)\big) = 0 \,. \qedhere
    \]
\end{lproof}

\subsection{Proof of \texorpdfstring{\cref{thm:finite} (Stochastic MDPs)}{Theorem (Finite)}}\label{sec:proof-finite-stoch}

\label{proof:finite}
\recall{theorem:finite}

We split this result into three parts: \cref{prop:finite-exact} for exact $Q$-values, and \cref{prop:finite-approximate,prop:lower-bound} for the upper and lower-bounds of the approximate setting. As foreshadowed in the main text, \cref{prop:finite-exact} is stronger than \cref{thm:finite}(a--b): informally, it states that fixing any three of the four components in $(P, \pi, r, \gamma)$, the set of values of the remaining component for which $M^\pi$ is full-rank, and thus $\exists! P(Q^\pi, \pi, r, \gamma)$, has full Lebesgue measure. \cref{thm:finite}(a) follows from part (b), while \cref{thm:finite}(b) follows from parts (c--d), where the failure set for discount factors is in fact \emph{finite}, strictly sharper in the main text. Part (a) of \cref{prop:finite-exact} moreover shows that ``almost all'' kernels $P$ are identified uniquely for fixed $(\pi, r, \gamma)$. \cref{prop:finite-exact} also deals with optimal and optimal entropy-regularised policies, which call for additional care because such policies themselves vary with $(P, r, \gamma)$.

\subsubsection{Proof of \texorpdfstring{\cref{thm:finite}(a--b)}{Proof of Theorem (Finite)(a)}}\label{sec:proof-finite-a}

To formally state the result, first write $n \eq |\cS|$, $m \eq |\cA|$, $L \eq |\cG|$ and let $\cP_\Delta \eq \{ P : \cS \times \cA \to \Delta(\cS) \}$ be the space of world models, which is isomorphic to the simplex product $(\Delta^n)^{nm}$ and inherits the Lebesgue measure from $\R^{nm(n-1)}$. We write $\cR = \{ r : \cS \times \cG \to \R \}$ for the space of reward functions, $\Pi = \{ \pi : \cS \times \cG \to \Delta(\cA) \}$ for the set of GC policies, and $\Pi^\circ$ for its deterministic subset. Finally, denote by $M^\star(P) = M^{\pi_l^\star(P)}(P) \in \mathbb R^{L \times n}$ the matrix corresponding to the optimal policy $\pi_l^\star(P)$; similarly for $M^\star(\gamma)$ and $M^\star(r)$. For part (d) we will additionally make use of the soft Bellman operator $\cT_\alpha : \R^{n^2 m} \to \R^{n^2 m}$, defined for any $\alpha > 0$ by
\[
    (\cT_\alpha Q)_{lij} = \sum_{k} P_{ijk} \left[ R_{lk} + \gamma \alpha \log\left( \sum_{j'} \exp(Q_{lkj'}/\alpha) \right) \right] \,,
\]
the unique fixed point $Q^\alpha$ (cf. \cref{lem:soft-bellman-contraction}), and the optimal entropy-regularised policy $\pi^\star_\alpha$ given by
\[
    (\pi^\star_\alpha)_{lij} = \frac{\exp(Q^\alpha_{lij}/\alpha)}{\sum_{j'} \exp(Q^\alpha_{lij'}/\alpha)} \,.
\]

Recall that $P$ is identified uniquely by $(Q^\pi, \pi, r, \gamma)$, written $\exists!P(Q^\pi)$, if there is a unique solution $P$ to the Bellman equation $\cT^\pi_P(Q^\pi) = Q^\pi$, which reduces to $M^\pi$ being full-rank (\cref{eq:bellman-linear}).

\begin{linked}[Finite MDPs, Exact]{proposition}{finite-exact}\label{prop:finite-exact}
    Fix any finite state space $\cS$ and goals $(\cG, r)$ that span $\cS$.
    \begin{enumerate}[label=(\alph*)]
        \item For any $(\cdot, r, \gamma, \pi)$, the set $\{ P \in \cP_\Delta \mid \rank(M^\pi(P)) < n \}$ has Lebesgue measure zero. So does the optimal-policy set $\{ P \in \cP_\Delta \mid \rank(M^\star(P)) < n \}$.
        \item For any $(P, \cdot, \gamma, \pi)$, the set $\left\{ r \in \cR \mid \rank(M^\pi(r)) < n \right\}$ has Lebesgue measure zero. So does the optimal-policy set $\left\{ r \in \cR \mid \rank(M^\star(r)) < n \right\}$.
        \item For any $(P, r, \cdot, \pi)$, the set $\{ \gamma \in [0, 1) \mid \rank(M^{\pi}(\gamma)) < n \}$ is finite. So is the optimal-policy set $\{ \gamma \in [0, 1) \mid \rank(M^\star(\gamma)) < n \}$. Moreover, there exist $\gamma^-, \gamma^+ \in (0, 1)$ such that $M^{\pi}(\gamma)$ and $M^\star(\gamma)$ are full-rank for all $0\leq \gamma < \gamma^-$ or $\gamma > \gamma^+$.
        \item For any $(P, r, \gamma, \cdot)$ the set $\{ \pi \in \Pi \mid \rank(M^\pi) < n \}$ has Lebesgue measure zero. Moreover, the set $\{ \alpha \in [0, \infty) \mid \rank(M^{\pi^\star_{\alpha}}) < n \}$ is finite, where $\pi^\star_{\alpha}$ is the optimal entropy-regularised policy with entropy coefficient $\alpha$, and there exist $\alpha^-, \alpha^+ \in (0, \infty)$ such that $M^{\pi^\star_{\alpha}}$ is full-rank for all $\alpha < \alpha^-$ or $\alpha > \alpha^+$.
    \end{enumerate}
\end{linked}

\begin{lproof}
    We use $n = |\cS|$, $m = |\cA|$, $L = |\cG|$ throughout, and recall the resolvent identity
    \[
        M_l = R_l\left(I-\gamma \bP^{\pi_l}\right)^{-1}
    \]
    from \cref{lem:resolvent}. Though each row is produced by an invertible matrix (scaled by $R_l$), the resulting concatenation $M$ may be rank-deficient. However, we can still prove that full-rankness is a generic property over each of the four variables, using the standard characterisation $\rank(M) < n \iff \det(M^\tr M) = 0$. The order is chosen to facilitate recycling along the way.

    \textbf{Part (c).} Fix $P, r$ and first consider \textbf{(i)} a fixed GC policy $\pi$, taking $M(\cdot)$ as a function of $\gamma$, omitting the superscript $\pi$ for simplicity. \cref{eq:resolvent} gives $M_{lk} = \sum_{k'} R_{lk'} \left(I-\gamma \bP^{\pi_l}\right)^{-1}_{k'k}$, so writing $C^{(l)}_{k'k}$ for the $(k',k)$th cofactor of $\left(I-\gamma \bP^{\pi_l}\right)$, we use Cramer's rule to obtain
    \begin{equation}
        M_{lk} = \frac{\sum_{k'} R_{lk'} C^{(l)}_{k'k}}{\det(I-\gamma \bP^{\pi_l})} \,. \label{eq:M-entries}
    \end{equation}
    Each cofactor $C^{(l)}_{k'k}$ is the signed determinant of an $(n-1) \times (n-1)$ submatrix of $\left(I-\gamma \bP^{\pi_l}\right)$, which is a polynomial of degree no greater than $n-1$ in $\gamma$. Since the coefficients $R_{lk'}$ are constants (independent of $\gamma$), the numerator of $M_{lk}$ is also a polynomial of degree at most $n-1$. Now for any $I \subset [L]$ with $|I| = n$, writing $M_I \in \R^{n \times n}$ for the submatrix of $M$ with rows indexed by $I$, the Leibniz formula gives
    \begin{align*}
        \det(M_I) = \sum_\sigma \text{sgn}(\sigma) \prod_{l \in I} M_{l\sigma(l)} = \frac{\sum_\sigma \text{sgn}(\sigma) \prod_{l \in I} \left(\sum_{k'} R_{l k'} C^{(l)}_{k'\sigma(l)}\right)}{\prod_{l \in I} \det(I-\gamma \bP^{\pi_l})} \,,
    \end{align*}
   which is a rational function of $\gamma$ whose numerator $A_I$ has degree
    \begin{align*}
        \deg(A_I) &\leq \max_\sigma \sum_{l \in I} \deg\left(\sum_{k'} R_{lk'} C^{(l)}_{k'\sigma(l)}\right) \leq \max_\sigma \sum_{l \in I} (n-1) = n(n-1) \,.
    \end{align*}
    By the Cauchy--Binet formula, $\det(M^\tr M) = \sum_{|I|=n} \det(M_I)^2$, which is rational in $\gamma$ with numerator of degree at most $2n(n-1)$. Finally note that $\det(M^\tr M)$ cannot be uniformly zero, since $M = R$ for $\gamma = 0$, and spanning goals imply $\rank(R) = n$, hence $\det(R^\tr R) \neq 0$. In particular, the set of roots of the non-zero rational function $f \eq \det(M^\pi(\cdot)^\tr M^\pi(\cdot)) : [0, 1) \to \R$ is finite and has cardinality bounded above by the degree of its numerator, so we conclude that $\{ \gamma \in [0, 1) \mid \rank(M^\pi(\gamma)) < n \}$ has cardinality $C \leq 2n(n-1)$.

    We now turn to \textbf{(ii)} optimal policies $\pi^\star$, which depend on $\gamma$. The key is noticing that the optimal value function $V^\star(\gamma)$ is componentwise equal to:
    \[
        V^\star(\gamma) = \max_{\pi \in \Pi^\circ} V^{\pi}(\gamma) \,,
    \]
    where $\Pi^\circ$ is the set of goal-conditioned \textit{deterministic} policies, which is a finite set bounded of size $m^{n^2}$ ($m^n$ possible deterministic policies for each of the $n$ goals). It follows immediately that 
    \[
        M^\star(\gamma) = R + \gamma V^\star(\gamma) = \max_{\pi \in \Pi^\circ} M^{\pi}(\gamma) \,,
    \]
    hence
    \[
        \{ \gamma \in [0, 1) \mid \rank(M^\star(\gamma)) < n \} \subseteq \bigcup_{\pi \in \Pi^\circ} \{ \gamma \in [0, 1) \mid \rank(M^{\pi}(\gamma)) < n \} \,.
    \]
    Applying part \textbf{(ii)} to each $\pi \in \Pi^\circ$, we obtain
    \begin{align*}
        C^\star &= \absB{ \{ \gamma \in [0, 1) \mid \rank(M^\star(\gamma)) < n \} } \\
        &\leq \absB{ \bigcup_{\pi \in \Pi^\circ} \{ \gamma \in [0, 1) \mid \rank(M^{\pi}(\gamma)) < n \} } \\
        &\leq \sum_{\pi \in \Pi^\circ} \absB{ \{ \gamma \in [0, 1) \mid \rank(M^{\pi}(\gamma)) < n \} } \leq 2 m^{n^2} n(n-1) \,.
    \end{align*}

    In particular, since the zeroes are finite, there exists $\gamma^+ \in (0,1)$ such that $M^\pi(\gamma)$ and $M^\star(\gamma)$ are full-rank for all $\gamma > \gamma^+$. Moreover, since $M^\pi$ is continuous in $\gamma$ and $M^\pi(0) = M^\star(0) = R$ has full rank by assumption that goals span the state space, there exists $\gamma^-$ such that $M^\pi(\gamma)$ and $M^\star(\gamma)$ are full-rank for all $0 \leq \gamma < \gamma^-$ (otherwise the set of zeroes would be infinite).

    \paragraph{Part (a).} Now fix $r, \gamma$ and begin with \textbf{(i)} a fixed GC policy $\pi$, taking $M(\cdot)$ as a function of $P$. Again, \cref{eq:M-entries} guarantees that each entry $M_{lk}(\cdot)$ is a rational function in the entries of $P$, so $f \eq \det(M(\cdot)^\tr M(\cdot)) : \cF_\Delta \to \R$ must also be rational. To prove $f \not\equiv 0$, consider the static world $P_0(s' \mid s, a) = \delta_{ss'}$, which satisfies $\bP_0^{\pi_l} = I$ for any policy $\pi_l$. Then using \cref{eq:resolvent},
    \[
        M(P_0) = R\left( I-\gamma I \right)^{-1} = \frac{R}{1-\gamma} \quad \implies \quad f(P_0) = \frac{\det(R^\tr R)}{(1-\gamma)^{2n}} \neq 0 \,,
    \]
    since spanning goals satisfy $\rank(R) = n$.
    Now consider the set $X = \{ x \in \R^{n-1} \mid x_i \geq 0, \sum_i x_i \leq 1 \} \subset \R^{n-1}$ and the bijective map $g : X \to \Delta^n$ given by $g(x) = (x_1, \ldots, x_{n-1}, 1-\sum_i x_i)$. The intrinsic Lebesgue measure $\lambda_\Delta$ on $\Delta^n \subset \R^n$ is naturally inherited from the Lebesgue measure on $\R^{n-1}$ by defining
    \begin{equation} \lambda_\Delta(A) \eq \lambda(g^{-1}(A)) \label{eq:measure-pullback} \end{equation}
    for any measurable set $A$. This extends to the product map $G = g^{mn} : X^{mn} \to (\Delta^n)^{mn} = \cF_\Delta$. We now consider the pulled-back function $h : X^{mn} \to \R$ given by $h = f \circ G$. Since $f$ and $G$ are both non-zero rational functions, and $G$ is bijective, $h$ is also a non-zero rational map. The set of roots of non-zero rational functions $\R^d \to \R$ has Lebesgue measure zero, which also holds for non-zero rational maps $U \to \R$ on any subset $U \subset \R^d$ (by monotonicity of the Lebesgue measure), hence
    \begin{equation} \lambda(h^{-1}(0)) = 0 \,. \label{eq:measure-zero-roots} \end{equation}
    We conclude that the bad set $B = \{ P \in \cF_\Delta \mid \rank(M(P)) < n \}$ satisfies
    \begin{align*}
        \lambda_\Delta(B) \ =  \lambda_\Delta(f^{-1}(0)) \ \eqby{\eqref{eq:measure-pullback}} \ \lambda(G^{-1}( f^{-1} (0) )) \ = \ \lambda \big( (f \circ G)^{-1}(0) \big) \ = \ \lambda(h^{-1}(0)) \ \eqby{\eqref{eq:measure-zero-roots}} \ 0 \,.
    \end{align*}

    The argument for \textbf{(ii)} optimal policies is similar to part \textbf{(a)(ii)}. We have
    \[
        M^\star(P) = R + \gamma V^\star(P) = \max_{\pi \in \Pi^\circ} M^{\pi}(P) \,,
    \]
    where $\Pi^\circ$ is the finite set of goal-conditioned deterministic policies, hence
    \[
        \{ P \in \cF_\Delta \mid \rank(M^\star(P)) < n \} \subseteq \bigcup_{\pi \in \Pi^\circ} \{ P \in \cF_\Delta \mid \rank(M^{\pi}(P)) < n \} \,.
    \]
    Applying part \textbf{(b)(i)} to each $\pi \in \Pi^\circ$, we obtain
    \begin{align*}
        \lambda_\Delta \big(\{ P \in \cF_\Delta \mid \rank(M^\star(P)) < n \} \big)
        &\leq \lambda_\Delta \big( \bigcup_{\pi \in \Pi^\circ} \{ P \in \cF_\Delta \mid \rank(M^{\pi}(P)) < n \} \big) \\
        &\leq \sum_{\pi \in \Pi^\circ} \lambda_\Delta \big( \{ P \in \cF_\Delta \mid \rank(M^{\pi}(P)) < n \} \big) = 0 \,.
    \end{align*}

    \paragraph{Part (b).} Now fix $P$, $\gamma$ and $\pi$ and take $M^\pi{(\cdot)}$ as a function of the reward $r \in \cR$, written as a matrix $R_{lk} \in \R^{L \times n}$. Since each entry $M_{lk}$ is linear in the entries of $R$ by \cref{eq:resolvent}, the function $f \eq \det(M^\pi(\cdot)^\tr M^\pi(\cdot)) : \R^{L \times n} \to \R$ is a polynomial of degree $2n$ in the entries of $R$. To prove that $f \not\equiv 0$, consider the reward matrix $R^0$ with rows $R^0_l = e_l(I-\gamma \bP^{\pi_l})$ for $l \in [n]$ and $R^0_l = 0$ for $l > n$. Then for $l \in [n]$,
    \[
        M_l = R^0_l (I-\gamma \bP^{\pi_l})^{-1} = e_l(I-\gamma \bP^{\pi_l})(I-\gamma \bP^{\pi_l})^{-1} = e_l \,,
    \]
    while $M_l = 0$ for $l > n$, so $M^\tr M = I$ and $f(R^0) = \det(I) \neq 0$. The zero set of a non-zero polynomial has Lebesgue measure zero, so
    \[
        \lambda \big( \{ R \in \R^{L \times n} \mid \rank(M^\pi(R)) < n \} \big) = \lambda(f^{-1}(0)) = 0 \,.
    \]
    The result for optimal policies is identical to part \textbf{(a)(ii)}.

    \paragraph{Part (d).} Now fix $P, r, \gamma$ and take $M^{(\cdot)}$ as a function of $\pi$. The proof over policies is identical to parts \textbf{(a-c)(i)}: $\det((M^\pi)^\tr M^\pi)$ is a rational function over the entries of $\pi$, and taking any goal-independent policy $\pi_0$ (eg the uniform policy) gives
    \[
        M^{\pi_0} = R \left( I - \gamma \bP^{\pi_0} \right)^{-1}
    \]
    which has rank $\rank(R) = n$ by assumption of spanning goals, so $\det((M^{\pi_0})^\tr M^{\pi_0}) \neq 0$. The pullback argument establishes zero measure.

    The second claim for entropy-regularised policies $\pi^\star_\alpha$ is less straightforward, because the matrix $M^{\pi^\star_\alpha}$ is not rational in $\alpha$. Recall from above that $Q^\alpha$ is the unique fixed point of the soft Bellman operator $\cT_\alpha$ (well-defined by \cref{lem:soft-bellman-contraction}), and $\pi^\star_\alpha$ is the softmax of $Q^\alpha$ at temperature $\alpha$. Now let $f : (0, \infty) \to \R$ be defined by $f(\alpha) = \det((M^{\pi^\star_\alpha})^\tr M^{\pi^\star_\alpha})$. By \cref{lem:definable-soft-Q}, the function $\alpha \mapsto Q^\alpha$ is analytic and definable in $\R_{\exp}$. The matrix $M^{\pi^\star_\alpha}$ is obtained from $Q^\alpha$ and $\pi^\star_\alpha$ using definable operations (polynomials, division, $\exp$, $\log$), and $\det((M^{\pi^\star_\alpha})^\tr M^{\pi^\star_\alpha})$ is a polynomial in the entries of $M^{\pi^\star_\alpha}$, so $f$ is definable. It is analytic for the same reasons. As $\alpha \to \infty$, the soft-optimal policy converges to the uniform policy $\pi_0$, which is goal-independent, so $\det((M^{\pi_0})^\tr M^{\pi_0}) \neq 0$. By continuity,
    \[ \lim_{\alpha \to \infty} f(\alpha) = \det((M^{\pi_0})^\tr M^{\pi_0}) \neq 0 \,, \]
    so $f$ is a non-zero, analytic and definable function. Since $\R_{\exp}$ is o-minimal \cite[Second Main Theorem]{Wilkie1996}, and $f$ is definable, the zero set $\{ \alpha \in (0, \infty) \mid f(\alpha) = 0 \}$ is a finite union of points and open intervals. But $f$ cannot be zero on an interval by the identity theorem for analytic functions, so the zero set is a finite set of $C$ points. In particular, there exist $\alpha^-,\alpha^+ \in (0, \infty)$ with $\alpha^- < \alpha^+$ such that $f(\alpha) \neq 0$ for all $\alpha \notin (\alpha^-, \alpha^+)$, since otherwise the zero set of $f$ would be infinite, and thus $M^{\pi^\star_{\alpha}}$ is full-rank for all $\alpha \notin (\alpha^-, \alpha^+)$. Adding the single boundary point $\alpha = 0$ preserves finiteness:
    \[
        \big| \{ \alpha \in [0, \infty) \mid \rank(M^{\pi^\star_\alpha}) < n \} \big| \subseteq \big\lvert \{ 0 \} \cup \{ \alpha \in (0, \infty) \mid f(\alpha) = 0 \} \big\rvert \leq 1+C < \infty \,. \qedhere
    \]
\end{lproof}

\subsubsection{Proof of \texorpdfstring{\cref{thm:finite}(c)}{Proof of Theorem (Finite)(c)}}\label{sec:proof-finite-b}

We split \cref{thm:finite}(c) into two parts: \cref{prop:finite-approximate} for the upper bound and \cref{prop:lower-bound} for the worst-case lower bound. Recall that a policy is called \textit{unconditional} if $\pi_g = \pi_{g'}$ for all $g \in \cG$.

\begin{linked}[Finite MDPs, Approximate]{proposition}{finite-approximate}\label{prop:finite-approximate}
    Fix a finite MDP and assume $(\cG, r)$ span $\cS$, $Q$ is $\ep$-approximate, and $M$ has full rank. Then the estimator $\hP_{ij} = M^{+} Q_{ij}$ satisfies
    \[
        \norm{\hP_{ij} - P_{ij}}_1 \leq \norm{M^+}_1 (1 + \ga m) \ep \,.
    \]
    Moreover, a sufficient condition for $M$ to be full-rank is for $M^\pi$ to be full-rank and $\ep < 1/\ga m \norm{(M^\pi)^+}_1$. If $\pi$ is unconditional and goals are states ($\cG = \cS$), then $M^\pi$ is always full-rank and the bound improves to
    \[
        \norm{\hP_{ij} - P_{ij}}_1 \leq \frac{\ep (1 + \ga)^2}{1 - \ep \ga (1 + \ga)} \leq \frac{4\ep}{1 - 2\ep} \in O(\ep)
    \]
    for all $\ep < 1/\ga (1 + \ga)$, which is satisfied whenever $\ep < 1/2$.
\end{linked}

\begin{lproof}
    We first decompose $M = M^\pi + E$, where $E_{lk} = \ga \sum_{j'} \pi_{lkj'} (Q_{lkj'} - Q^\pi_{lkj'})$ satisfies
    \begin{align*}
        \norm{E}_1 &= \max_k \sum_l \abs{E_{lk}} \leq \ga \max_k \sum_l \sum_{j'} \pi_{l,kj'} \abs{Q_{lkj'} - Q^\pi_{lkj'}} \\
        &\leq \ga \max_k \sum_l \sum_{j'} \abs{Q_{lkj'} - Q^\pi_{lkj'}} = \ga \max_k \sum_{j'} \norm{Q_{kj'} - Q^\pi_{kj'}}_1 \leq \ga m \ep \,.
    \end{align*}
    If $M$ is full-rank, then $M^+M^\pi = M^+(M-E) = I-M^+E$ and
    \begin{align*}
        \hP_{ij} - P_{ij} &= M^+Q_{ij} - P_{ij} = M^+(Q_{ij} - Q^\pi_{ij}) + (M^+M^\pi - I)P_{ij} \\
        &= M^+(Q_{ij} - Q^\pi_{ij}) - M^+ E P_{ij} = M^+\big(Q_{ij} - Q^\pi_{ij} - E P_{ij}\big) \,.
    \end{align*}
    Taking norms and using $\norm{P_{ij}}_1 = 1$,
    \begin{equation*}
        \norm{\hP_{ij} - P_{ij}}_1 \leq \norm{M^+}_1 \left( \ep + \ga m \ep \right) = \norm{M^+}_1 (1 + \ga m) \ep \,.
    \end{equation*}
    A sufficient condition for $M$ to be full-rank is $\ep < 1/\ga m \norm{(M^\pi)^+}_1$, since then $(M^\pi)^+ M = I - (M^\pi)^+ E$ is invertible if $\norm{(M^\pi)^+}_1 \norm{E}_1 < 1$, and $(M^\pi)^+ M$ being invertible forces $M$ full-rank since $M x = 0$ implies $x = 0$. When $\pi$ is unconditional and $\cG = \cS$, factoring out $\pi_{lkj'} = \pi_{kj'}$ sharpens the bound to $\norm{E}_1 \leq \ga \ep$ and \cref{eq:resolvent} gives $M^\pi = (I - \ga (P^\pi)^\tr)^{-1}$, implying $M^\pi$ is always invertible with $\norm{(M^\pi)^{-1}}_1 = 1 + \ga$ by row-stochasticity of $P^\pi$. Then
    \begin{align*}
        \norm{\hP_{ij} - P_{ij}}_1 &\leq \norm{(M^\pi)^{-1}}_1 \norm{M^\pi \hP_{ij} - M^\pi P_{ij}}_1 \\
        &= (1 + \ga) \norm{(M^\pi - M)\hat{P}_{ij} + Q_{ij} - Q^\pi_{ij} }_1 \\
        &\leq (1+\ga) \left( \norm{E}_1 \norm{\hat{P}_{ij}}_1 + \norm{Q_{ij} - Q^\pi_{ij}}_1 \right) \\
        &\leq (1+\ga) \left( \ga \ep \Big( \norm{\hat{P}_{ij}-P_{ij}}_1 + 1 \Big) + \ep \right) \,.
    \end{align*}
    Solving for $\norm{\hP_{ij} - P_{ij}}_1$, we conclude
    \[
        \norm{\hP_{ij} - P_{ij}}_1 \leq \frac{\ep (1 + \ga)^2}{1 - \ep \ga (1 + \ga)} \leq \frac{4\ep}{1 - 2\ep} \in O(\ep)
    \]
    for all $\ep < 1/\ga (1 + \ga)$, which is satisfied whenever $\ep < 1/2$.
\end{lproof}

We now show that the upper bound of \cref{prop:finite-approximate} is tight up to a constant factor.

\begin{linked}[Worst-Case Lower Bound]{proposition}{lower-bound}\label{prop:lower-bound}
    For any $n \geq 2$, $m \geq 2$, $\ga \in [0, 1)$, and $\ep > 0$ sufficiently small, there exist two MDPs with transition kernels $P, P' \in \cP_\Delta$, a goal set $\cG$ that spans the state space, and a policy $\pi$ such that
    \begin{enumerate}[label=(\roman*)]
        \item $\norm{Q^\pi_{ij}(P) - Q^\pi_{ij}(P')}_1 \leq \ep$ for all $(i, j)$;
        \item $\norm{P_{ij} - P'_{ij}}_1 \geq (1+\ga)\ep$ for some $(i, j)$.
    \end{enumerate}
    Consequently, any estimator $\hat P$ mapping $\ep$-approximate $Q$-values to a world model incurs worst-case error at least $(1+\ga)\ep/2 \in \Omega(\ep)$.
\end{linked}

\begin{lproof}
    Take $n = 2$, $m = 2$, $L = 2$ with indicator goals $\cG = \cS$. The construction below extends to any $n \geq 2, m \geq 2$ by adding identity-transition states $s_3, \ldots, s_n$ and self-loop actions $a_3, \ldots, a_m$. We define two transition kernels as follows: $P$ always swaps states regardless of actions, i.e. $P(s_2 \mid s_1, \cdot) = 1$, $P(s_1 \mid s_2, \cdot) = 1$, and $P'$ agrees with $P$ everywhere except $P'(\cdot s_1, a_1) = (\delta, 1-\delta)$ for some $\delta \in (0, 1)$. Now take the deterministic policy $\pi(a_2 \mid s_k) = 1$ for $k \in \{1, 2\}$ that always selects the second action $a_2$. Since $a_1$ is off-policy, the value function $V^\pi$ depends only on $a_2$-transitions, which are identical in $P$ and $P'$, hence $V^\pi(P) = V^\pi(P')$ and $M^\pi(P) = M^\pi(P')$. The policy-induced transition $\bP^\pi = (P^\pi)^\tr$ is the swap matrix, so
    \[
        M^\pi = (I - \ga \bP^\pi)^{-1} = \frac{1}{1-\ga^2}\begin{pmatrix} 1 & \ga \\ \ga & 1 \end{pmatrix} \,.
    \]
    Comparing action-value functions, we have $Q_{lij}(P) = \sum_{k} P_{ijk} M^\pi_{lk} =  Q_{lij}(P')$ for all $(i,j) \neq (1, 1)$ since $P, P'$ agree everywhere except at $(s_1, a_1)$. For the latter,
    \[
        Q^\pi_{l11}(P) = \sum_k P(s_k \mid s_1, a_1) M^\pi_{lk} = M^\pi_{l2} \,, \quad Q^\pi_{l11}(P') = M^\pi_{l2} + \delta(M^\pi_{l1} - M^\pi_{l2}) \,,
    \]
    so
    \[ \norm{Q^\pi_{11}(P) - Q^\pi_{11}(P')}_1 = \delta \sum_l \abs{ M_{l1}^\pi - M_{l2}^\pi } = \frac{2\delta}{1-\ga^2} \abs{ 1- \gamma} = \frac{2\delta}{1+\gamma} \,. \]
    Setting $\delta \eq (1+\ga)\ep/2$ gives $\norm{Q^\pi_{ij}(P) - Q^\pi_{ij}(P')}_1 \leq \ep$ for all $(i, j)$ and $\norm{P_{1, 1} - P'_{1, 1}}_1 = 2\delta = (1+\ga)\ep$, as claimed. For $\ep < 2/(1+\ga)$ we have $\delta < 1$, so $P'$ is a valid world model. In particular, for any estimator $\hat P$, let $\hQ \eq Q^\pi(P)$ and notice that $(P, \hQ)$ and $(P', \hQ)$ satisfy $\norm{\hQ - Q^\pi(P)}_1 = 0 \leq \ep$ and $\norm{\hQ - Q^\pi(P')}_1 \leq \ep$. The triangle inequality gives
    \[
        \norm{\hat P(\hQ) - P}_1 + \norm{\hat P(\hQ) - P'}_1 \geq \norm{P - P'}_1 = (1+\ga)\ep \,,
    \]
    so $\max \bigl(\norm{\hat P(\hQ) - P}_1, \norm{\hat P(\hQ) - P'}_1\bigr) \geq (1+\ga)\ep/2$. Hence at least one of $P, P'$ is at distance $\geq (1+\ga)\ep/2$ from $\hat P(\hQ)$, and since $\hat P$ cannot distinguish between them (both are $\ep$-feasible given $\hQ$), the worst-case error of $\hat P$ on $\ep$-approximate inputs is at least $(1+\ga)\ep/2$.
\end{lproof}

\subsubsection{Supporting Results}\label{sec:proofs:multi-TTT:d}

We prove supporting results for \cref{prop:finite-exact}, starting with the soft Bellman equation.

\begin{lemma}[Soft Bellman]\label{lem:soft-bellman-contraction}
    For any $\alpha > 0$, $\gamma \in [0, 1)$, the soft Bellman operator $\cT_\alpha$ given by
    \[
        (\cT_\alpha Q)_{lij} = \sum_{k} P_{ijk} \left[ R_{lk} + \gamma \alpha \log\left( \sum_{j'} \exp(Q_{lkj'}/\alpha) \right) \right]
    \]
    is a $\gamma$-contraction in the supremum norm: $\| \cT_\alpha Q - \cT_\alpha Q' \|_\infty \leq \gamma \| Q - Q' \|_\infty$.
\end{lemma}
\begin{lproof}
    The log-sum-exp function $\alpha \log(\sum_{j'} \exp(Q_{lkj'}/\alpha))$ is $1$-Lipschitz in $Q$ with respect to the supremum norm. Since $P_{ijk}$ sums to $1$ over $k$,
    \begin{align*}
        |(\cT_\alpha Q)_{lij} - (\cT_\alpha Q')_{lij}| &= \gamma \left| \sum_{k} P_{ijk} \alpha \log\left( \frac{\sum_{j'} \exp(Q_{lkj'}/\alpha)}{\sum_{j'} \exp(Q'_{lkj'}/\alpha)} \right) \right| \\
        &\leq \gamma \sum_{k} P_{ijk} \left|\alpha  \log\left( \frac{\sum_{j'} \exp(Q_{lkj'}/\alpha)}{\sum_{j'} \exp(Q'_{lkj'}/\alpha)} \right) \right| \\
        &\leq \gamma \sum_{k} P_{ijk} \| Q - Q' \|_\infty = \gamma \| Q - Q' \|_\infty \,. \qedhere
    \end{align*}
\end{lproof}

We will use the following background on \textit{definable} functions to show that the map $\alpha \mapsto Q^\alpha$ is particularly well-behaved.

\begin{definition}
    A set $S \subseteq \R^n$ is \emph{definable} in $\R_{\exp} = (\R, <, +, \cdot, \exp)$ if it can be described by a first-order formula using the symbols $<, +, \cdot, \exp$ and quantifiers over $\R$. For example, the graph of the logarithm function $G_\text{log} = \{(x,y) \mid x > 0 \ \land \ \exp(y) = x\}$ is definable. A function $f : A \to \R^m$ (where $A \subseteq \R^n$) is \emph{definable} if its graph $\{(x, f(x)) \mid x \in A\} \subseteq \R^{n+m}$ is a definable set. For example, $\log : \R^+ \to \R$ is definable, but $\sin : \R \to \R$ is not.
\end{definition}

A foundational result of \citet[Second Main Theorem]{Wilkie1996} establishes that $\R_{\exp}$ is \emph{o-minimal}: every definable subset of $\R$ is a finite union of points and open intervals. This is a strong tameness property which implies, for instance, that the set of zeroes of a definable function $f : \R \to \R$ is a finite union of points and open intervals. If $f$ is also a non-trivial analytic function, then it cannot be zero on an open interval, so it must have a finite set of zeroes (analogous to polynomials!)

\begin{lemma}[Soft Bellman definability]\label{lem:definable-soft-Q}
    For each $\alpha > 0$, let $Q^\alpha$ be the unique fixed point of the soft Bellman operator $\cT_\alpha$. Then the function $\alpha \mapsto Q^\alpha$ is definable and real analytic in $\R_{\exp}$.
\end{lemma}

\begin{lproof}
    The set $R = \{(\alpha, Q) \in (0, \infty) \times \R^{n^2m} : \cT_\alpha Q = Q\}$ is definable in $\R_{\exp}$, since $\cT_\alpha$ involves only polynomial operations and the functions $\exp$ and $\log$. By Lemma \ref{lem:soft-bellman-contraction}, $\cT_\alpha$ is a contraction, so there is a unique $Q^\alpha$ such that $\cT_\alpha Q^\alpha = Q^\alpha$ for each $\alpha > 0$. In particular, $R$ is the graph of $\alpha \mapsto Q^\alpha$, so the map is definable.
    
    Now define $F : (0, \infty) \times \R^{n^2 m} \to \R^{n^2 m}$ by $F(\alpha, Q) = Q - \cT_\alpha Q$, which is analytic because it is the composition of analytic functions (polynomials, $\exp$ and $\log$). The Jacobian of $F$ with respect to $Q$ is the $n^2 m \times n^2 m$ matrix $D_Q F = I - D_Q \cT_\alpha$. By \cref{lem:soft-bellman-contraction}, $\cT_\alpha$ is a $\gamma$-contraction in the sup-norm. For any unit vector $v$ with $\|v\|_\infty = 1$,
    \[
        \|D_Q \cT_\alpha \cdot v\|_\infty = \lim_{t \to 0} \frac{\|\cT_\alpha(Q + tv) - \cT_\alpha Q\|_\infty}{|t|}  \leq \lim_{t \to 0} \frac{\gamma \|tv\|_\infty}{|t|} = \gamma \|v\|_\infty = \gamma \,,
    \]
    so taking the supremum over unit vectors implies $\|D_Q \cT_\alpha\|_\infty \leq \gamma < 1$. It follows that the Neumann series $\sum_{k \geq 0} (D_Q \cT_\alpha)^k$ converges in operator norm to $(I - D_Q \cT_\alpha)^{-1}$, so $D_Q F = I - D_Q \cT_\alpha$ is invertible. In particular, $D_Q F$ and its inverse are continuous in $(\alpha, Q)$, so for any $(\alpha_0, Q_0)$ such that $F(\alpha_0, Q_0) = 0$, the analytic implicit function theorem \citep{KrantzParks2002} implies that there is an analytic $G_U : U \to \R^{n^2 m}$ such that $F(\alpha, G(\alpha)) = 0$ for all $\alpha$ in some neighbourhood $U$ of $\alpha_0$. Uniqueness of solutions $Q^\alpha$ to $F(\alpha, Q^\alpha) = 0$ implies that the local maps $G_U$ agree everywhere: for two different neighbourhoods $U, V$, the analytic functions $G_U$ and $G_V$ must satisfy $F(\alpha, G_U(\alpha)) = 0 = F(\alpha, G_V(\alpha))$ for all $\alpha \in U \cap V$, but this implies $G_U(\alpha) = Q^\alpha = G_V(\alpha)$, so $G_U = G_V$. In particular, there is a global analytic function $G : (0, \infty) \to \R^{n^2 m}$ such that $F(\alpha, G(\alpha)) = 0$ for all $\alpha > 0$. By uniqueness, $Q^\alpha = G(\alpha)$, so $\alpha \mapsto Q^\alpha$ is real analytic on $(0, \infty)$.
\end{lproof}

\subsection{Proofs of \texorpdfstring{\cref{cor:perturbed-rewards-stochastic} (Perturbed Rewards) and \cref{prop:concrete-rewards-stochastic} (Gaussian Rewards)}{Proposition (Concrete Rewards, Stochastic)}}\label{sec:proof-concrete-rewards-stochastic}

\cref{prop:finite-deterministic-a} for deterministic kernels guaranteed that $\exists!P(Q^\pi)$ almost surely under any absolutely continuous perturbation $\xi$ of a single-goal base reward $r_0 \in \R^{|\cS|}$. The same translation-invariance argument extends to the stochastic setting of \cref{thm:finite}, viewing the reward as a matrix and the perturbation as matrix-valued.

\begin{linked}[Perturbed Rewards, Stochastic]{corollary}{perturbed-rewards-stochastic}\label{cor:perturbed-rewards-stochastic}
    Fix any finite MDP, policy $\pi$ goal set $|\cG| \geq |\cS|$, base reward $R_0 \in \R^{|\cG| \times |\cS|}$ and random matrix $\Xi \in \R^{|\cG| \times |\cS|}$ whose distribution $\mu$ is absolutely continuous with respect to Lebesgue measure (e.g.\ entrywise independent Gaussians). Then the perturbed reward $R \eq R_0 + \Xi$ satisfies $\exists!P(Q^\pi)$ almost surely.
\end{linked}

\begin{lproof}
    Viewing reward functions $r : \cS \times \cG \to \R$ as matrices $R \in \R^{|\cG| \times |\cS|}$ with $R_{lk} \eq r(s_k, g_l)$, \cref{prop:finite-exact}(b) establishes that the set
    \[ B \eq \{ R \in \R^{|\cG| \times |\cS|} \mid \rank(M^\pi(R)) < |\cS| \} \]
    has zero Lebesgue measure. Now fix any base reward $R_0 \in \R^{|\cG| \times |\cS|}$ and any random matrix $\Xi \in \R^{|\cG| \times |\cS|}$ whose distribution $\mu$ is absolutely continuous with respect to Lebesgue measure $\lambda$. By translation-invariance of Lebesgue measure, $\lambda(B - R_0) = \lambda(B) = 0$, and absolute continuity ($\mu \ll \lambda$) implies $\mu(B - R_0) = 0$, i.e.\ $\P_{\Xi \sim \mu}(R_0 + \Xi \in B) = 0$. By definition of $B$, $M^\pi(R_0 + \Xi)$ is almost surely full-rank, hence $\exists! P(Q^\pi, \pi, R_0 + \Xi, \gamma)$ almost surely.
\end{lproof}

We similarly extend \cref{prop:concrete-rewards}, where a single Gaussian reward generically implies $\exists!P(Q^\pi)$ for deterministic MDPs, to the stochastic setting. Extra care is required to define $\pi$ independently from $g$, but the result again extends to include optimal policies defined as a function of $g$.

\begin{linked}[Gaussian Rewards]{proposition}{concrete-rewards-stochastic}\label{prop:concrete-rewards-stochastic}
    Fix any finite MDP, $L \geq |\cS|$, policies $\pi_1, \ldots, \pi_L : \cS \to \Delta(\cA)$, and $\sigma_1, \ldots, \sigma_L > 0$. For any injective map $\phi : \cS \hookrightarrow \R^d$, the goal set $\cG = \{g_1, \ldots, g_L\}$ with Gaussian rewards $r_{g_l}(s) = \exp(-\|\phi(s) - g_l\|^2/2\sigma_l^2)$ and policy $\pi(\cdot|s, g_l) \eq \pi_l(\cdot|s)$ satisfies $\exists! P(Q^\pi)$ for Lebesgue-almost every $(g_1, \ldots, g_L) \in (\R^d)^L$. The same conclusion also holds for the optimal (instead of fixed) policy $\pi^\star$ with respect to $(g_1, \ldots, g_L)$.
\end{linked}

\begin{lproof}
    Write $n \eq |\cS|$ and recall that $\exists! P(Q^\pi)$ if $\rank(M^\pi) = n$, which for $L \geq n$ is equivalent to $\det((M^\pi)^\tr M^\pi) \neq 0$. By \cref{eq:resolvent}, $M^\pi$ has rows
    \[
        M^\pi_l = r_{g_l} H_l \,, \quad \text{where} \quad H_l \eq (I - \ga \bP^{\pi_l})^{-1} \,.
    \]
    Writing $R_{lk} \eq r_{g_l}(s_k) = \exp(-\norm{\phi(s_k) - g_l}^2/2\sig_l^2)$, the matrix $R$ is real-analytic in the goal tuple $\boldg \eq (g_1, \ldots, g_L) \in (\R^d)^L$. Since each $M^\pi_l = R_l H_l$ is linear in $R_l$ with goal-independent $H_l$,
    \[
        f(g_1, \ldots, g_L) \eq \det\big((M^\pi)^\tr M^\pi\big)
    \]
    is real-analytic on $(\R^d)^L$. Now the zero set of a non-zero real-analytic function on $(\R^d)^L$ has Lebesgue measure zero, so it suffices to exhibit one tuple of goals for which $f \neq 0$. We first show that for any fixed $\sig > 0$, the family $S_\sig \eq \{r_{g, \sig} \in \R^n \mid g \in \R^d\}$ spans $\R^n$. Assume otherwise for contradiction; then there exists $w \neq 0$ such that
    \[ \sum_{k=1}^n w_k e^{-\norm{\phi(s_k) - g}^2/2\sig^2} = w^\tr r_{g, \sig} = 0 \]
    for all $g \in \R^d$. Since $\phi$ is injective, the centres $x_k = \phi(s_k)$ are distinct, and the Gaussians $\{g \mapsto e^{-\norm{x_k - g}^2/2\sig^2}\}_{k = 1}^n$ centred at any distinct points $x_1, \ldots, x_n \in \R^d$ are linearly independent as functions on $\R^d$, so we conclude $w_k = 0$ for all $k$, a contradiction. Hence $\text{span}(S_\sig) = \R^n$. We now construct $(g_l)_{l=1}^n$ inductively such that $\{M^\pi_l\}_{l=1}^n = \{r_{g_l} H_l\}_{l=1}^n$ are linearly independent vectors. Proceed by induction on $l$: at step $l \in \{1, \ldots, n\}$, with $g_j$ already chosen for $j < l$, set $V_0 = \{0\}$ and
    \[ V_{l-1} \eq \text{span}\{M^\pi_j\}_{j < l} \]
    for $l-1 \geq 1$, of dimension no greater than $l - 1 < n$. Since $H_l$ is invertible,
    \[ \text{span}(\{r_{g, \sig_l} H_l : g \in \R^d\}) = \text{span}(\{r_{g, \sig_l} : g \in \R^d\}) = \text{span}(S_{\sig_l}) = \R^n \,, \]
    so the family $\{r_{g, \sig_l} H_l : g \in \R^d\}$ cannot be contained in the proper subspace $V_{l-1}$, hence there exists $g_l \in \R^d$ with $r_{g_l} H_l \notin V_{l-1}$. After $n$ steps, we obtain $V_n = \text{span}\{M_j^\pi\}_{j=1}^n = \R^n$ and thus $\rank(M^\pi) = n$. Choosing the remaining $g_l$ for $l > n$ arbitrarily, we obtain $f(g_1, \ldots, g_L) \neq 0$ and thus $f \not\equiv 0$. We conclude $\lambda(f^{-1}(0)) = 0$, and finally
    \[
        \lambda\bigg(\left\{ \boldg \in (\R^d)^L \mid \neg \exists! P(Q^\pi)\right\}\bigg) \leq \lambda\bigg(\left\{\boldg \in (\R^d)^L \mid \rank(M^\pi(\boldg)) < n \right\} \bigg) = \lambda(f^{-1}(0)) = 0 \,.
    \]

    For the optimal-policy claim, let $\Pi^\circ$ denote the finite set of goal-conditioned deterministic policies $\pi : \cS \times [L] \to \cA$, of size $|\cA|^{|\cS| \cdot L}$. For each goal tuple $\boldg \in (\R^d)^L$, define $\pi^\star(\boldg) \in \Pi^\circ$ by $\pi^\star(\boldg)(\cdot, l) \eq \pi^\star(\cdot, g_l)$, the optimal policy for goal $g_l$. Then
    \[
        \{\boldg : \rank(M^{\pi^\star(\boldg)}(\boldg)) < n\} \subseteq \bigcup_{\pi \in \Pi^\circ} \{\boldg : \rank(M^\pi(\boldg)) < n\} \,.
    \]
    For each fixed $\pi \in \Pi^\circ$, decomposing into per-slot policies $\pi_l \eq \pi(\cdot, l)$ and applying the first part of the proposition gives measure zero for the corresponding term. By finite subadditivity,
    \[
        \lambda\big(\{\boldg : \rank(M^{\pi^\star(\boldg)}(\boldg)) < n\}\big) \leq \sum_{\pi \in \Pi^\circ} \lambda\big(\{\boldg : \rank(M^\pi(\boldg)) < n\}\big) = 0 \,. \qedhere
    \]
\end{lproof}

\subsection{Example of Rank Deficiency}\label{sec:proof-singularity}

\cref{thm:finite} proves that $M^\pi$ is almost always full-rank, but singularities can occur for specific $(P, \pi, \gamma)$ even when goals are states ($\cG = \cS$). We construct an explicit example for $n = m = 3$. Let $\cS = \{0, 1, 2\} = \cA$, with deterministic transition kernel $P(s, a) = a$ and deterministic policies $\pi_0(i) \equiv 0$; $\pi_1(0)=1, \pi_1(1) = 1, \pi_1(2) = 2$; $\pi_2(0)=2, \pi_2(1) = 1, \pi_2(2) = 2$. Writing $c = 1/(1-\ga)$, one can analytically compute:
\[
    M^\pi = \begin{pmatrix} c & c-1 & c-1 \\ c-1 & c & 0 \\ c-1 & 0 & c \end{pmatrix} = I + (c-1)\begin{pmatrix} 1&1&1\\1&1&0\\1&0&1 \end{pmatrix} \,.
\]
The rhs matrix has eigenvalues $1+\sqrt{2}$, $1$, $1-\sqrt{2}$, so setting $\ga = 1/\sqrt{2}$ gives $1 + (c-1)(1-\sqrt{2}) = 0$ and thus $\det(M^\pi) = 0$. However, note that the matrix is full-rank for all $\ga \neq 1/\sqrt{2}$, and for any perturbation of the policies, making singularities ``exceptionally rare'' as guaranteed by \cref{thm:finite}.

\subsection{\texorpdfstring{Arbitrary $Q$-Values}{Arbitrary Q-Values}}\label{sec:proof-arbitrary-Q}

\cref{thm:deterministic,thm:finite} give conditions under which the true kernel $P$ is identified uniquely by exact $Q$-values, or approximately by $\ep$-approximate $Q$-values. \cref{prop:duality-deterministic,prop:duality-stochastic} below show that analogous results apply to \textit{arbitrary} $Q \in \cQ$. This is particularly relevant for deep RL agents, which seldom have converged $Q$-values, but may nonetheless encode a unique world model that we denote $P^\star$ -- as opposed to $P$, since there is no guarantee the extracted world model will approximate the true kernel. While the proof techniques are very similar (constructing a map from $Q$ to $M$, instead of $R$ to $M$, whose preimages preserve measure-zero sets), neither follows from the other. First, recall that we write $M$ for the matrix from \cref{eq:bellman-linear} induced by $(Q, \pi, r, \gamma)$.

\begin{linked}[General Uniqueness, Stochastic]{proposition}{duality-stochastic}\label{prop:duality-stochastic}
    Fix a finite MDP, a policy $\pi$, and a reward function $r : \cS \times \cG \to \R$. Then the set $\{ Q \in \cQ \mid \rank(M) < \min(L, n) \}$ has Lebesgue measure zero. Moreover, if $|\cG| \geq |\cS|$, then for every $Q$ outside this set, $P^\star(s,a) \eq M^+ Q(s,a)$ is the unique minimiser of the Bellman residual $\|\cT^\pi_{P^\star}(Q) - Q\|^2$, and if $|\cG| = |\cS|$, then $\cT^\pi_{P^\star}(Q) = Q$.
\end{linked}

\begin{lproof}
    Recall the notation $L = |\cG|, n = |\cS|$ and define the map $F : \cQ \to \R^{L \times n}$ by $F(Q) = R+\gamma V = R+\gamma \pi Q$, i.e. $V_{li} = \sum_{j} \pi_{lij} Q_{lij}$. Note that $F$ is affine and surjective, since for any $M \in \R^{L \times n}$, defining $Q_{lij} = (M_{li} - R_{li})/\gamma$ gives $V_{li} = M_{li} - R_{li}$ and thus $F(Q) = M$. Then the set of $Q \in \cQ$ for which $M$ has deficient-rank is
    \[
        \{ Q \in \cQ \mid \rank(M) < \min(L, n) \} = F^{-1}(\{ M \in \R^{L \times n} \mid \rank(M) < \min(L, n) \})
    \]
    which has Lebesgue measure zero as the preimage of a Lebesgue-measure-zero set under a surjective affine map: the rank-deficient set $\{ M \in \R^{L \times n} \mid \rank(M) < \min(L, n) \}$ is the zero set of $\det(M^\tr M)$ when $L \geq n$, or of $\det(M M^\tr)$ when $L \leq n$, which is a non-zero polynomial in the entries of $M$. Therefore, $M$ has full-rank for almost every $Q \in \cQ$. Now recall that the Bellman equation $\cT^\pi_{P^\star}(Q) = Q$ is a linear system $M P^\star_{ij} = Q_{ij}$ for each $i,j$ (see Equation \ref{eq:bellman-linear}). If $L \geq n$, the rank-nullity theorem gives $\dim(\text{ker}(M)) = n - \dim(\text{im}(M)) = 0$, so the system $M P^\star_{ij} = Q_{ij}$ has at most one solution, and the unique least-squares solution is given by $P^\star_{ij} = M^+ Q_{ij}$. If $L = n$, we have $P^\star_{ij} = M^{-1} Q_{ij}$ and thus $M P^\star_{ij} = Q_{ij}$ exactly, i.e. $\cT^\pi_{P^\star}(Q) = Q$.
\end{lproof}

\begin{linked}[General Uniqueness, Deterministic]{proposition}{duality-deterministic}\label{prop:duality-deterministic}
    Fix a finite MDP, a policy $\pi$, and a reward function $r : \cS \times \cG \to \R$. Then the set $\{ Q \in \cQ \mid \exists k \neq k' \text{ with } M_{\cdot, k} = M_{\cdot, k'} \}$ has Lebesgue measure zero. Moreover, for any $|\cG| \geq 1$ and any $Q$ outside this set, $P^\star(s,a) \eq \argmin_{s'} \norm{Q(s,a) - M_{\cdot, s'}}_1$ is the unique deterministic world model minimising the Bellman residual $\|\cT^\pi_{P^\star}(Q) - Q\|^2$, and at most one deterministic model $P^\star$ satisfies $\cT^\pi_{P^\star}(Q) = Q$.
\end{linked}

\begin{lproof}
    Let $F : \cQ \to \R^{L \times n}$ be the affine surjection from the proof of \cref{prop:duality-stochastic}. For any $k \neq k'$, the set $\{M \in \R^{L \times n} : M_{\cdot, k} = M_{\cdot, k'}\}$ is a proper subspace of $\R^{L \times n}$, so it has Lebesgue measure zero. Taking a finite union over the $\binom{n}{2}$ pairs $k \neq k'$, the set
    \[
        \{ Q \in \cQ \mid \exists k \neq k' \text{ with } M_{\cdot, k} = M_{\cdot, k'} \} = F^{-1}\Bigl( \textstyle\bigcup_{k \neq k'} \{ M \in \R^{L \times n} \mid M_{\cdot, k} = M_{\cdot, k'} \} \Bigr)
    \]
    has Lebesgue measure zero since it is the preimage of a Lebesgue-measure-zero set under a surjective affine map. Therefore, $M$ has pairwise distinct columns for almost every $Q \in \cQ$. Now the Bellman equation $\cT^\pi_{P^\star}(Q) = Q$ reduces in deterministic MDPs to $Q_{ij} = M_{\cdot, P^\star(s_i, a_j)}$ for each $(i,j)$. Since columns are pairwise distinct, the argmin $P^\star(s_i, a_j) = \argmin_{k} \norm{Q_{ij} - M_{\cdot, k}}_1$ is unique and minimises $\norm{Q_{ij} - M_{\cdot, P^\star(s_i, a_j)}}^2 = \norm{\cT^\pi_{P^\star}(Q) - Q}^2$.
\end{lproof}

We note that all results above refer to world models $P^\star \in \cP$ which may lie outside the probability simplex; we refer to \cref{sec:proof-simplex} for discussion.

\subsection{Impossibility Result for Multiple Discount Factors and Policies}\label{sec:proof-multi-horizon}

Below we prove that, unlike multiple goals, multiple discount factors and multiple policies are not usually sufficient to break value equivalence.

\begin{linked}{proposition}{multi-horizon}\label{prop:multi-horizon}
    For any $n \geq 4$, there is a family of finite MDPs with $n$ states in which $P$ cannot be uniquely identified from the set of $Q$-values $\{Q^\pi_{r,\gamma} \mid \pi \in \Pi,\ \ga \in [0, 1)\}$ over \textit{all} policies $\pi$ and \textit{all} discount factors $\ga$, for \emph{any} reward function $r : \cS \to \R$. The non-identifiable family has dimension at least $1$ when $\{r_2, r_3, r_4\}$ are not all equal, and dimension at least $2$ when $r_2 = r_3 = r_4$.
\end{linked}

\begin{lproof}
    Take $n = 4$, $m \geq 1$, and let $s_2, s_3, s_4$ be absorbing states: $P(s_k \mid s_k, a_j) = 1$ for $k \in \{2, 3, 4\}$ and every $a_j$. From $s_1$, fix any interior escape, e.g.
    \[
        P(\cdot \mid s_1, a_1) = \big(p_1, \tfrac{1-p_1}{3}, \tfrac{1-p_1}{3}, \tfrac{1-p_1}{3}\big)
    \]
    for some $p_1 \in (0, 1)$, and $P(\cdot \mid s_1, a_j)$ for $j \geq 2$ arbitrary. For any reward function $r \in \R^4$, we construct a family of transition kernels $P_\ep \neq P$ such that $Q$-values are identical for all $P_\ep$, for all $\pi, \gamma$. First, let
    \[
        \cC_r \eq \{\Delta \in \R^4 : \Delta_1 = 0,\ \mathbf{1}^\top \Delta = 0,\ r^\top \Delta = 0\} \,.
    \]
    If $\{r_2, r_3, r_4\}$ are not all equal, note that
    \[
        \Delta_r \eq (0,\ r_3 - r_4,\ r_4 - r_2,\ r_2 - r_3)
    \]
    is a non-zero element of $\cC_r$ (and in fact spans the one-dimensional vector space $\cC_r$). If on the other hand $r_2 = r_3 = r_4$, the constraint $r^\top \Delta = 0$ reduces to $\mathbf{1}^\top \Delta = 0$. In particular the pair-swaps
    \[
        \Delta^{1}_r \eq (0, 1, -1, 0) \,, \qquad \Delta^{2}_r \eq (0, 0, 1, -1)
    \]
    are non-zero, linearly independent elements of $\cC_r$ (and in fact span the two-dimensional vector space $\cC_r$). In either case $\cC_r \neq \{0\}$, so the family $\{P_\ep\}_{\ep \Delta \in \cC_r}$ constructed below gives at least a one-parameter family of distinct stochastic kernels (two-parameter in the second case). Now pick any $0 \neq \Delta \in \cC_r$. Since $P(\cdot \mid s_1, a_1)$ is interior, there exists $\ep_0 > 0$ such that for every $\ep \in (-\ep_0, \ep_0)$,
    \[
        P_\ep(\cdot \mid s_1, a_1) \eq P(\cdot \mid s_1, a_1) + \ep \Delta
    \]
    is a valid transition function (non-negativity by interior, and $\mathbf{1}^\top \Delta = 0$ preserves the sum); define $P_\ep$ to agree with $P$ everywhere else. Now fix any policy $\pi$ and $\ga \in [0, 1)$. For $k \in \{2, 3, 4\}$, the Bellman equation at an absorbing state gives $V^\pi(s_k) = r_k/(1-\ga)$, depending on $r$ alone. For $k = 1$, writing $\al_l(\ep) \eq \sum_j \pi(a_j \mid s_1) P_\ep(s_l \mid s_1, a_j)$, the Bellman equation reads
    \[
        V^\pi(s_1) = \al_1(\ep) (r_1 + \ga V^\pi(s_1)) + \sum_{l=2}^4 \al_l(\ep) r_l/(1-\ga) \,,
    \]
    depending on the kernel only through $\al_1(\ep)$ and $\sum_{l=2}^4 \al_l(\ep) r_l$. Now $\Delta_1 = 0$ implies $\al_1(\ep) = \al_1(0)$ for all $\ep$, and since $r^\top \Delta = 0$,
    \[
        \sum_{l=2}^4 (\al_l(\ep) - \al_l(0)) r_l = \ep \pi(a_1 \mid s_1) r^\top \Delta = 0 \,,
    \]
    so $\sum_{l=2}^4 \al_l(\ep) r_l$ is also $\ep$-independent. Hence $V^\pi_{P_\ep}(s_1) = V^\pi_{P}(s_1)$. Now consider the $Q$-values
    \[
        Q^\pi_{P_\ep}(s, a; \ga) = \sum_{l=1}^4 P_\ep(s_l \mid s, a) [r_l + \ga V^\pi(s_l)] \,.
    \]
    The only entry dependent on $\ep$ is $(s_1, a_1)$, for which
    \[ Q^\pi_{P_\ep}(s_1,a_1) - Q^\pi_{P}(s_1,a_1) = \sum_{l=2}^4 \Delta_l r_l /(1-\ga) = \ep r^\top \Delta /(1-\ga) = \ep r^\top \Delta /(1-\ga) = 0 \,,
    \]
    hence $Q^\pi_{P_\ep} = Q^\pi_{P}$ for every $\ep \in (-\ep_0, \ep_0)$ and every $(\pi, \ga)$.
\end{lproof}

\subsection{Proof of \texorpdfstring{\cref{prop:local} (Local MDPs)}{Proposition (Local)}}\label{sec:proof-local}

As a middle-ground between deterministic and stochastic, we present a result referenced at the end of \cref{sec:finite} for \textit{$N$-local MDPs}, defined by $\supp_{s,a}(P) \eq |\{s' : P(s' \mid s, a) > 0\}| \leq N$ for all $(s, a)$. To restrict the usual hypothesis class $\cP = \{ P : \cS \times \cA \to \Delta(\cS) \}$, we introduce two classes corresponding to $N$-local MDPs with \textit{known} or \textit{unknown} support respectively
\begin{align*}
    \cP_\Gamma &\eq \{ P \in \cP : \supp_{s,a}(P) = \Gamma(s,a) \ \ \forall s,a \} \,, \\
    \cP_N &\eq \{ P \in \cP : |\supp_{s,a}(P)| \leq N \ \ \forall s,a \} \,,
\end{align*}
where $\Gamma : \cS \times \cA \to 2^\cS$ is a fixed support function with $|\Gamma(s,a)| \leq N$ for all $(s,a)$. For example, a deterministic MDP is a $1$-local MDP with typically unknown support (the successor state is not known), while an MDP with locally constrained dynamics may have known support if the agent has some prior knowledge about the environment. In the extreme case, a stochastic MDP is simply an $n$-local (i.e. global) MDP with known support (the entire state space).

\begin{linked}[Local]{theorem}{local}\label{prop:local}
    Fix any $N$-local finite MDP, policy $\pi$ and support $\Gamma$.
    \begin{enumerate}[label=(\alph*)]
        \item Known support: For any $|\cG| \geq N - 1$, the set of reward functions $r : \cS \times \cG \to \R$ for which $P \in \cP_{\Gamma}$ is uniquely determined by $Q^\pi$ has full Lebesgue measure.
        \item Unknown support: For any $|\cG| \geq 2N-1$, the set of reward functions $r : \cS \times \cG \to \R$ for which $P \in \cP_N$ is uniquely determined by $Q^\pi$ has full Lebesgue measure.
    \end{enumerate}
\end{linked}

\begin{lproof}
    Write $L = |\cG|$, and $\supp(i, j) \eq \{k \in [n] : P_{ijk} > 0\}$ for the support of $P(\cdot \mid s_i, a_j)$.

    \textbf{Part (a).} When $\supp(i, j) = \Gamma(s_i, a_j)$ is known, the Bellman equation (Equation \ref{eq:bellman-linear}) reduces to a linear system of size $\leq N$ per state-action pair. Taking $|\supp(i,j)| = N$ without loss of generality (the proof follows identically when the support is smaller), we have
    \[ M^\pi_{ij} \eq M^\pi_{:,\supp(i,j)} \in \R^{L \times N} \]
    and augmenting $M^\pi_{ij}$ and $Q_{ij}$ with ones to obtain $\mathbf{M}^\pi_{ij} \in \R^{(L+1)\times N}$ and $\mathbf{Q}_{ij} \in \R^{L+1}$, the transition function $P$ satisfies
    \begin{equation}\label{eq:local-bellman-augmented}
        \mathbf{M}^\pi_{ij} P_{ij} = \mathbf{Q}_{ij}
    \end{equation}
    for all $i,j$, where $P_{ij} \coloneqq P_{i,j,\supp(i,j)} \in \Delta^N$. In particular, $P \in \cP_{\Gamma}$ is uniquely determined if $\rank(\mathbf{M}^\pi_{ij}) \geq N$. The proof now follows similarly to that of \cref{prop:finite-deterministic-a}\textbf{(b)}. First recall the resolvent identity from \cref{lem:resolvent}:
    \[ M^\pi_{l} = R_l (I - \gamma \bP^{\pi_l})^{-1} \,, \]
    where $\bP^\pi = (P^\pi)^\tr$. Since $(I - \gamma \bP^{\pi_l})^{-1}$ is invertible for each $l$, the map $F : \R^{L \times n} \to \R^{L \times n}$ given by $F(R) = M^\pi$ is linear and invertible. Now $P \in \cP_{\Gamma}$ is uniquely determined by \cref{eq:local-bellman-augmented} if $\rank(\mathbf{M}^\pi_{ij}) \geq N$ for all $i,j$. Taking the converse statement,
    \begin{align*}
        \{ R \in \R^{L \times n} \mid \neg \exists! P(Q) \} &\subseteq \{ R \in \R^{L \times n} \mid \exists i,j \text{ s.t. } \rank(\mathbf{M}^\pi_{ij}) < N \} \\
        &= F^{-1}\bigg( \{ M \in \R^{L \times n} \mid \exists i,j \text{ s.t. } \rank(\mathbf{M}_{ij}) < N \} \Bigg) \\
        &= F^{-1}\bigg( \bigcup_{i,j} \{ M \in \R^{L \times n} \mid \rank(\mathbf{M}_{ij}) < N \} \bigg) \,.
    \end{align*}
    For generic $M \in \R^{L \times n}$, the augmented matrix $\mathbf{M} \in \R^{(L+1)\times n}$ is also generic, and so is any slice of columns $\mathbf{M}_{ij}$, with $\rank(\mathbf{M}_{ij}) = \min(L+1, N) = N$ almost surely, for all $L \geq N-1$. In particular,
    \[ \lambda \left( \{ M \in \R^{L \times n} \mid \rank(\mathbf{M}_{ij}) < N \} \right) = 0 \]
    for each $i,j$, and under measure-zero preservation through finite unions and $F^{-1}$,
    \[ \lambda \left( \{ R \in \R^{L \times n} \mid \neg \exists! P(Q) \} \right) = 0 \,. \]

    \textbf{Part (b).} $P \in \cP_N$ is uniquely determined if for each $i,j$, there are no $P_{ij} \neq P'_{ij} \in \Delta^n_N$ such that
    \[ \mathbf{M}^\pi P_{ij} = \mathbf{Q}_{ij} = \mathbf{M}^\pi P'_{ij} \,, \]
    Subtracting one side from the other, note that $P_{ij}-P'_{ij} \in \Lambda^n_{2N} \coloneqq \{ T \in \R^n \mid \|T\|_0 \leq 2N \}$, so $P \in \cP_N$ is uniquely determined if there is no $0 \neq T_{ij} \in \Lambda^n_{2N}$ such that
    \[ \mathbf{M}^\pi T_{ij} = 0 \,, \]
    noting that the last-row constraint $\sum_{k} T_{ijk} = 0$ is implied by $\sum_{k} P_{ijk} - P'_{ijk} = 0$. In particular, writing $\mathbf{M}_{:,K}$ for the matrix sliced to the columns indexed by $K \subset [n]$ with $|K| = 2N$, this is equivalent to $\mathbf{M}_{:,K} T_{ij} = 0$ having no non-trivial solutions for $T_{ij} \in \R^{2N}$ for all such $K$, which is implied by $\rank(\mathbf{M}_{:,K}) \geq 2N$. Now the map $F : \R^{L \times n} \to \R^{L \times n}$ given by $F(R) = M^\pi$ is linear and invertible, and $P \in \cP_N$ is uniquely determined if $\rank(\mathbf{M}^\pi_{:,K}) \geq 2N$ for all $K \subset [n]$ with $|K| = 2N$. Taking the converse statement,
    \begin{align*}
        \{ R \in \R^{L \times n} \mid \neg \exists! P(Q) \} &\subseteq \{ R \in \R^{L \times n} \mid \exists |K| = 2N \text{ s.t. } \rank(\mathbf{M}^\pi_{:,K}) < 2N \} \\
        &= F^{-1}\bigg( \{ M \in \R^{L \times n} \mid \exists |K| = 2N \text{ s.t. } \rank(\mathbf{M}_{:,K}) < 2N \} \bigg) \\
        &= F^{-1}\left( \ \bigcup_{|K| = 2N} \{ M \in \R^{L \times n} \mid \rank(\mathbf{M}_{:,K}) < 2N \} \right) \,.
    \end{align*}
    For generic $M \in \R^{L \times n}$, the augmented matrix $\mathbf{M} \in \R^{(L+1)\times n}$ is also generic, and so is any slice of columns $\mathbf{M}_{:,K}$, with $\rank(\mathbf{M}_{:,K}) = \min(L+1, 2N) = 2N$ almost surely, for all $L \geq 2N-1$. In particular,
    \[ \lambda \left( \{ M \in \R^{L \times n} \mid \rank(\mathbf{M}_{:,K}) < 2N \} \right) = 0 \]
    for each $K$, and under measure-zero preservation through finite unions and $F^{-1}$,
    \[ \lambda \left( \{ R \in \R^{L \times n} \mid \neg \exists! P(Q) \} \right) = 0 \,. \qedhere \]
\end{lproof}

\section{Proofs for Continuous MDPs}\label{sec:proofs:continuous}

This section contains all proofs for continuous MDPs; note that \cref{lem:per-goal-resolvent,lem:unconditional-determining} hold for arbitrary Borel state spaces $\cS$, but in order to specify concrete reward families (indicator and Gaussian) for which $\exists! P(Q^\pi)$ holds, we specialise the goal-augmented MDP definition of \cref{sec:background} to Borel \textit{continuous} state spaces $\cS \subseteq \R^d$. Formally, the condition on $P$ being a Markov kernel means that it satisfies two measurability conditions: (i) for each $(s,a)$, $B \mapsto P(B \mid s,a)$ is a Borel probability measure on $\cS$, and (ii) for each Borel set $B \in \Sigma_\cS$, $(s,a) \mapsto P(B \mid s,a)$ is measurable, ensuring that Bellman integrals $\int_\cS V(s') P(ds' \mid s,a)$ are well-defined. No absolute continuity of $P$ with respect to Lebesgue measure is assumed unless stated explicitly.

\subsection{Termination and the Bellman Family}\label{sec:proofs:termination}

In order to match the goal-dependent termination functions commonly used in GCRL (often written in code as \texttt{dones}), and thereby to formally state \cref{thm:deterministic-sparse,thm:deterministic-gaussian}, we generalise the goal-augmented MDP formalism (\cref{sec:background}) to include a \textit{termination function} $D : \cS \times \cG \to \{0, 1\}$ that specifies pairs $(s, g)$ beyond which rewards are set to zero. We use a similar formalism as the state-based discount factors $\gamma : \cS \to [0,1]$ used by \citet{Sutton2011} and \citet{White2017}, but with a goal-augmented domain. Writing $r_g(\cdot), D_g(\cdot) \eq r(\cdot, g), D(\cdot, g)$, The return of a trajectory $\tau = (s^t, a^t)_{t\geq 0}$ with respect to $g$ is now given by
\[ R(\tau, g) \eq \sum_{t\ge 0} \gamma^t r_g(s_{t+1}) \prod_{k=1}^{t}D_g(s_k) \,, \]
where the empty product ($t=0$) is defined as $1$, and which reduces to the usual return when $D \equiv 1$. Note that the multiplicative factor makes returns path-dependent, and could not be folded into the Markovian reward $r$. Terminating environments could instead be handled by adding aborbing states \citep{White2017}, but this imposes a modification on the state space, while transition-based termination does not, and cannot handle goal-dependence. With this in hand, the value and action-value functions are defined exactly as before: $V^\pi(s, g) \eq \E_{\tau \sim \pi_g} \left[ R(\tau, g) \mid s^0 = s \right]$ and $Q^\pi(s, a, g) \eq \E_{\tau \sim \pi_g} \left[ R(\tau, g) \mid s^0 = s, a^0 = a \right]$, with $\tau \sim \pi_g$ shorthand for independently sampling $s_0 \sim \mu, a_t \sim \pi(s_t, g), s_{t+1} \sim P(s_t, a_t)$.
In integral form, the Bellman equation becomes
\begin{equation*}\label{eq:bellman-continuous}
    Q^\pi_g(s,a) = \int_{\cS} \Big(r_g(s') + \gamma D_g(s')\, V^\pi_g(s')\Big) P(\mathrm{d} s' \mid s,a)\,.
\end{equation*}
While the product $\prod_{k=1}^t D_g(s_k)$ induces path-dependent returns, the Bellman operators inherit the usual $\gamma$-contraction properties, and their unique fixed point admits an optimal \textit{Markovian} policy \citep{White2017}.

\begin{definition}[Bellman family]\label{def:bellman-family}
    For any fixed $(Q^\pi, \pi, r, \gamma)$ with goal set $\cG$, we define the \textit{Bellman family} as the collection $\{m^{\pi}_{g} : g \in \cG\}$ of bounded Lebesgue-measurable functions
    \[
        m^{\pi}_{g}(s') \eq r_g(s') + \gamma D_g(s') V^\pi_g(s') \,, \quad \text{with} \quad V^\pi_g(s') =  \int_{\cA} Q^\pi_g(s',a') \pi(\mathrm{d} a' \mid s',g) \,.
    \]
    In particular, for any $(s,a)$, the Bellman equation
    \[
        \forall s,a,g.\quad Q^\pi(s,a,g) = \int m^{\pi}_{g}(s') P(\mathrm{d} s' \mid s,a)
    \]
    requires that action-values be a collection of expectations of the \textit{known} test functions $m^{\pi}_{g}$ under the \textit{unknown} measure $P(\cdot \mid s,a)$.
\end{definition}

\subsection{Determining Measures}\label{sec:proofs:continuous-family}

The two reward function families we consider for continuous MDPs are the following, adjoined with the notion of \textit{termination upon arrival}.

\begin{definition}\label{def:rewards-termination}
    Given $\cS, \cG \subseteq \R^d$, we define the Gaussian and indicator reward functions $r : \cS \times \cG \to \R$ as $r(s, g)=\exp(-\|s-g\|^2/(2\sigma^2))$ and $r(s,g) = \mathbf{1}[\norm{s - g} < \sigma]$ respectively, where $\sigma > 0$ is a fixed variance / threshold we omit from $r$ for brevity. For indicator rewards, we say that we assume \textit{termination upon arrival} if $D_g(s) = \mathbf{1}[\norm{s - g} \geq \sigma] = 1-r_g(s)$. For Gaussian goals, we say that we assume \textit{$\rho$-termination} if $D_g(s) = \mathbf{1}[\norm{s - g} \geq \rho]$, noting that the Gaussian variance $\sigma$ is not necessarily equal to the termination threshold $\rho$.
\end{definition}

To prove that $P$ can be identified uniquely from $(Q^\pi, \pi, r, \gamma)$, we need to show that the Bellman family $\{m^{\pi}_{g} : g \in \cG\}$ is \textit{measure determining}, defined as follows.

\begin{definition}[Measure-determining family]\label{def:measure-determining}
    A set $\mathcal{F}$ of bounded measurable functions $f:\cS\to \R$ is \emph{measure determining} if for any probability measures $\mu,\nu \in \Delta(\cS)$,
    \[
        \int f \mathrm{d}\mu = \int f \mathrm{d}\nu \quad \forall f\in\mathcal{F}
        \quad\implies\quad
        \mu=\nu.
    \]
\end{definition}

When policies are unconditional, Lemma \ref{lem:unconditional-determining} shows that Bellman families are measure-determining if the underlying reward family is signed-measure-determining, as defined below.

\begin{definition}[Signed-measure-determining family]\label{def:signed-measure-determining}
    A set $\mathcal{F}$ of bounded measurable functions $f:\cS\to\R$ is \emph{signed-measure-determining} if for any finite signed measures $\eta$ and $\tilde\eta$ on $\cS$,
    \[
        \int f \mathrm{d}\eta = \int f \mathrm{d}\tilde\eta \quad \forall f\in\mathcal{F}
        \quad\Longrightarrow\quad
        \eta=\tilde\eta.
    \]
\end{definition}

\cref{def:signed-measure-determining} is stronger than \cref{def:measure-determining}, since probability measures are finite signed measures.

\subsection{Proofs for Unconditional Policies (Stochastic \& Deterministic MDPs)}\label{sec:proofs:unconditional}

The results below apply to arbitrary stochastic continuous MDPs, which includes deterministic ones. However, they are restricted to unconditional policies $\pi_g = \pi_{g'}$ for all $g \in \cG$. The next section will deal with general (goal-conditioned) policies, but only apply to deterministic MDPs. Results for the challenging intersection of stochastic kernels and goal-conditioned policies is beyond the scope of this paper, and left for future work.

\subsubsection{Proof of \texorpdfstring{\cref{thm:gaussian-stochastic} (Gaussian Goals)}{Theorem (Gaussian Stochastic)}}\label{sec:stochastic-gaussian}

\begin{linked}{theorem}{gaussian-stochastic}\label{thm:gaussian-stochastic}
    Fix any continuous stochastic MDP, unconditional policy $\pi$ and termination function $D$, $\sigma > 0$, and a goal set $\cG \subseteq \R^d$ with non-empty interior and Gaussian rewards $r_g(s') = \exp(-\norm{s'-g}^2/(2\sigma^2))$. Then the Bellman family $\{m^\pi_{g}\}_{g \in \cG}$ is measure determining on $\cS$, and thus $\exists! P(Q^\pi)$. The conclusion also holds when $\cG$ is a countable set dense in some non-empty open subset of $\R^d$.
\end{linked}

\begin{lproof}
    We prove that $\{r_g : g \in \cG\}$ is determining for finite signed measures on $\cS$, which by \cref{lem:unconditional-determining} implies that $\{m^\pi_g\}$ is measure determining on $\cS$. Let $\eta$ be a finite signed measure on $\cS$ with $\int_\cS r_g \mathrm{d}\eta = 0$ for all $g \in \cG$. Writing $\varphi_\sigma(x) \coloneqq \exp(-\|x\|^2 / (2\sigma^2))$, note that $r_g(s') = \varphi_\sigma(g - s')$, so
    \[
        F(g) \coloneqq \int_\cS \varphi_\sigma(g - s')\, \eta(\mathrm{d}s') = (\eta * \varphi_\sigma)(g)
    \]
    vanishes on $\cG$ by hypothesis. By \cref{lem:gauss-conv-analytic}, $F$ is real-analytic on $\R^d$. Since $\cG \subseteq \R^d$ has non-empty interior in $\R^d$, $F$ vanishes on an open ball $U \subseteq \cG$, and the identity theorem for real-analytic functions on $\R^d$ \citep{KrantzParks2002} gives $F \equiv 0$ on $\R^d$. The Fourier transform argument derived in \cref{lem:gauss-fourier} implies $\eta = 0$. If $\cG$ is instead countable and dense in some non-empty open $U \subseteq \R^d$, then $F$ vanishes on a dense subset of $U$, and continuity (implied by real-analyticity) gives $F = 0$ on $U$, and the identity theorem applies as before.
\end{lproof}

\subsubsection{Proof of \texorpdfstring{\cref{thm:stochastic-sparse} (Indicator Goals)}{Theorem (Indicator Stochastic)}}\label{sec:proof-sparse-stochastic}

For indicator rewards, the goal set is required to contain the open ball of radius $\sigma$ around the state space. Recall that $B_\sigma(0)\coloneqq \{s \in \R^d \mid \lVert s \rVert<\sigma \}$, and $\cS \oplus B_\sigma(0)\coloneqq \{ s+x \mid s \in S, x \in B_\sigma(0)\}$

\begin{linked}{theorem}{stochastic-sparse}\label{thm:stochastic-sparse}
    Fix any continuous stochastic MDP, unconditional policy $\pi$ and termination function $D$, $\sigma > 0$, and goal set $\cG \supseteq \cS \oplus B_\sigma(0)$ with indicator rewards $r_g(s') = \mathbf{1}[\norm{s'-g} < \sigma]$. Then the Bellman family $\{m^\pi_g\}_{g \in \cG}$ is measure determining on $\cS$, and thus $\exists! P(Q^\pi)$.
\end{linked}

\begin{lproof}
    We prove that $\{r_g : g \in \cG\}$ is determining for finite signed measures on $\cS$, which by \cref{lem:unconditional-determining} implies that $\{m^\pi_g\}$ is measure determining on $\cS$. Let $\eta$ be a finite signed measure on $\cS$ with $\int_\cS r_g \mathrm{d}\eta = 0$ for all $g \in \cG$. Writing $\psi_\sigma(x) \coloneqq \mathbf{1}[\|x\| < \sigma]$, we have $r_g(s') = \psi_\sigma(g - s')$, so
    \[
        F(g) \coloneqq \int_\cS \psi_\sigma(g - s')\, \eta(\mathrm{d}s') = (\eta * \psi_\sigma)(g)
    \]
    vanishes on $\cG$ by hypothesis. For any $g \notin \cS \oplus B_\sigma(0)$, every $s' \in \cS$ satisfies $\|g - s'\| \geq \sigma$, so $\psi_\sigma(g - s') = 0$ and thus $F(g) = 0$. Combined with $F = 0$ on $\cG \supseteq \cS \oplus B_\sigma(0)$, we have $F \equiv 0$ on $\R^d$. The Fourier transform argument proven in \cref{lem:sparse-fourier} allows us to conclude $\eta = 0$.
\end{lproof}

\subsection{Proofs for Goal-Conditioned Policies (Deterministic MDPs)}

We now consider deterministic MDPs, where $P(\cdot \mid s, a) = \delta_{f(s,a)}(\cdot)$ for a successor function $f : \cS \times \cA \to \cS$ and the Dirac delta $\delta$. In this setting, measure determination reduces to a simpler condition called \textit{point separation}.

\begin{definition}[Point-separating goals]\label{def:point-separating}
    A family $\mathcal{F}$ of real-valued functions on $\cS$ is \emph{point separating} if for any $x\neq y$ in $\cS$ there exists $f\in\mathcal{F}$ such that $f(x)\neq f(y)$.
\end{definition}

In particular, if the MDP is deterministic and the family $\{m^\pi_{g}:g\in\cG\}$ is point separating on $\cS$, the successor function $f$ is uniquely identified by $(\pi,Q^\pi)$: any $f, f'$ satisfying the Bellman equation
\begin{equation}\label{eq:point-separation-lemma}
    m^\pi_{g} \big(f(s,a)\big)=Q^\pi(s,a,g)=m^\pi_{g} \big(f'(s,a)\big)\qquad \forall g, s, a
\end{equation}
must have $f(s,a) = f'(s,a)$ for every $(s,a)$.

\subsubsection{\texorpdfstring{Proof of \cref{thm:deterministic-sparse} (Deterministic MDPs, Indicator Goals with Termination)}{Theorem (Deterministic, Indicator)}}\label{sec:proof-deterministic-sparse}

\label{proof:deterministic-sparse}
\recall{theorem:deterministic-sparse}

\begin{lproof}
    \textbf{(i) Reward separation.} Fix any $x \neq y \in \cS$. We first exhibit $g \in \cG$ with $r_g(x) = 1$ and $r_g(y) = 0$. If $\norm{x - y} \geq \sigma$, taking $g = x \in \cS \cG$ gives $\norm{x - g} = 0 < \sigma$, hence $r_g(x) = 1$, while $\norm{y - g} = \norm{y - x} \geq \sigma$ implies $r_g(y) = 0$. Otherwise, let $\delta \eq \frac{1}{2}\norm{x - y} \in (0, \sigma/2)$ and define
    \[
        g \eq x + (\sigma - \delta) \frac{x - y}{\norm{x - y}} \,.
    \]
    Then $\norm{g - x} = \sigma - \delta < \sigma$, so $g \in B_\sigma(x) = x + B_\sigma(0) \subseteq \cS \oplus B_\sigma(0) \subseteq \cG$ and $r_g(x) = 1$. Writing $u \eq (x - y)/\norm{x - y}$, we have $g - y = (x - y) + (\sigma - \delta) u = (\norm{x - y} + \sigma - \delta) u$, so
    \[ \norm{g - y} = \norm{x - y} + \sigma - \delta = \tfrac{1}{2}\norm{x - y} + \sigma > \sigma \,, \]
    hence $r_g(y) = 0$. In either case, the family $\{r_g : g \in \cG\}$ separates $x$ from $y$.

    \textbf{(ii) Bellman family expansion.} Fix $g$ as constructed in step (i). By \cref{lem:per-goal-resolvent},
    \[ m_g^\pi = \sum_{k \geq 0} \ga^k T_g^k r_g \,, \quad \text{where} \quad (T_g h)(s) = D_g(s)\, \E_{s' \sim P^{\pi_g}(\cdot \mid s)}[h(s')] \,, \]
    is the per-goal Bellman operator. We claim that for all $k \geq 0$,
    \begin{equation}\label{eq:bellman-trajectory}
        (T_g^k r_g)(s) = \E_\tau \left[ \prod_{t < k} D_g(s_t)\, r_g(s_k) \,\middle|\, s_0 = s \right] \,,
    \end{equation}
    where $\tau = (s_t)_{t \geq 0}$ is the policy-induced trajectory with $a_t \sim \pi_g(\cdot \mid s_t)$ and $s_{t+1} = f(s_t, a_t)$. The base case $k = 0$ holds by definition of the empty product equalling 1. For the inductive step, we use the definition of $T_g$ and the inductive hypothesis applied to $s_1$ to get
    \begin{align*}
        (T_g^{k+1} r_g)(s) &= D_g(s) \E \left[(T_g^k r_g)(s_1) \,\middle|\, s_0 = s\right] \\
        &= D_g(s) \E\left[\E_{\tau'}\left[\prod_{t < k} D_g(s'_t)\, r_g(s'_k) \,\middle|\, s'_0 = s_1\right] \,\middle|\, s_0 = s\right] \\
        &= \E\left[D_g(s_0) \prod_{t < k} D_g(s_{t+1})\, r_g(s_{k+1}) \,\middle|\, s_0 = s\right]  \\
        &= \E\left[\prod_{t < k+1} D_g(s_t)\, r_g(s_{k+1}) \,\middle|\, s_0 = s\right] \,,
    \end{align*}
    where the third equality uses the Markov property (the inner trajectory $\tau'$ started at $s_1$ has the same law as $(s_{t+1})_{t \geq 0}$ under the original chain) to merge the nested expectations.

    \textbf{(iii) Bellman separation.} Since $r_g(x) = 1$, the assumption of termination upon arrival (\cref{def:rewards-termination}) gives $D_g(x) = 1-r_g(x) = 0$, so the product in Equation \ref{eq:bellman-trajectory} contains the factor $D_g(s_0) = 0$ for every $k \geq 1$, conditional on $s_0 = x$. Therefore $(T_g^k r_g)(x) = 0$ for all $k \geq 1$, and only the $k = 0$ term survives:
    \[ m_g^\pi(x) = r_g(x) = 1 \,. \]
    For $y$, we have $r_g(y) = 0$ and hence $D_g(y) = 1$. Conditioning on $s_0 = y$ throughout, we define the hitting time
    \[ T \eq \inf\{t \geq 1 : s_t \in B_\sigma(g)\} \in \{1, 2, \dots\} \cup \{\infty\} \,. \]
    Note first that the $k = 0$ term contributes $(T_g^0 r_g)(y) = r_g(y) = 0$. For $k \geq 1$, we evaluate $(T_g^k r_g)(y)$ by splitting on whether $T < k$, $T = k$, or $T > k$, recalling that $s_0 = y$ satisfies $\norm{y - g} > \sigma$, so $s_0 \notin B_\sigma(g)$ and $D_g(s_0) = 1$. On $\{T < k\}$, the index $T \in \{1, \dots, k-1\}$ falls inside the product $\prod_{t < k} D_g(s_t)$, and $s_T \in B_\sigma(g)$ gives $D_g(s_T) = 1 - r_g(s_T) = 0$, killing the product. On $\{T = k\}$, for every $1 \leq t < k$ we have $s_t \notin B_\sigma(g)$ by definition of $T$, so $D_g(s_t) = 1$; combined with $D_g(s_0) = 1$, the entire product equals $1$, while $s_k \in B_\sigma(g)$ gives $r_g(s_k) = 1$, so the integrand equals $1$. On $\{T > k\}$, $s_k \notin B_\sigma(g)$, so $r_g(s_k) = 0$ kills the integrand. Combining, the $\{0, 1\}$-valued integrand $\prod_{t < k} D_g(s_t)\, r_g(s_k)$ is precisely the indicator of $\{T = k\}$, so $(T_g^k r_g)(y) = \E_\tau[\mathbf{1}\{T = k\} \mid s_0 = y]$ for $k \geq 1$. Swapping sum and expectation since all terms are nonnegative (Fubini \citep{Folland1999}), and using $\sum_{k \geq 1} \ga^k \mathbf{1}\{T = k\} = \ga^T$ (with the convention $\ga^\infty = 0$),
    \[
        m_g^\pi(y) = \sum_{k \geq 1} \ga^k \E_\tau[\mathbf{1}\{T = k\} \mid s_0 = y] = \E_\tau\left[\ga^T \,\middle|\, s_0 = y\right] \leq \ga < 1 = m_g^\pi(x) \,,
    \]
    where the final bound uses $T \geq 1$. We conclude that $\{m_g^\pi : g \in \cG\}$ separates $x$ from $y$. Since $(x, y)$ was arbitrary, the Bellman family is point separating on $\cS$, and \eqref{eq:point-separation-lemma} implies that the $P$ is uniquely identified by $(\pi, Q^\pi, r, \gamma)$.
\end{lproof}

\subsubsection{Proof of \texorpdfstring{\cref{thm:deterministic-gaussian} (Deterministic MDPs, Gaussian Goals with Termination)}{Theorem (Deterministic, Gaussian)}}\label{sec:deterministic-gaussian}

We now turn to Gaussian goals $r_g(s) = \exp(-\norm{s - g}^2/(2\sigma^2))$ paired with \textit{$\rho$-termination}, i.e. a termination function $D_g(s) = \mathbf{1}[\norm{s - g} \geq \rho]$ (see \cref{def:rewards-termination}). Unlike indicator rewards, $D_g \neq 1 - r_g$, with the termination threshold $\rho$ being independent from the Gaussian variance $\sigma$.

\begin{linked}{theorem}{deterministic-gaussian}\label{thm:deterministic-gaussian}
    Fix any continuous deterministic MDP with compact $\cS \subseteq \R^d$, policy $\pi$ and $\sigma, \rho > 0$ satisfying the smallness condition
    \begin{equation*}
        \sigma^2 < \frac{\rho^2}{-4\log(1-\gamma)} \,.
    \end{equation*}
    Then there is a finite goal set $\cG \subseteq \cS \oplus B_\rho(0)$ with Gaussian rewards $r_g(s) = \exp(-\norm{s-g}^2/(2\sigma^2))$ and $\rho$-termination $D_g(s) = \mathbf{1}[\norm{s - g} \geq \rho]$ such that $\{m_g^\pi : g \in \cG\}$ is point separating on $\cS$, hence $\exists! P(Q^\pi)$. In particular, $\exists! P(Q^\pi)$ for any $\cG \supseteq \cS \oplus B_\rho(0)$.
\end{linked}

\begin{lproof}
    \textbf{(i) Constructing $\cG$.} First note that the smallness condition is equivalent to
    \begin{equation}\label{eq:gauss-smallness}
        \exp(-\rho^2/(2\sigma^2)) < (1 - \gamma)^2 \,, \quad \text{or equivalently,} \quad \sigma^2 < \frac{\rho^2}{-4\log(1-\gamma)} \;.
    \end{equation}
    Now pick $\eta \in (0, \rho/2)$ small enough that
    \begin{equation}\label{eq:gauss-eta-condition}
        \exp(-2\eta^2/\sigma^2) > \frac{\exp(-\rho^2/(2\sigma^2))}{1 - \gamma} + \gamma \,,
    \end{equation}
    which exists because the LHS tends to $1$ as $\eta \to 0$ while the RHS is smaller than $1$ by Equation \ref{eq:gauss-smallness}. By compactness, we can choose locii $c_1, \dots, c_N \in \cS$ such that $\cS \subseteq \bigcup_{i = 1}^N B_\eta(c_i)$. Let $e_1, \dots, e_d$ denote the standard basis of $\R^d$ and define, for each $i \in [N]$ and $j \in [d]$, the finite set of goals $\cG \eq \{g_{ij} \mid 1 \leq i \leq N,\, 0 \leq j \leq d\}$ with
    \[ g_{i0} \eq c_i \,, \quad \text{and} \quad g_{ij} \eq c_i + \eta e_j \quad \text{for $j \in [d]$} \,. \]
    Since $c_i \in \cS$ and $\norm{g_{ij} - c_i} = \norm{\eta e_j} = \eta < \rho$, we have $g_{ij} \in c_i + B_\rho(0)$ and thus $\cG \subseteq \cS \oplus B_\rho(0)$.

    \textbf{(ii) Bellman expansion.} Fix any $g \in \cG$. By \cref{lem:per-goal-resolvent}, we have $m_g^\pi = \sum_{k \geq 0} \gamma^k T_g^k r_g$ with $(T_g h)(s) = D_g(s) \E_{s' \sim P^{\pi_g}(\cdot \mid s)}[h(s')]$. The trajectory formula (Equation \ref{eq:bellman-trajectory}) in the proof of \cref{thm:deterministic-sparse} holds for any bounded measurable reward, so for all $k\geq 0$,
    \[ (T_g^k r_g)(s) = \E_\tau \left[ \; \prod_{t < k} D_g(s_t) r_g(s_k) \,\middle|\, s_0 = s\right] \,. \]

    \textbf{(iii) Bellman separation.} Fix distinct $x, y \in \cS$ and pick $i$ with $x \in B_\eta(c_i)$. For each $j \in \{0, \dots, d\}$,
    \[ \norm{x - g_{ij}} \leq \norm{x - c_i} + \norm{c_i - g_{ij}} < \eta + \eta = 2\eta < \rho \,, \]
    by assumption that $\eta \in (0,\rho/2)$. Hence $D_{g_{ij}}(x) = 0$ and the product in the trajectory formula above contains the factor $D_g(s_0) = 0$ for every $k \geq 1$ when $s_0 = x$. Only the $k = 0$ term survives:
    \begin{equation}\label{eq:gauss-separation-x}
        m_{g_{ij}}^\pi(x) = r_{g_{ij}}(x) = \exp(-\norm{x - g_{ij}}^2/(2\sigma^2)) > \exp(-2\eta^2/\sigma^2) \,.
    \end{equation}
    For $y$, we distinguish two cases.

    \emph{Case A: some $j$ satisfies $\norm{y - g_{ij}} \geq \rho$.} Fix any such $j$ and write $g \eq g_{ij}$. Conditioning on $s_0 = y$, define the hitting time
    \[ T \eq \inf\{t \geq 1 : s_t \in B_\rho(g)\} \in \{1, 2, \dots\} \cup \{\infty\} \,, \]
    which automatically satisfies $T \geq 1$ since $s_0 = y \notin B_\rho(g)$. We bound the pathwise integrand $\sum_{k \geq 0} \gamma^k \prod_{t < k} D_g(s_t)\, r_g(s_k)$ by case analysis on $k$ relative to $T$. For $0 \leq k < T$, every $s_t$ with $0 \leq t < k$ satisfies $s_t \notin B_\rho(g)$ (true at $t = 0$ since $\norm{y - g} \geq \rho$, and at $1 \leq t < k < T$ by definition of $T$), so $D_g(s_t) = 1$. On the other hand, $\norm{s_k - g} \geq \rho$ gives $r_g(s_k) \leq \exp(-\rho^2/(2\sigma^2))$. For $k = T$ (only when $T < \infty$), the product $\prod_{t < T} D_g(s_t) = 1$ as above and $r_g(s_T) \leq 1$, contributing $\gamma^T \leq \gamma$ since $T \geq 1$. For $k > T$, the product contains the factor $D_g(s_T) = 0$. Summing,
    \[
        \sum_{k \geq 0} \gamma^k \prod_{t < k} D_g(s_t)\, r_g(s_k) \leq \sum_{k = 0}^{T - 1} \gamma^k \exp(-a^2/(2\sigma^2)) + \mathbf{1}\{T < \infty\} \gamma^T \leq \frac{\exp(-a^2/(2\sigma^2))}{1 - \gamma} + \gamma \,.
    \]
    Taking expectations on both sides (Fubini--Tonelli \citep{Folland1999}, since all terms are nonnegative) and applying the trajectory formula above,
    \[ m_g^\pi(y) = \sum_{k \geq 0} \gamma^k (T_g^k r_g)(y) \leq \frac{\exp(-a^2/(2\sigma^2))}{1 - \gamma} + \gamma < \exp(-2\eta^2/\sigma^2) < m_g^\pi(x) \,, \]
    where the strict inequalities use Equations \ref{eq:gauss-eta-condition} and \ref{eq:gauss-separation-x} respectively. Hence $m_g^\pi(y) \neq m_g^\pi(x)$.

    \emph{Case B: $\norm{y - g_{ij}} < \rho$ for every $j = 0, \dots, d$.} Then $D_{g_{ij}}(y) = 0$ for all $j$ and the same $k = 0$ collapse gives $m_{g_{ij}}^\pi(y) = r_{g_{ij}}(y)$ for all $j$. Now assume for contradiction that $m_{g_{ij}}^\pi(x) = m_{g_{ij}}^\pi(y)$ for every $j$; then substituting $m_{g_{ij}}^\pi(x) = r_{g_{ij}}(x)$ by Equation \ref{eq:gauss-separation-x}, we have $\norm{x - g_{ij}}^2 = \norm{y - g_{ij}}^2$ for every $j$. Expanding squared norms and subtracting the $j = 0$ identity from the $j$-th gives $2\langle g_{ij} - g_{i0}, x - y\rangle = 0$, which reduces to $\eta\langle e_j, x - y\rangle = 0$ for $j = 1, \dots, d$, forcing $x = y$, a contradiction. Hence $m_{g_{ij}}^\pi(x) \neq m_{g_{ij}}^\pi(y)$ for some goal $g_{ij}$.

    Since $(x, y)$ was arbitrary, $\{m_g^\pi : g \in \cG\}$ is point separating on $\cS$, and \eqref{eq:point-separation-lemma} implies that $P$ is uniquely identified by $(\pi, Q^\pi, r, \gamma)$. The last claim is immediate: if a goal set $\cG'$ satisfies $\cG' \supseteq \cS \oplus B_\rho(0)$ then $\cG' \supseteq \cG$, hence $\{m_g^\pi : g \in \cG'\} \supseteq \{m_g^\pi : g \in \cG\}$, which inherits point separation.
\end{lproof}

\begin{remark}[Finite, but large!]\label{rem:gauss-explicit}
    For $\cS \subseteq [-R, R]^d$, a grid with linear spacing $\eta/\sqrt{d}$ along each axis covers $\cS$ by $\eta$-balls: each grid cell is a hypercube of side $\eta/\sqrt{d}$ and diagonal $\eta$, hence fits in an $\eta$-ball. This gives an upper bound
    \[ N \leq \big\lceil 2R\sqrt{d}/\eta \big\rceil^d \quad \implies \quad |\cG| \leq (d+1) \big\lceil 2R\sqrt{d}/\eta \big\rceil^d \,, \]
    which is $O(d\exp(d\log(d)))$ in the dimension. The construction, albeit finite, is therefore impractical, but serves to illustrate the stronger mathematical guarantees for Gaussian goals (finite goals) as opposed to indicator goals (infinite goals) in this setting.
\end{remark}

\subsubsection{Proof of \texorpdfstring{\cref{thm:deterministic-gaussian-generic} (Deterministic MDPs, Gaussian Goals with Analytic Regularity)}{Theorem (Deterministic, Gaussian, Unconditional)}}\label{sec:deterministic-gaussian-generic}

The construction of \cref{thm:deterministic-gaussian} relies on the smallness condition $\sig^2 < \rho^2/(-4\log(1-\ga))$ and a goal-ball termination function, producing a finite but exponentially-sized goal set (see \cref{rem:gauss-explicit}). Trading termination for real-analytic regularity on $D$ and $V^\pi$ yields a sharper guarantee: $L \geq 2d + 1$ Gaussian goals in \emph{generic position} suffice for point separation, with no constraint on $\sigma$.

\begin{linked}{theorem}{deterministic-gaussian-generic}\label{thm:deterministic-gaussian-generic}
    Fix any continuous deterministic MDP with open $\cS \subseteq \R^d$, unconditional policy $\pi$, $\sig > 0$, Gaussian rewards $r_g(s) = \exp(-\norm{s - g}^2/(2\sig^2))$, and assume that $V^\pi$ and $D$ are real-analytic on $\cS \times \R^d$ and $\cS$ respectively. Then for every $L \geq 2d + 1$, the set of goal tuples $\boldg = (g_1, \ldots, g_L) \in \cS^L$ for which $\{m_{g_l}^\pi\}_{l = 1}^L$ is point separating on $\cS$ has full Lebesgue measure in $\cS^L$. In particular, for Lebesgue-almost every $L$-element goal set $\cG = \{g_1, \ldots, g_L\} \subseteq \cS$, we have $\exists! P(Q^\pi)$ for all $L \geq 2d + 1$.
\end{linked}

\begin{lproof}
    \textbf{(i) Resolvent identity.} Following the proof of \cref{lem:unconditional-determining}, notice that the Markov operator $(Th)(s) \eq D(s) \int_\cS h(s') P^\pi(\mathrm{d}s' \mid s)$ is independent from $g$, and its adjoint $T^*$ on finite signed measures satisfies $\norm{T^*}_{\mathrm{TV}} \leq 1$ and $\int_\cS Th \mathrm{d}\eta = \int_\cS h \mathrm{d}(T^*\eta)$. For each $x \in \cS$, define
    \begin{equation}\label{eq:gen-mux}
        \mu_x  \eq (I - \ga T^*)^{-1} \del_x  = \sum_{t \geq 0} \ga^t (T^*)^t \del_x \,,
    \end{equation}
    a finite non-negative Borel measure on $\cS$. By construction, $\mu_x$ satisfies the resolvent identity
    \begin{equation}\label{eq:gen-resolvent-mu}
        \mu_x - \ga T^* \mu_x  = \del_x \,.
    \end{equation}
    Using $m_g^\pi = (I-\ga T)^{-1}r_g$ from \cref{lem:per-goal-resolvent} and following the proof of \cref{lem:unconditional-determining} gives
    \begin{align*}
        m_g^\pi(x) &= \int_\cS m_g^\pi \mathrm{d}\del_x = \sum_{t \geq 0} \ga^t \int_\cS (T^t r_g) \mathrm{d}\del_x \\
        &= \sum_{t \geq 0} \ga^t \int_\cS r_g \mathrm{d}((T^*)^t \del_x) \\
        &= \int_\cS r_g \mathrm{d}\Big(\sum_{t \geq 0} \ga^t (T^*)^t \del_x\Big) = \int_\cS r_g \mathrm{d}\mu_x = (\mu_x \ast \phi_\sig)(g) \,,
    \end{align*}
    where $\phi_\sig(u) \eq \exp(-\norm{u}^2/(2\sig^2))$ and $r_g(s') = \phi_\sig(g - s')$.

    \textbf{(ii) Non-vanishing slices.} For $x, y \in \cS$, we define the slice function $h_{x, y} : \R^d \to \R$ by
    \[ h_{x, y}(g) \eq m_g^\pi(x) - m_g^\pi(y) = ((\mu_x - \mu_y) \ast \phi_\sig)(g) \,, \]
    which is real-analytic on $\R^d$ by \cref{lem:gauss-conv-analytic} applied to the finite signed measure $\mu_x - \mu_y$. We prove that $h_{x, y} \not\equiv 0$ for all $x \neq y$ via the contrapositive: if $h_{x, y} \equiv 0$, then $(\mu_x - \mu_y) \ast \phi_\sig \equiv 0$, so $\mu_x = \mu_y$ by \cref{lem:gauss-fourier}, and applying Equation \ref{eq:gen-resolvent-mu} yields $\del_x = \del_y$, which implies $x = y$. Writing $\Lambda \eq \{(x, x) : x \in \cS\} \subseteq \cS^2$ for the diagonal, we conclude that $h_{x,y} \not\equiv 0$ for all $(x,y) \in \cS^2 \setminus \Lambda$.

    \textbf{(iii) The bad set.} In order to characterise the set of goal tuples that fail to induce point-separation, we define $\cU \eq (\cS^2 \setminus \Lambda) \times (\R^d)^L \subseteq \R^{2d + Ld}$ and the map $H : \cU \to \R^L$ by
    \begin{equation*}
        H\big(x, y, \boldg\big) \eq \big(h_{x, y}(g_1), \dots, h_{x, y}(g_L)\big) \,,
    \end{equation*}
    where $\boldg = (g_1, \ldots, g_L)$ is a tuple of $L$ goals, and let
    \[ E \eq H^{-1}(0) = \{(x, y, \boldg) \in \cU : h_{x, y}(g_l) = 0 \ \forall \ l \in [L]\} \,. \]
    Now notice that the Bellman family $\{m_{g_l}^\pi\}_{l = 1}^L$ fails to be point-separating on $\cS$ if and only if there exist $x \neq y \in \cS$ with $h_{x, y}(g_l) = 0$ for every $l$. Defining $\pi_2 : \cU \to (\R^d)^L$ as the coordinate projection $(x, y, \boldg) \mapsto \boldg$, this leads us to the ``bad set''
    \[ B \eq \pi_2(E) = \{\boldg \in (\R^d)^L : \exists \, x \neq y \ \ \text{such that} \ \ h_{x, y}(g_l) = 0 \ \forall \ l \in [L]\} \]
    of points $\boldg \in B$ that induce a ``bad'' Bellman family. Our goal is therefore to prove that $B$ has Lebesgue measure zero in $\R^{Ld}$. To do so, we will partition $E$ into a countable union based on the \textit{vanishing order} of partial derivatives.
    
    \textbf{(iv) Partitioning $E$.} For a multi-index $\beta \in \N^d$, we write $\partial_g^\beta$ to denote the corresponding partial derivative in the $\R^d$-variable $g$, as distinct from the bold tuple $\boldg$, with order $|\beta| \eq \sum_{i=1}^d \beta^i$. For $(x, y, \boldg) \in E$ and each $l \in [L]$, we define
    \[ \Omega_l(x,y,\boldg) \eq \inf\{|\beta| \geq 0 : \partial_g^\beta h_{x, y}(g_l) \neq 0\} \]
    as the corresponding \textit{vanishing order}. Now real-analyticity of $D$, $V^\pi$ and $r$ implies that $m_g^\pi(x) = r_g(x) + \ga D(x) V^\pi(x, g)$ is jointly real-analytic in $g$ and $x$, and in particular, $(x, y, g) \mapsto h_{x, y}(g)$ is real-analytic on the open set $(\cS^2 \setminus \Lambda) \times \R^d$. Since $h_{x,y} \not\equiv 0$ for $x \neq y$ by part \textbf{(ii)}, the identity theorem \citep{KrantzParks2002} implies that the Taylor series of $h_{x,y}$ at $g_l$ has at least one non-zero coefficient, so $\Omega_l(x,y,\boldg) < \infty$ for all $(x,y,\boldg) \in E$. Moreover, $\Omega_l(x,y,\boldg) \geq 1$ since $h_{x, y}(g_l) = 0$ by assumption that $(x,y, \boldg) \in E$. We thus have $\Omega_l \in \N$ (positive natural numbers), and can partition $E$ as a countable (disjoint) union
    \[ E = \bigcup_{\boldN \in \N^L} E_{\boldN} \,, \qquad E_\boldN \eq \{(x, y, \boldg) \in E : \Omega_l(x,y,\boldg) = N_l \ \forall \ l \in [L]\} \,. \]

    \textbf{(v) Local dimension.} Fix $\boldN$ and $p = (\bx, \by, \bboldg) \in E_\boldN$. By definition, for each $l \in [L]$, there is a multi-index $\beta_l \in \N^d$ with $|\beta_l| = N_l$ and $\partial_g^{\beta_l} h_{\bx, \by}(\bg_l) \neq 0$. Since $|\beta_l| \geq 1$, pick a coordinate $c_l \in [d]$ with $\beta_l^{c_l} \geq 1$, and let $\alpha_l \in \N^d$ be $\beta_l$ with its $c_l$-th entry decremented by $1$, so $|\alpha_l| = N_l - 1$ and
    \begin{equation}\label{eq:gen-beta-split}
        \partial_g^{\beta_l} = \partial_{g^{c_l}} \partial_g^{\alpha_l} \,,
    \end{equation}
    where $g^{c_l}$ denotes the $c_l$-th scalar coordinate of $g \in \R^d$. We now define the map
    \[ S_p : \cU \to \R^L \,, \qquad S_{p}\big(x, y, \boldg\big)_l \eq \partial_g^{\alpha_l} h_{x, y}(g_l) \,, \]
    and its zero set
    \[ Z_p \eq S_p^{-1}(0) = \{(x, y, \boldg) \in \cU : S_p(x, y, \boldg) = 0\} \,. \]
    The motivation for this construction is that $E_\boldN \subseteq Z_p\,$: any $(x, y, \boldg) \in E_\boldN$ satisfies $\Omega_l(x, y, \boldg) = N_l$ for every $l$, so for $|\alpha_l| = N_l - 1 < N_l$, we have
    \[ S_{p}(x, y, \boldg)_l = \partial_g^{\alpha_l} h_{x, y}(g_l) = 0 \quad \forall \ l \in [L] \,, \]
    hence $(x, y, \boldg) \in Z_p$. We now leverage this inclusion to show that $Z_p$ has the local structure of a real-analytic submanifold of $\cU$ in a neighbourhood of $p$. First, the map $S_p$ is real-analytic on $\cU$ since partial derivatives of real-analytic functions are real-analytic. To invoke the implicit function theorem, we split the coordinates of $\cU \subseteq \R^{2d + Ld}$ into $w \eq (g_1^{c_1}, \ldots, g_L^{c_L}) \in \R^L$, comprising the $L$ scalar coordinates of $\boldg$ singled out by the indices $c_1, \ldots, c_L$ chosen above, and the remaining $2d + L(d-1)$ coordinates $v$. This identifies $\cU$ with an open subset of $\R^{2d + L(d-1)} \times \R^L$ via the bijection $(x, y, \boldg) \leftrightarrow (v, w)$, and we write $p = (v_p, w_p)$ accordingly. The key is now to notice that the partial Jacobian $\mathrm{D}_w S_p(p) \in \R^{L \times L}$ is diagonal:
    \[
        \mathrm{D}_w S_p(p) = \begin{pmatrix} \partial_g^{\beta_1} h_{\bx, \by}(\bg_1) & \cdots & 0 \\ \vdots & \ddots & \vdots \\ 0 & \cdots & \partial_g^{\beta_L} h_{\bx, \by}(\bg_L) \end{pmatrix} \,,
    \]
    where the diagonal entries $\partial_{g^{c_l}} \partial_g^{\alpha_l} h_{\bx, \by}(\bg_l) = \partial_g^{\beta_l} h_{\bx, \by}(\bg_l)$ use \cref{eq:gen-beta-split}, and are non-zero by definition of $\beta_l$. In particular, $\mathrm{D}_w S_p(p)$ is invertible. Combined with $S_p(p) = 0$, the real-analytic implicit function theorem \citep{KrantzParks2002} provides open neighbourhoods $V_p \subseteq \R^{2d + L(d-1)}$ of $v_p$ and $W_p \subseteq \R^L$ of $w_p$, and a real-analytic function $G_p : V_p \to W_p$ such that $S_p(v, G_p(v)) = 0$ for all $v \in V_p$, and moreover these are the only zeros of $S_p$ in $V_p \times W_p$:
    \[ Z_p \cap (V_p \times W_p) = \{(v, G_p(v)) : v \in V_p\} \,. \]
    In other words, setting $U_p \eq V_p \times W_p$ and $\psi_p(v) \eq (v, G_p(v))$ for brevity, the zero set $Z_p$ is the image of the real-analytic map $\psi_p$ in a neighbourhood of $p$, namely,
    \[ Z_p \cap U_p = \psi_p(V_p) \,. \]
    Returning to our set $E_\boldN \subseteq Z_p$, we obtain the local inclusion $E_\boldN \cap U_p \subseteq \psi_p(V_p)$.

    \textbf{(vi) Measure-zero conclusion.} The open set $\cU \subseteq \R^{2d + Ld}$ is second-countable, so the open cover $E_\boldN \subseteq \bigcup_{p \in E_\boldN} U_p$ admits a countable subcover $E_\boldN \subseteq \bigcup_{n \in \N} U_{p_n}$, and the local inclusion $E_\boldN \cap U_p \subseteq \psi_p(V_p)$ from part \textbf{(v)} gives
    \[ E_\boldN = E_\boldN \cap \bigcup_{n \in \N} U_{p_n} = \bigcup_{n \in \N} E_\boldN \cap  U_{p_n} \subseteq \bigcup_n \psi_{p_n}(V_{p_n}) \,. \]
    At last, we invoke the assumption $L \geq 2d + 1$ to get
    \[ 2d + L(d - 1) = 2d + Ld - L \leq 2d + Ld - (2d+1)  < Ld \,, \]
    so the parameterising domain $V_{p_n} \subseteq \R^k$ with $k \eq 2d + L(d-1)$ has dimension strictly less than $Ld$. In particular, since projection is also real-analytic, $\pi_2(\psi_{p_n}(V_{p_n}))$ is the image of the real-analytic map $\pi_2 \circ \psi_{p_n} : V_{p_n} \to \R^{Ld}$, from an open subset of $\R^k$ with $k < Ld$. By Sard's theorem \cite{Lee2013}, or a direct proof covering $V_{p_n}$ with countably many cubes, it immediately follows that $\la_{Ld}(\pi_2(\psi_{p_n}(V_{p_n}))) = 0$ for every $n$. Moreover, since $\pi_2$ is monotonic and preserves unions, $\pi_2(E_\boldN) \subseteq \bigcup_n \pi_2(\psi_{p_n}(V_{p_n}))$. We sidestep a direct proof that $\pi_2(E_\boldN)$ is Lebesgue measurable  by working with the \emph{Lebesgue outer measure} $\la_{Ld}^*$, which is defined everywhere, agrees with the Lebesgue measure on measurable sets, and satisfies countable subadditivity \citep{Folland1999}. Using this property, have
    \[ \la_{Ld}^*(\pi_2(E_\boldN)) \leq \sum_{n \in \N} \la_{Ld}^*(\pi_2(\psi_{p_n}(V_{p_n}))) = 0 \,. \]
    Using union preservation once more, the bad set is given by $B = \pi_2(E) = \bigcup_{\boldN \in \N^L} \pi_2(E_\boldN)$, hence
    \[ \la_{Ld}^*(B) \leq \sum_{\boldN \in \N^L} \la_{Ld}^*(\pi_2(E_\boldN)) = 0 \,. \]
    Completeness of Lebesgue measure \citep{Folland1999} guarantees that every set with Lebesgue outer measure zero is Lebesgue-measurable with measure zero, so we conclude $\la_{Ld}(B) = 0$. Now recall that a Bellman family $\{m_{g_l}^\pi\}_{l = 1}^L$ is point-separating on $\cS$ iff $\boldg \notin B$. Hence the goal tuples in $\cS^L$ that fail to induce point-separation are exactly $B \cap \cS^L$, of $\la_{Ld}$-measure zero, and the complement has full Lebesgue measure in $\cS^L$. In particular, $\la_{Ld}$-almost every $L$-element goal set $\cG = \{g_1, \ldots, g_L\} \subseteq \cS$ satisfies $\exists! P(Q^\pi)$ since the diagonal $\bigcup_{i \neq j}\{g_i = g_j\} \subseteq \cS^L$ has $\la_{Ld}$-measure zero, so $\la_{Ld}$-almost every tuple has distinct entries and corresponds to an $L$-element goal set.
\end{lproof}

\subsection{Supporting Results}\label{sec:proofs:gaussian:supporting}

\begin{lemma}[Resolvent]\label{lem:per-goal-resolvent}
    Let $\gamma \in [0,1)$, a measurable policy $\pi$, measurable termination function $D$ and bounded measurable reward function $r$. For any goal $g$, let $T_g$ be the Markov operator
    \[
        (T_g h)(s) \coloneqq D_g(s) \int_\cS h(s') P^{\pi_g}(\mathrm{d}s' \mid s) \,, \quad \text{where} \quad P^\pi_g(\cdot\mid s) \coloneqq \int_{\cA} P(\cdot\mid s,a) \pi_g(\mathrm{d}a\mid s) \,.
    \]
    Then for each $g \in \cG$, $m_g^\pi = (I - \ga T_g)^{-1} r_g = \sum_{t \geq 0} \ga^t T_g^t r_g$ as bounded measurable functions on $\cS$.
\end{lemma}

\begin{lproof}
    We note that the identity $m_g^\pi = (I-\gamma T^\pi)^{-1} r_g$ is essentially identical to \citet[Theorem~2]{Blier2021}, but with a goal-conditioned policy baked in; we include the proof for completeness. First recall the Bellman equations
    \[ V^\pi_g(s) = \int_\cA Q^\pi_g(s, a) \pi(\mathrm{d}a \mid s, g) \qquad \text{and} \qquad Q^\pi_g(s, a) = \int_\cS m_g^\pi(s') P(\mathrm{d}s' \mid s, a) \]
    from \cref{def:bellman-family}. Substituting, marginalising over $\pi(\cdot \mid s, g)$ and multiplying by $D_g(s)$:
    \[
        D_g(s) V^\pi_g(s) = D_g(s) \int_\cS m_g^\pi(s') P^{\pi_g}(\mathrm{d}s' \mid s) = (T_g m_g^\pi)(s) \,.
    \]
    Now $m_g^\pi(s) = r_g(s) + \ga D_g(s) V^\pi(s, g) = r_g + \ga T_g m_g^\pi$, hence $(I - \ga T_g) m_g^\pi = r_g$. Since $0 \leq D \leq 1$ and $P^{\pi_g}(\cS \mid s) = 1$, we have $\norm{T_g h}_\infty \leq \norm{D}_\infty \cdot \norm{h}_\infty \leq \norm{h}_\infty$, so $\norm{T_g}_{\infty \to \infty} \leq 1$. For any $\ga < 1$, this implies that $I - \ga T_g$ is invertible on bounded measurable functions, with Neumann series $(I - \ga T_g)^{-1} = \sum_{t \geq 0} \ga^t T_g^t$. Boundedness of $r_g$ ensures $m_g^\pi$ is bounded.
\end{lproof}

\begin{lemma}[Determining rewards $\Rightarrow$ determining Bellman family]\label{lem:unconditional-determining}
     Assume $\pi(\cdot\mid s,g) = \pi(\cdot\mid s)$ and $D(s, g) = D(s)$ are unconditional. If the family $\{r_g : g\in\cG\}$ is signed-measure-determining (\cref{def:signed-measure-determining}), then the Bellman family $\{m_g^\pi : g\in\cG\}$ is measure determining (\cref{def:measure-determining}).
\end{lemma}

\begin{lproof}
    By unconditionality, the Markov operator $(T_gh)(s) \eq D_g(s) \int_{\cS} h(s')P^{\pi_g}(\mathrm{d}s'\mid s)$ from \cref{lem:per-goal-resolvent} is independent from $g$, so we write $T \eq T_g$. Now let $\mu,\nu\in\Delta(\cS)$ satisfy $\int m_g^\pi \mathrm{d}\mu=\int m_g^\pi \mathrm{d}\nu$ for all $g$, and write $\eta\coloneqq \mu-\nu$. Define the adjoint $T^*$ on finite signed measures by
    \[ (T^*\eta)(B) \coloneqq \int_\cS D(s) P^\pi(B\mid s) \eta(\mathrm{d}s) \]
    for any Borel set $B \subseteq \cS$. By Fubini's theorem \citep{Folland1999}, $T^*\eta$ is a finite signed measure satisfying, for any bounded measurable function $h$, the following adjoint relation:
    \begin{align*}
        \int_{\cS} (T h) \mathrm{d}\eta &\eq \int_{\cS} \int_{\cS} D(s)  h(s') P^\pi(\mathrm{d} s'\mid s) \eta(\mathrm{d}s) \\
        &= \int_{\cS} h(s') \int_{\cS} D(s) P^\pi(\mathrm{d} s'\mid s) \eta(\mathrm{d}s) = \int_{\cS} h(s') T^*\eta(\mathrm{d}s') \qe \int_{\cS} h \mathrm{d}(T^*\eta) \,.
    \end{align*}
    Moreover, iterating this relation gives $\int (T^t h) \mathrm{d}\eta = \int h \mathrm{d}(T^t\eta)$ for all $t \in \N$. Now $\|T^*\eta\|_{\mathrm{TV}} \le \|\eta\|_{\mathrm{TV}}$ since $0 \leq D \leq 1$ and $P^\pi(\cS\mid s)=1$, which implies $\gamma\|T^*\|_{\mathrm{TV}} < 1$, so the resolvent
    \[\tilde\eta \coloneqq (I - \gamma T^*)^{-1}\eta = \sum_{t \geq 0} \gamma^t (T^*)^t\eta\]
    converges in total variation. Now, by \cref{lem:per-goal-resolvent}, $m_g^\pi = \sum_{t \geq 0} \gamma^t T^t r_g$. Interchanging sum and integral (by dominated convergence), applying the adjoint relation and summing the Neumann series gives
    \begin{align*}
        \int_\cS m_g^\pi \mathrm{d}\eta &= \sum_{t \geq 0} \gamma^t \int_\cS  (T^t r_g) \mathrm{d}\eta \\
        &= \sum_{t \geq 0} \gamma^t \int_\cS r_g \mathrm{d} ((T^*)^t\eta) \\
        &= \int_\cS r_g \mathrm{d} \Big( \sum_{t \geq 0} \gamma^t (T^*)^t\eta \Big) \qe \int_\cS r_g \mathrm{d}\tilde\eta  \,.
    \end{align*}
    Notice the structural reduction from the Bellman family to the reward family: the LHS integrates $m_g^\pi$ against the \textit{original} measure $\eta$, while the RHS integrates $r_g$ against the \textit{resolvent} measure $\tilde\eta = (I - \gamma T^*)^{-1}\eta$. To conclude the proof: by hypothesis, $\int_\cS m_g^\pi \mathrm{d}\eta$ vanishes for all $g$, so $\int r_g \mathrm{d}\tilde\eta = 0$ for all $g$. Since $\{r_g\}$ is determining for finite signed measures, $\tilde\eta = 0$. This immediately implies $\eta = (I - \gamma T^*)\tilde\eta = 0$, so we obtain $\mu = \nu$ as required.
\end{lproof}

\begin{lemma}\label{lem:gauss-conv-analytic}
    Let $\mu$ be a finite signed measure on $\R^d$ and $\varphi_\sigma(x) \coloneqq \exp(-\|x\|^2/(2\sigma^2))$ with $\sigma > 0$. Then $g \mapsto (\mu * \varphi_\sigma)(g)$ is real-analytic on $\R^d$.
\end{lemma}

\begin{lproof}
    Write $(\mu * \varphi_\sigma)(g) = \int_{\R^d} \varphi_\sigma(g - s')\, \mu(\mathrm{d}s')$. For each $s' \in \R^d$, the map
    \[ z \mapsto \varphi_\sigma(z - s') = \exp\left(-\frac{1}{2\sigma^2} \sum_{k=1}^d (z_k - s'_k)^2 \right) \]
    is the composition of polynomials and the exponential function, which are entire (i.e. holomorphic on $\C^d$), so $z \mapsto \varphi_\sigma(z - s')$ is entire for all $s' \in \R^d$. Now for any $z = x + iy \in \C^d$ with $x, y \in \R^d$ and $\|y\|_\infty \leq \sigma$, we have
    \[ \operatorname{Re} \sum_{k=1}^d (z_k - s'_k)^2 = \sum_{k=1}^d (x_k - s'_k)^2 - \sum_{k=1}^d y_k^2 \geq - \sum_{k=1}^d y_k^2 \geq -d\sigma^2 \,, \]
    so $|\varphi_\sigma(z - s')| \leq e^{d/2}$ uniformly in $s' \in \R^d$ and $z \in \{z \mid \|\operatorname{Im} z\|_\infty \leq \sigma\}$. The polydisc of radius $\sigma$ centred at any $g \in \R^d$, $\{z \mid |z_k - g_k| \leq \sigma \text{ for all } k\}$, lies entirely in the strip $\{z \mid \|\operatorname{Im} z\|_\infty \leq \sigma\}$, so by Cauchy's estimate on polydiscs \citep{KrantzParks2002},
    \[
        \sup_{s' \in \R^d} \bigl|\partial_g^\alpha \varphi_\sigma(g - s')\bigr| \leq \frac{\alpha!}{\sigma^{|\alpha|}} e^{d/2}
    \]
    for every multi-index $\alpha \in \N^d$. Since this bound is uniform in $s'$ and $|\mu|(\R^d) < \infty$, dominated convergence justifies differentiating under the integral, giving
    \[
        \bigl|\partial_g^\alpha (\mu * \varphi_\sigma)(g) \bigr| \leq \frac{\alpha!}{\sigma^{|\alpha|}} e^{d/2}\, |\mu|(\R^d) \,.
    \]
    A smooth function whose derivatives satisfy $|\partial^\alpha f| \leq A B^{|\alpha|} \alpha!$ uniformly on $\R^d$ has a convergent Taylor series at every point, hence is real-analytic on $\R^d$ \citep{KrantzParks2002}.
\end{lproof}

\begin{lemma}\label{lem:gauss-fourier}
    Let $\eta$ be a finite signed measure on $\R^d$ and $\varphi_\sigma(x) \coloneqq \exp(-\|x\|^2/(2\sigma^2))$ with $\sigma > 0$. If $(\eta * \varphi_\sigma)(g) = 0$ for all $g \in \R^d$, then $\eta = 0$.
\end{lemma}

\begin{lproof}
    Taking Fourier transforms
    \[ \widehat\eta(\xi)\coloneqq\int_{\R^d} e^{-2\pi i\xi\cdot s'}\eta(\mathrm{d}s') \quad \text{and} \quad \widehat{\varphi_\sigma}(\xi)=\int_{\R^d} e^{-2\pi i\xi\cdot x} \varphi_\sigma(x) \mathrm{d}x \,, \]
    the convolution theorem \citep{Folland1999} gives $\widehat\eta\cdot\widehat{\varphi_\sigma} = \widehat{\eta*\varphi_\sigma} = \widehat{0} = 0$ on $\R^d$. Now
    \[ \widehat{\varphi_\sigma}(\xi)=(2\pi\sigma^2)^{d/2}e^{-2\pi^2\sigma^2\|\xi\|^2}>0 \]
    for all $\xi$, so we conclude $\widehat\eta = 0$ everywhere. Using the Jordan decomposition $\eta = \eta^+ - \eta^-$ into finite positive Borel measures $\eta^\pm$ on $\R^d$, linearity of the Fourier transform gives $\widehat{\eta^+} = \widehat{\eta^-}$ on $\R^d$, so Bochner's theorem \citep{Bochner1955} implies $\eta^+ = \eta^-$, and thus $\eta = \eta^+ - \eta^- = 0$.
\end{lproof}

\begin{lemma}\label{lem:sparse-fourier-zeros}
    For $\sigma > 0$ and $\psi_\sigma(x) \coloneqq \mathbf{1}[\|x\| < \sigma]$ on $\R^d$, the Fourier transform of $\psi_\sigma$ is
    \[
        \widehat{\psi_\sigma}(\xi) = \sigma^{d/2}\, \frac{J_{d/2}(2\pi \sigma \|\xi\|)}{\|\xi\|^{d/2}} \,,
    \]
    where $J_\nu$ denotes the Bessel function of the first kind, and the zero set $\{\xi \in \R^d : \widehat{\psi_\sigma}(\xi) = 0\}$ has Lebesgue measure zero.
\end{lemma}

\begin{lproof}
    Since $\psi_\sigma$ is radial, i.e. depends only on $\|x\|$, the Fourier transform $\widehat{\psi_\sigma}$ depends only on $\|\xi\|$. Writing $r = \|\xi\|$, we use the radial Fourier formula \citep{SteinWeiss1971}
    \[
        \widehat{\psi_\sigma}(\xi) = 2\pi r^{-(d/2-1)} \int_0^\sigma J_{d/2-1}(2\pi r t) t^{d/2} \mathrm{d}t \,.
    \]
    Substituting $u = 2\pi r t$ gives $\mathrm{d}t = \mathrm{d}u/(2\pi r)$, with $t \in [0, \sigma]$ becoming $u \in [0, 2\pi r \sigma]$, and applying the Bessel identity $\int_0^T u^{\nu+1} J_\nu(u)\, \mathrm{d}u = T^{\nu+1} J_{\nu+1}(T)$ with $\nu = d/2 - 1$ and $T = 2\pi r \sigma$:
    \begin{align*}
        \widehat{\psi_\sigma}(\xi)
        &= \frac{2\pi}{r^{d/2-1}} \times \frac{1}{(2\pi r)^{d/2+1}} \int_0^{2\pi r \sigma} u^{d/2} J_{d/2-1}(u)\, \mathrm{d}u \\
        &= \frac{2\pi}{r^{d/2-1}} \times \frac{(2\pi r \sigma)^{d/2} J_{d/2}(2\pi r \sigma)}{(2\pi r)^{d/2+1}}
        = \sigma^{d/2} \frac{J_{d/2}(2\pi \sigma r)}{r^{d/2}} \,,
    \end{align*}
    which settles the first claim. First notice that $\xi = 0$ is not a zero of $\widehat{\psi_\sigma}$, since $J_{d/2}(z)/z^{d/2}$ converges to $1/(2^{d/2}\Gamma(d/2 + 1)) \neq 0$ as $z \to 0$.\footnote{This is irrelevant for the conclusion of Lebesgue measure zero, but gives a neater characterisation of the zero set.} For $\xi \neq 0$, the Bessel function $J_{d/2}$ has a sequence of strictly increasing positive zeros $\{a_{d/2,k}\}_{k \in \N}$ with no finite accumulation point. In particular, $\widehat{\psi_\sigma}(\xi) = 0$ iff $J_{d/2}(2\pi \sigma r) = 0$, i.e. $r = a_{d/2,k}/(2\pi \sigma)$ for some $k \geq 1$. Hence $Z$ is the countable union
    \[ \bigcup_{k \geq 1} \{\xi : \|\xi\| = a_{d/2,k}/(2\pi \sigma) \} \,, \]
    where each set is a $(d-1)$-dimensional sphere in $\R^d$, which has Lebesgue measure zero. Countable unions of measure-zero sets have zero measure, so we are done.
\end{lproof}

\begin{lemma}\label{lem:sparse-fourier}
    Let $\eta$ be a finite signed measure on $\R^d$ and $\psi_\sigma(x) \coloneqq \mathbf{1}[\|x\| < \sigma]$ with $\sigma > 0$. If $(\eta * \psi_\sigma)(g) = 0$ for all $g \in \R^d$, then $\eta = 0$.
\end{lemma}

\begin{lproof}
    Taking Fourier transforms, the convolution theorem \citep{Folland1999} gives $\widehat{\eta} \cdot \widehat{\psi_\sigma} = \widehat{\eta * \psi_\sigma} = \widehat{0} = 0$ on $\R^d$. Letting $Z = \{\xi : \widehat{\psi_\sigma}(\xi) = 0\}$, it follows that $\widehat{\eta}(\xi) = 0$ for all $\xi \in \R^d \setminus Z$. We claim that $\widehat{\eta}$ is continuous: if $\xi_n \to \xi$ in $\R^d$, then $e^{-2\pi i \xi_n \cdot s'} \to e^{-2\pi i \xi \cdot s'}$ pointwise, while $|e^{-2\pi i \xi_n \cdot s'}| = 1$ is integrable with respect to the finite measure $|\eta|$, so dominated convergence gives $\widehat{\eta}(\xi_n) \to \widehat{\eta}(\xi)$. By \cref{lem:sparse-fourier-zeros}, $Z$ has Lebesgue measure zero, so $\R^d \setminus Z$ is dense in $\R^d$ (every open ball has positive Lebesgue measure, so meets $\R^d \setminus Z$). Continuity of $\widehat{\eta}$ combined with vanishing on the dense set $\R^d \setminus Z$ gives $\widehat{\eta} \equiv 0$ on $\R^d$. As in the proof of \cref{lem:gauss-fourier}, we conclude $\eta = 0$.
\end{lproof}
\section{Reacher and MountainCar Experiments}\label{sec:appendix-experiments}

In this section we provide additional details for the continuous-control experiments of \cref{sec:experiments}. We train goal-conditioned $Q$-functions with PQN \citep{Gallici2025} and HER \citep{Andrychowicz2017}, then fit a world model $P_\phi$ to the converged $Q$-values via $P$-learning (\cref{sec:method}). Throughout, the WM is parameterised as a residual MLP $P_\phi(s,a) = s + h_\phi(s,a)$, trained with $\ell_1$ Bellman residual loss function on independently sampled $(s,a,g)$ batches. The sampling distribution is given by $d = \mu_\cS \otimes \mathrm{Unif}(\cA) \otimes \mathrm{Unif}(\cG)$, with $\mu_\cS$ and all hyperparameters specified per-environment in \cref{tab:hparams-reacher} and \cref{tab:hparams-mountaincar}. To train policies on unseen goals using the WM, we run value iteration on $\hat P$ over a discretised grid (resolution $50$ for \texttt{Reacher}, $200$ for \texttt{MountainCar}) to obtain a WM-derived policy.

\paragraph{Compute.}\label{par:compute}
All experiments were run on a single H100 GPU, leveraging vectorised environments implemented by Gymnax \citep{Gymnax2022} in JAX \citep{bradbury_jax_2018}. PQN training takes < 5min per \texttt{Reacher} run and < 2min per \texttt{MountainCar} or \texttt{FourRooms} run; $P$-learning takes < 10s in all cases. The largest single experiment is the $42$-architecture \texttt{Reacher} sweep (\cref{sec:appendix-reacher-arch}), totalling $\sim$35 GPU-hours.
\subsection{Reacher}\label{sec:appendix-reacher}

The \texttt{Reacher} state $(\theta_1, \theta_2, \omega_1, \omega_2, x, y) \in \R^6$ comprises the two joint angles $\theta_i$, their angular velocities $\omega_i$, and the fingertip position $(x,y) = (\cos\theta_1 + \cos\theta_2, \sin\theta_1 + \sin\theta_2)$. Since the fingertip is a deterministic function of the joints, training goals only specify $(x,y)$ and the world model is restricted to the dynamic dimensions $(\theta_1, \theta_2, \omega_1, \omega_2)$. The state distribution $\mu_\cS$ used by $P$-learning is the env reset distribution, $\theta_i \sim U[-\pi, \pi]$ and $\omega_i \sim U[-1, 1]$. The 2D action space is discretised into 9 actions (3 per joint) in order to train agents with PQN. Full hyperparameters in \cref{tab:hparams-reacher}.

\subsection{Reacher architecture sweep}\label{sec:appendix-reacher-arch}

\begin{figure}[ht]
    \centering
    \begin{subfigure}[t]{0.49\textwidth}
        \centering
        \includegraphics[width=\linewidth]{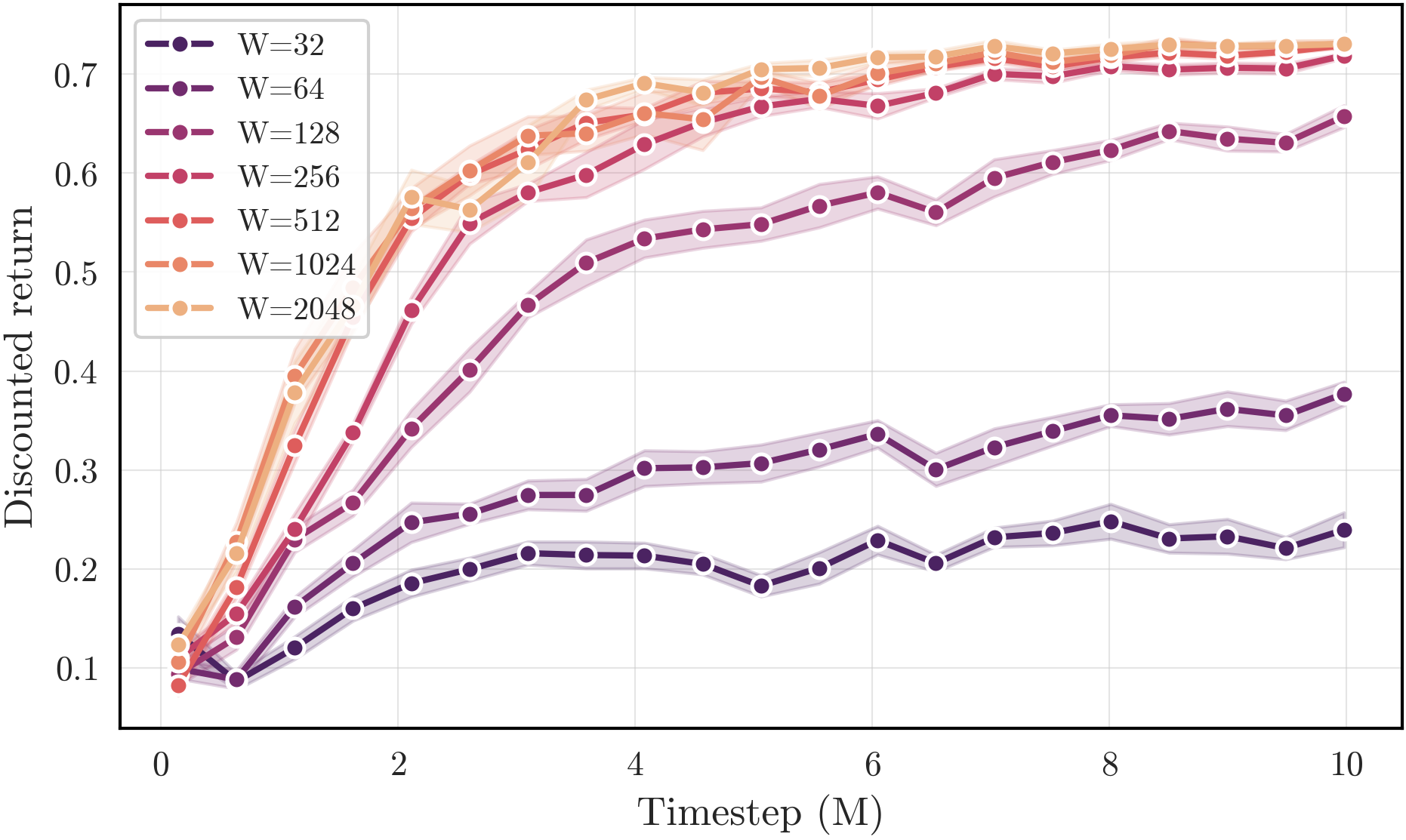}
        \caption{PQN return, slice at $D = 4$ (one curve per $W$).}
    \end{subfigure}
    \hfill
    \begin{subfigure}[t]{0.49\textwidth}
        \centering
        \includegraphics[width=\linewidth]{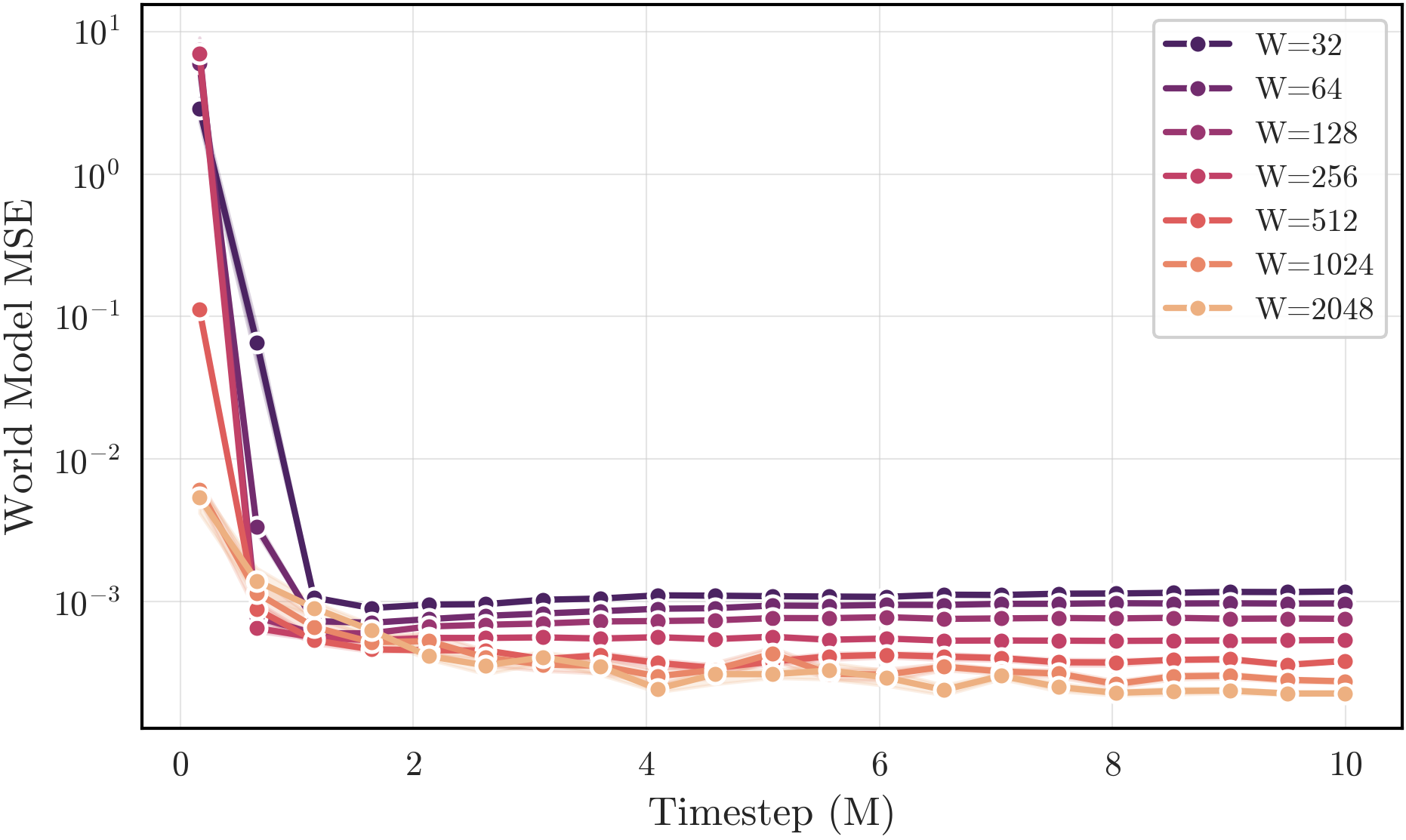}
        \caption{WM dynamics MSE (log-y), slice at $D = 4$.}
    \end{subfigure}\\[0.3em]
    \begin{subfigure}[t]{0.49\textwidth}
        \centering
        \includegraphics[width=\linewidth]{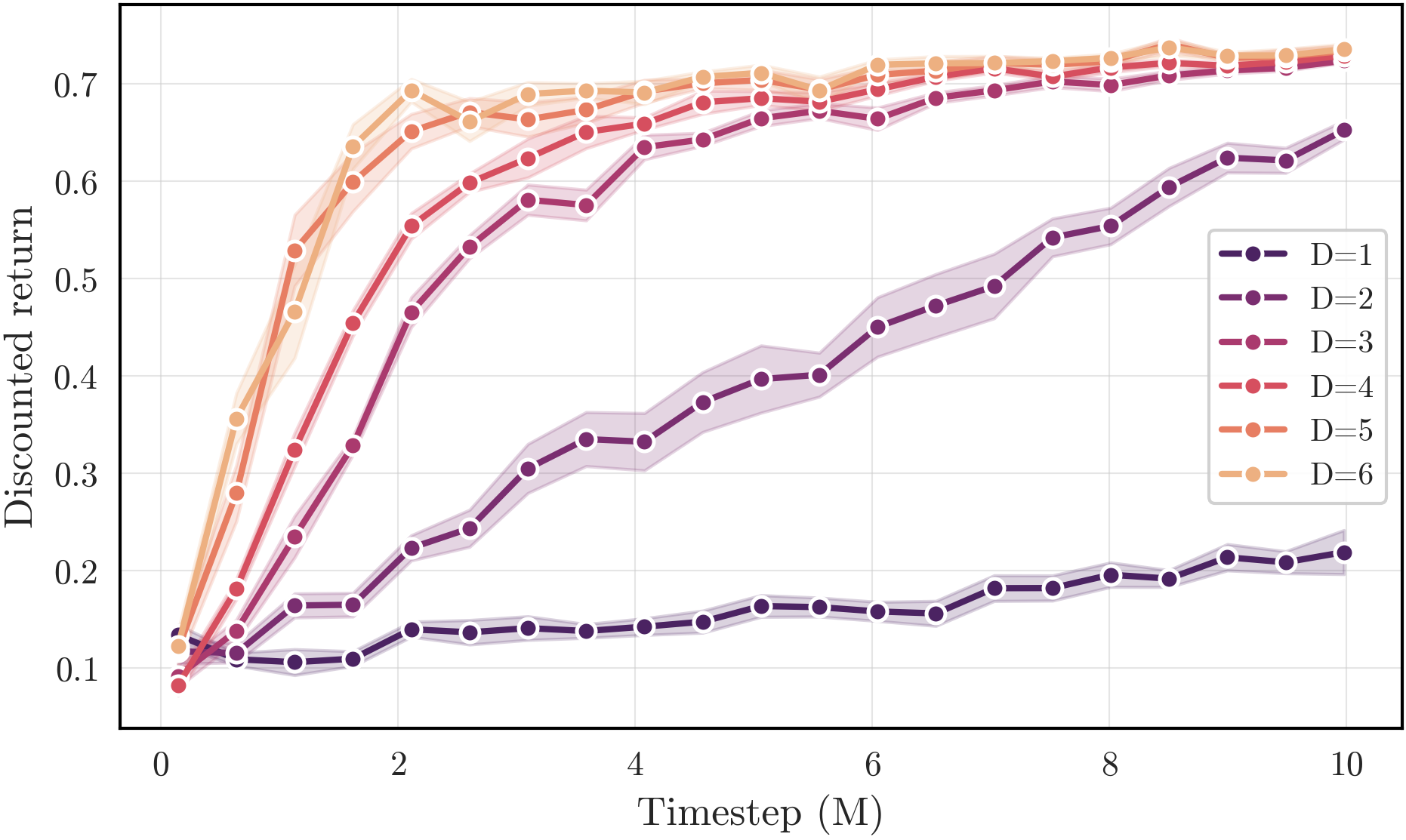}
        \caption{PQN return, slice at $W = 512$ (one curve per $D$).}
    \end{subfigure}
    \hfill
    \begin{subfigure}[t]{0.49\textwidth}
        \centering
        \includegraphics[width=\linewidth]{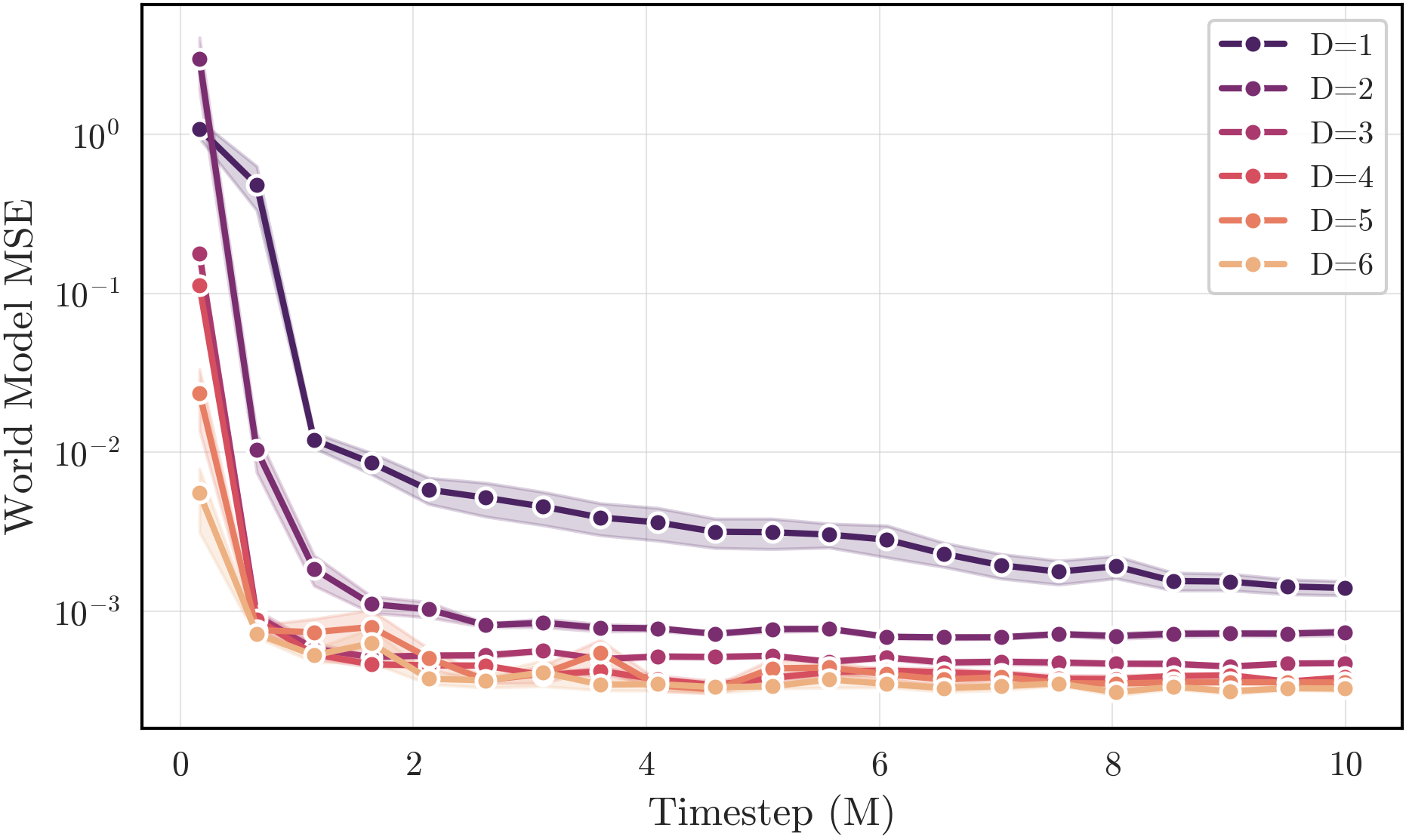}
        \caption{WM dynamics MSE (log-y), slice at $W = 512$.}
    \end{subfigure}
    \caption{PQN return (left) and WM dynamics MSE (right) vs training step for \texttt{Reacher}, with varying depth (top) / width (bottom) at fixed $D = 4$ and $W = 512$. Mean $\pm$ SE over $10$ seeds.}
    \label{fig:reacher-arch-slices}
\end{figure}

We expand the architecture-capacity analysis behind \cref{fig:reacher-correlation} by sweeping $42$ architectures over depths $D \in \{1, 2, 3, 4, 5, 6\}$ and widths $W \in \{32, 64, 128, 256, 512, 1024, 2048\}$, with $10$ seeds each, and record agent return and extracted WM error (MSE) over 20 checkpoints (a total of $420$ \texttt{Reacher} runs and 8400 checkpoints). All hyperparameters match \cref{tab:hparams-reacher} apart from the $Q$-network depth and width, which we vary. We record three quantities: (i) the PQN return averaged over the four training goals, (ii) the WM dynamics MSE on a held-out grid, and (iii) the WM-derived policy return on three unseen goals (\textit{far fingertip}, \textit{target angle}, \textit{target velocity}). \cref{fig:reacher-arch-slices} in the main text shows the results across all 42 architectures at the end of training. \cref{tab:reacher-arch-correlations} below reports the final-checkpoint correlations, confirming that more performant goal-conditioned agents tend to have more accurate implicit world models, and that these in turn yield higher zero-shot returns on unseen goals.\vspace{-1em}

\begin{table}[ht]
    \centering
    \small
    \caption{Final-checkpoint correlations across the $n = 42$ \texttt{Reacher} architectures, among PQN agent return, WM dynamics error, and WM-policy unseen-goal discounted return. $95\%$ confidence intervals in brackets ($2000$ resamples).}\label{tab:reacher-arch-correlations}\vspace{3pt}
    \setlength{\tabcolsep}{8pt}
    \begin{tabular}{@{}l cc@{}}
        \toprule
        \textbf{Pair} & Spearman $\rho$ & Pearson $r$ \\
        \midrule
        PQN return vs.\ WM error           & $-0.98$ $[-0.99, -0.95]$ & $-0.82$ $[-0.88, -0.76]$ \\
        PQN return vs.\ WM-policy return   & $+0.95$ $[+0.89, +0.97]$ & $+0.95$ $[+0.92, +0.97]$ \\
        WM error vs.\ WM-policy return     & $-0.96$ $[-0.98, -0.91]$ & $-0.82$ $[-0.88, -0.77]$ \\
        \bottomrule
    \end{tabular}
\end{table}

\begin{figure}[ht]
    \centering
    \includegraphics[width=0.75\textwidth]{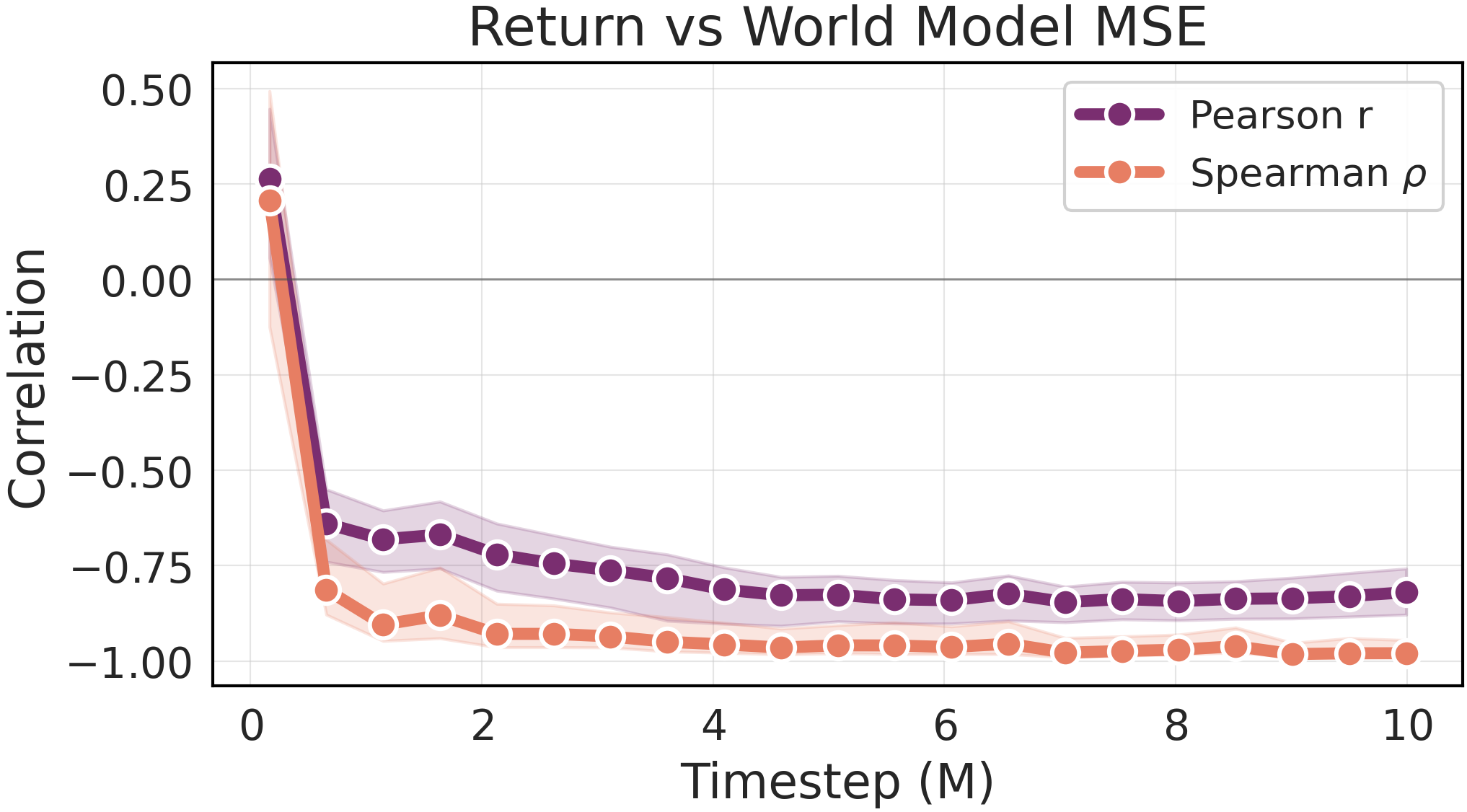}
    \caption{Pearson and Spearman correlations over time between \texttt{Reacher} agent return and WM MSE, across the $n = 42$ architectures, with $95\%$ confidence intervals shown in shaded regions.}\label{fig:reacher-correlations-over-time}
\end{figure}

To investigate whether this correlation only holds at the end of training, we plot the PQN return and WM MSE \textit{during} training in \cref{fig:reacher-arch-slices}, sliced at fixed depth $D = 4$ with varying $W$ (top) and at fixed width $W = 512$ with varying $D$ (bottom). The plots visually suggest a strong correlation between agent performance and WM accuracy, not only at the end of training but throughout. This is confirmed numerically by the Pearson and Spearman correlations over time reported in \cref{fig:reacher-correlations-over-time}.

\subsection{MountainCar}\label{sec:appendix-mountaincar}

\texttt{MountainCar} has a 2D position-velocity state space $\cS = [-1.2, 0.6] \times [-0.07, 0.07]$ and three acceleration actions $\cA = \{\text{left}, \text{none}, \text{right}\}$, with initial position uniform on $[-0.6, -0.4]$ and zero initial velocity. The state distribution $\mu_\cS$ used by $P$-learning is uniform over the full state space $\cS$. We consider two agents that share the hyperparameters listed in \cref{tab:hparams-mountaincar}: a position-trained agent with 4 linearly spaced goals $g_x \in \{-1.2, -0.6, 0, 0.6\}$ and indicator reward with $\sigma = 0.1$, matching the original environment-provided goal for $g_x = 0.6$, and a velocity-trained counterpart \cref{fig:mountaincar-platonic} with linearly spaced goals $g_v \in \{-0.07, -0.023, 0.023, 0.07\}$ and threshold $\sigma = 0.01$ to match the approximately 10x smaller velocity state space. All results are provided in the main text.

\begin{table}[ht]
    \centering
    \small
    \caption{Hyperparameters for \texttt{Reacher}.}\label{tab:hparams-reacher}\vspace{3pt}
    \begin{tabular}{@{}l l p{0.55\linewidth}@{}}
        \toprule
        \textbf{Group} & \textbf{Hyperparameter} & \textbf{Value} \\
        \midrule
        \multirow{4}{*}{Env}
            & State / action dims & $6 \,/\, 9$ \\
            & Number of goals $|\cG|$ & $4$ (cardinal $(x,y)$ targets on unit circle) \\
            & Reward / radius / termination & indicator $\mathbf{1}[\norm{s-g}<\sigma]$, $\sigma=0.2$, terminate on arrival \\
            & Discount $\gamma$ / Episode length & $0.99$ / $100$ steps \\
        \midrule
        \multirow{8}{*}{PQN}
            & Network & MLP, $6 \times 2048$, layer norm, sigmoid outputs \\
            & Backup & 1-step TD ($\lambda = 0$) \\
            & Optimiser & RAdam, global-norm gradient clipping at $10$ \\
            & Learning rate & $10^{-4}$, linear decay \\
            & Total env steps & $10^7$ ($256$ envs $\times\, 64$ rollout steps $\times\, 610$ updates) \\
            & Minibatch / epochs per update & $256$ / $1$ \\
            & $\varepsilon$-greedy schedule & $1.0 \to 0.1$ over first $50\%$ of training \\
            & HER / resets & $k=4$, ``future'' strategy; optimistic resets, ratio $16$ \\
        \midrule
        \multirow{5}{*}{WM}
            & Network & MLP, $2 \times 256$, $\tanh$, no layer norm \\
            & Output & residual $P_\phi(s,a) = s + h_\phi(s,a)$ (init scale $0.01$) \\
            & Optimiser & Adam, no gradient clipping, weight decay $0$ \\
            & Loss / LR & $\ell_1$; $10^{-4}$, linear decay \\
            & Batch / training steps & $1024 \,/\, 1000$ \\
        \bottomrule
    \end{tabular}
\end{table}

\begin{table}[ht]
    \centering
    \small
    \caption{Hyperparameters for \texttt{MountainCar}. The velocity-conditioned agent of \cref{fig:mountaincar-platonic} uses an identical configuration with goals $g_v \in \{-0.07, -0.023, 0.023, 0.07\}$ and $\sigma = 0.01$.}\label{tab:hparams-mountaincar}\vspace{3pt}
    \begin{tabular}{@{}l l p{0.55\linewidth}@{}}
        \toprule
        \textbf{Group} & \textbf{Hyperparameter} & \textbf{Value} \\
        \midrule
        \multirow{4}{*}{Env}
            & State / action dims & $2 \,/\, 3$ \\
            & Number of goals $|\cG|$ & $4$ ($x$-positions $\{-1.2, -0.6, 0, 0.6\}$) \\
            & Reward / radius / termination & indicator $\mathbf{1}[\norm{s-g}<\sigma]$, $\sigma=0.1$, terminate on arrival \\
            & Discount $\gamma$ / Episode length & $0.99$ / $200$ steps \\
        \midrule
        \multirow{8}{*}{PQN}
            & Network & MLP, $4 \times 1024$, layer norm, sigmoid outputs \\
            & Backup & 1-step TD ($\lambda = 0$) \\
            & Optimiser & RAdam, global-norm gradient clipping at $10$ \\
            & Learning rate & $10^{-4}$, linear decay \\
            & Total env steps & $5\times 10^6$ ($256$ envs $\times\, 64$ rollout steps $\times\, 305$ updates) \\
            & Minibatch / epochs per update & $256$ / $1$ \\
            & $\varepsilon$-greedy schedule & $1.0 \to 0.1$ over first $50\%$ of training \\
            & HER / resets & $k=4$, ``random'' strategy; optimistic resets, ratio $16$ \\
        \midrule
        \multirow{5}{*}{WM}
            & Network & MLP, $2 \times 256$, $\tanh$, no layer norm \\
            & Output & residual $P_\phi(s,a) = s + h_\phi(s,a)$ (init scale $0.01$) \\
            & Optimiser & Adam, no gradient clipping, weight decay $0$ \\
            & Loss / LR & $\ell_1$; $10^{-4}$, cosine decay \\
            & Batch / training steps & $4096 \,/\, 20{,}000$ \\
        \bottomrule
    \end{tabular}
\end{table}

\clearpage
\section{FourRooms Experiments (Stochastic Deterministic)}\label{sec:appendix-experiments-gridworld}\label{sec:exp-finite}

\begin{figure}[ht]
    \centering
    \includegraphics[width=\linewidth]{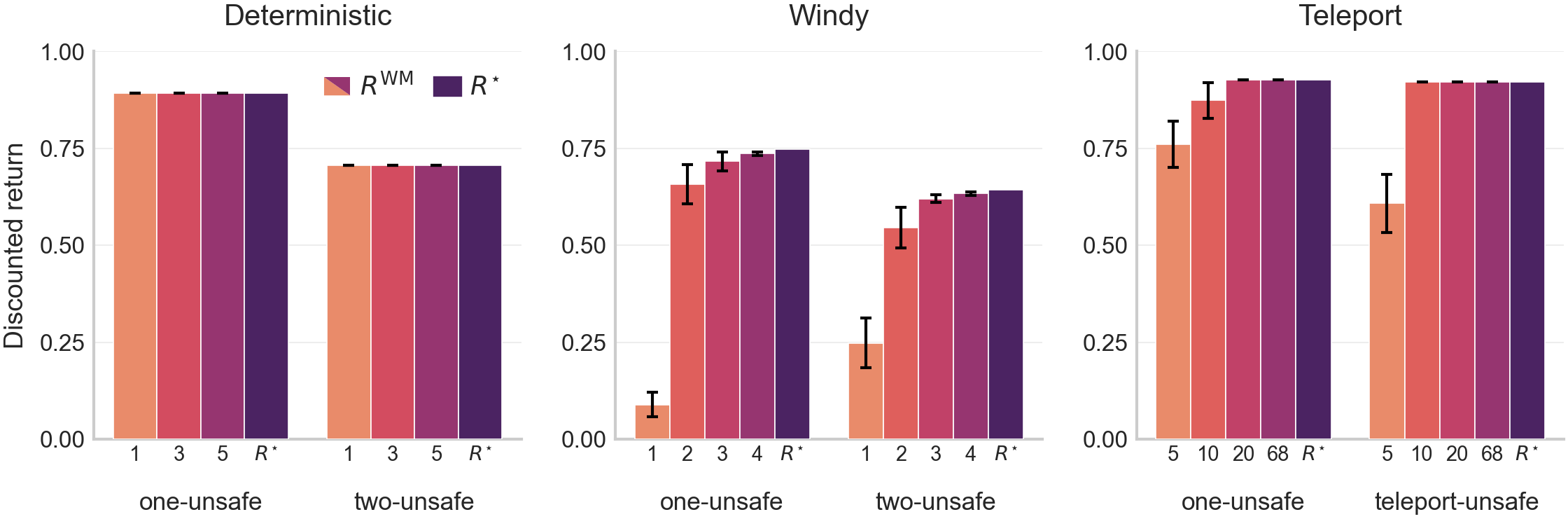}
    \caption{Mean return $\pm$ SE (10 training seeds), for optimal $(R^\star)$ vs WM-trained $(R^\text{WM})$ policies, on two unseen goals, in each \texttt{FourRooms} variant, for varying number of training goals $|\cG|$ on the $x$-axis. In the \textbf{deterministic} variant, recovery is exact at every $|\cG| \geq 1$, in line with \cref{thm:deterministic}. In the \textbf{windy} variant, $|\cG| = 4$ training goals already drive the WM-derived policy to within $\sim 1\%$ of optimal return. In the \textbf{teleporting} variant, $|\cG| = 20$ training goals -- well short of the $|\cG| = |\cS| = 68$ worst case of \cref{thm:finite} -- also induce quasi-optimal policies on unseen goals.}\label{fig:rooms-bar-plot}
\end{figure}

\begin{figure}[ht]
    \centering
    \begin{subfigure}{0.48\textwidth}
        \centering
        \includegraphics[width=\linewidth]{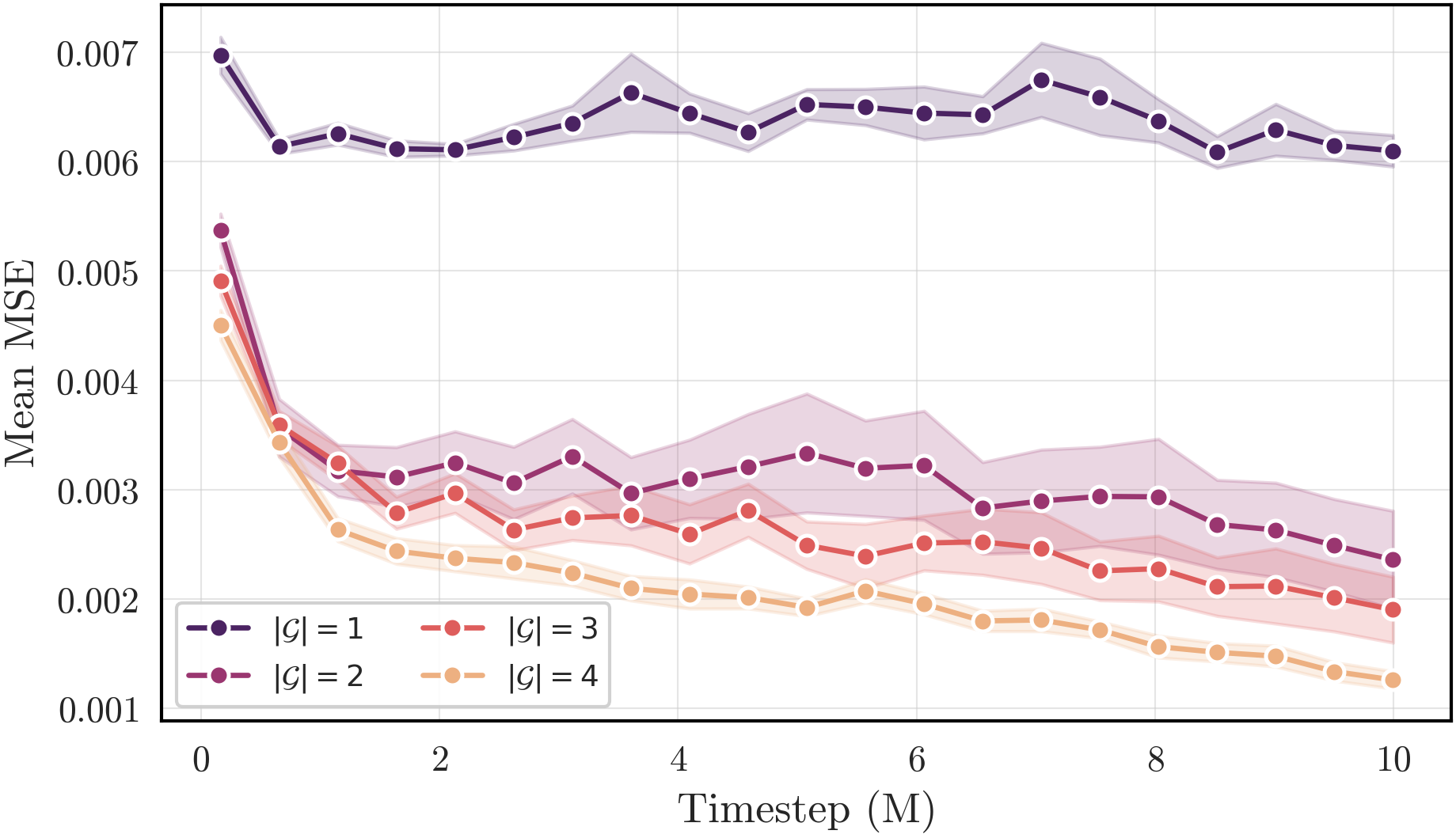}
        \caption{Transition MSE vs.\ training step.}
    \end{subfigure}\hfill
    \begin{subfigure}{0.48\textwidth}
        \centering
        \includegraphics[width=\linewidth]{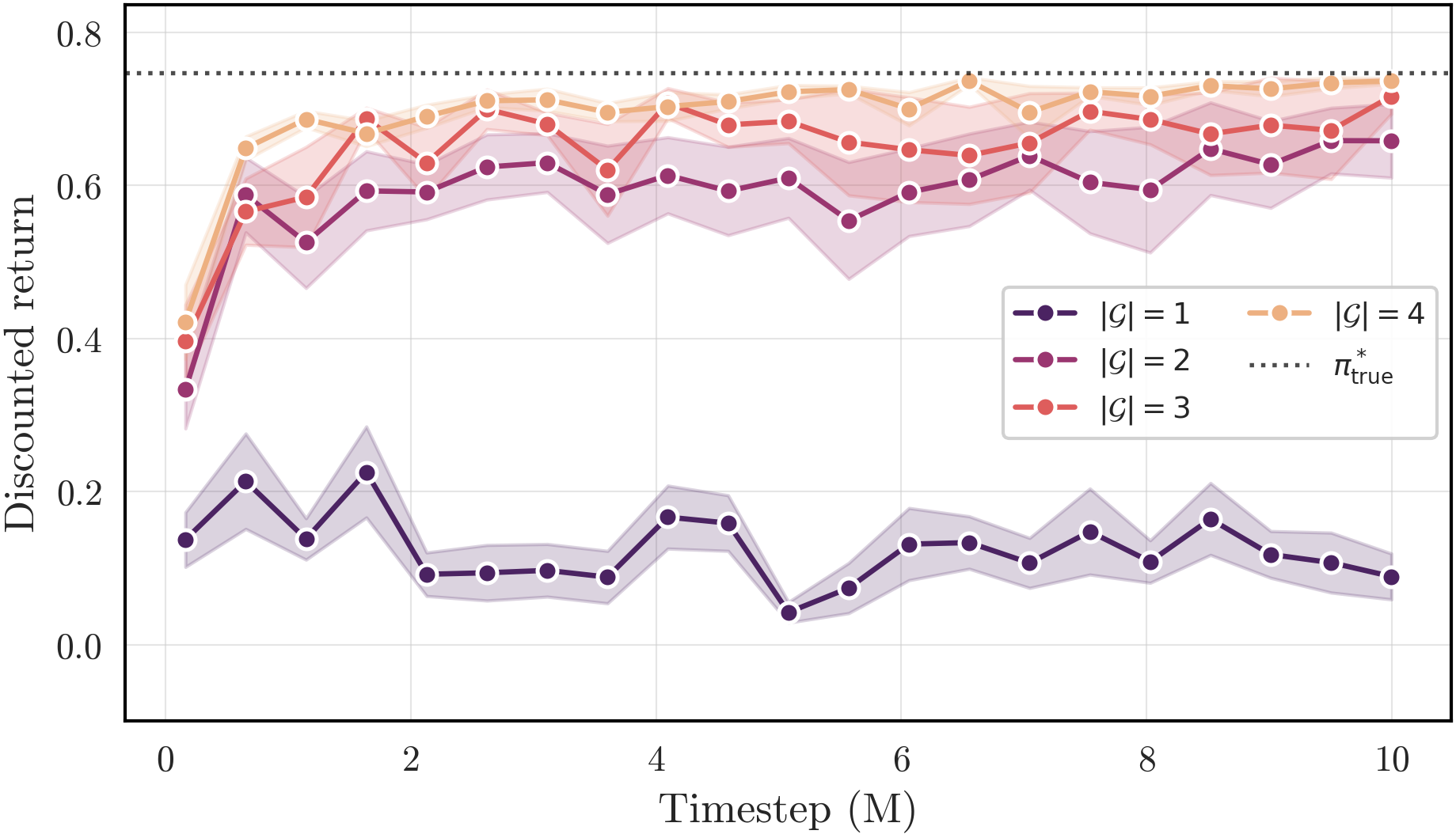}
        \caption{Discounted return on \textit{one-unsafe}.}
    \end{subfigure}\\[0.4em]
    \begin{subfigure}{0.48\textwidth}
        \centering
        \includegraphics[width=\linewidth]{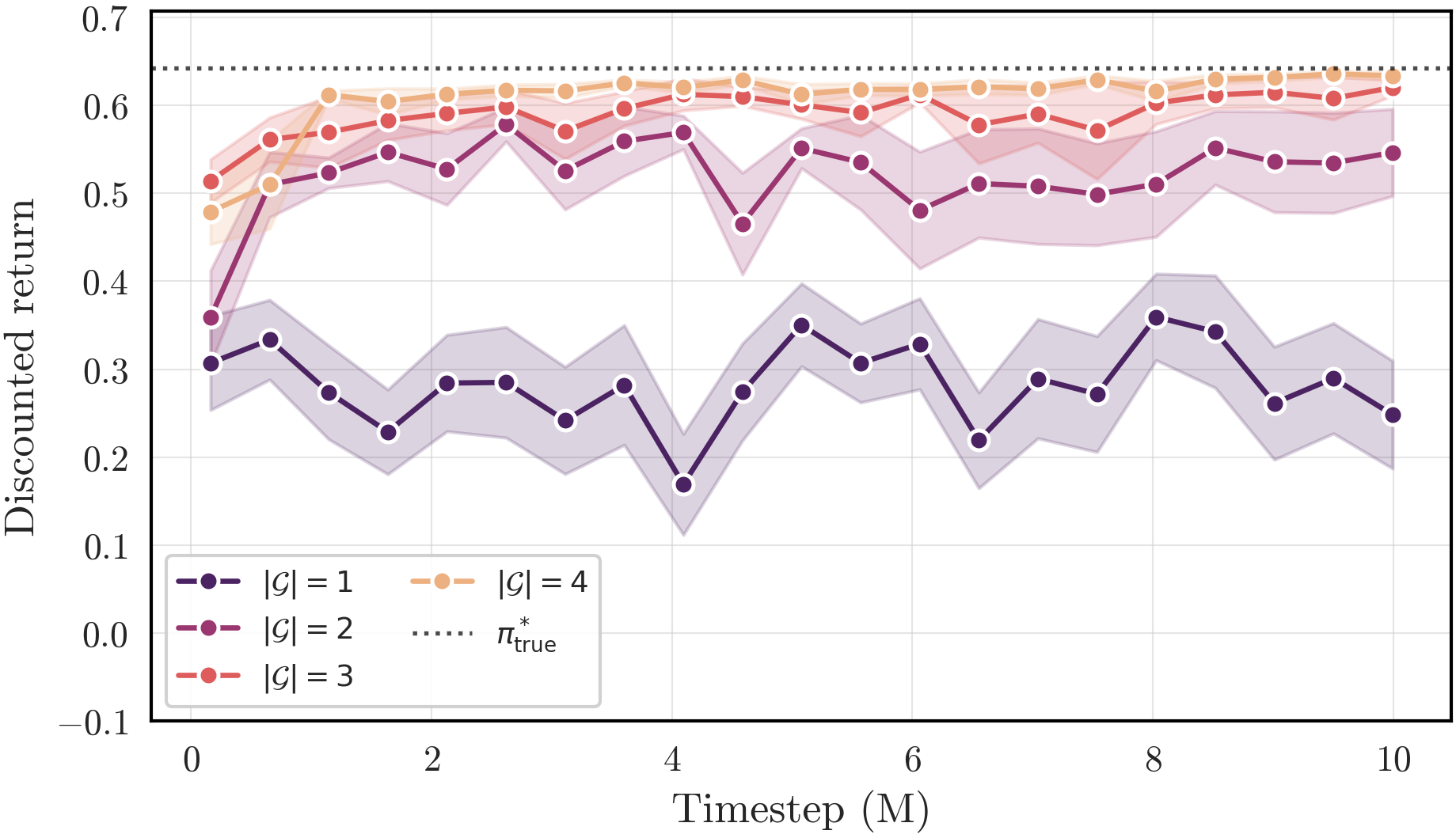}
        \caption{Discounted return on \textit{two-unsafe}.}
    \end{subfigure}\hfill
    \begin{subfigure}{0.49\textwidth}
        \centering
        \includegraphics[width=\linewidth]{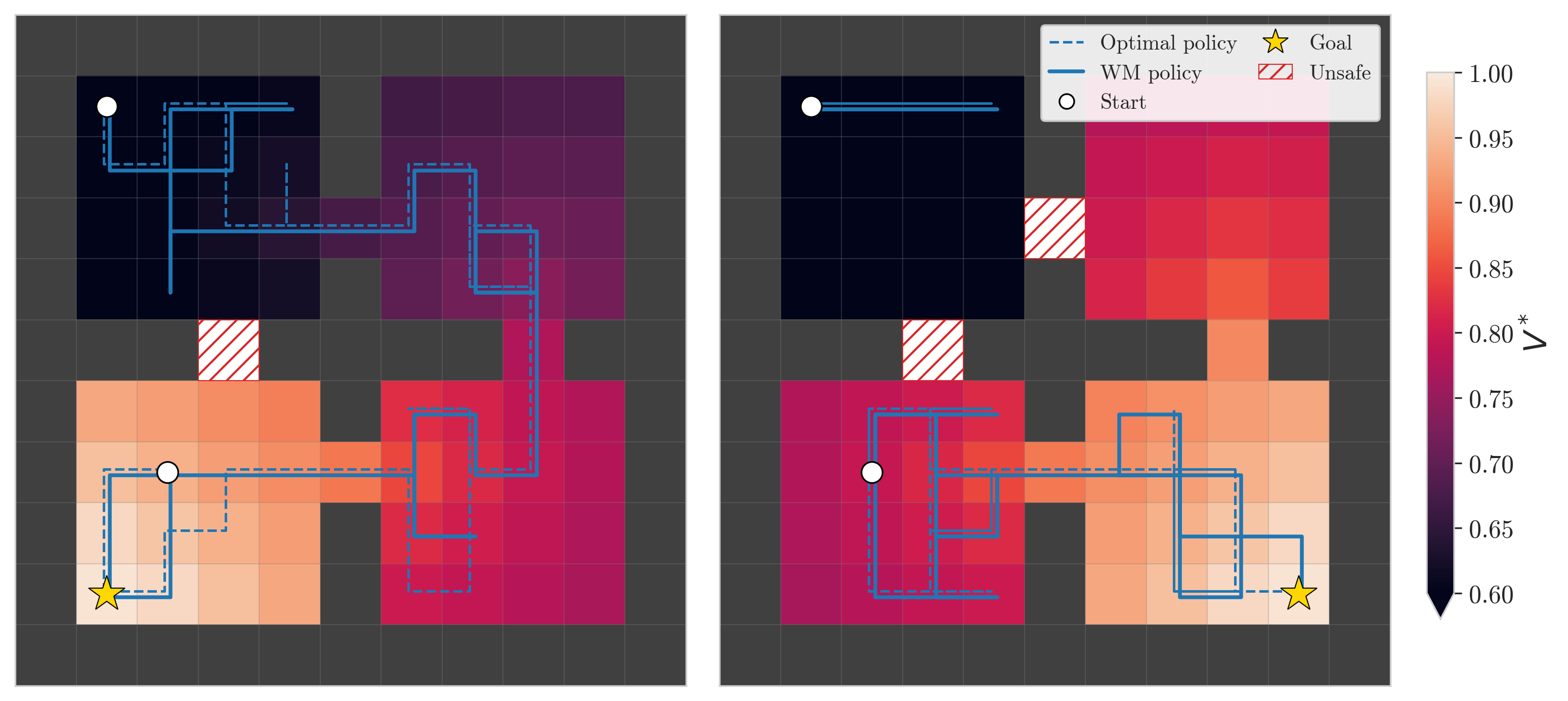}
        \caption{$V^\star$ heatmap with trajectories of optimal (dashed red) and WM-derived (solid blue) policies on both unseen goals, for an agent trained with $|\cG| = 4$ goals.}
    \end{subfigure}
    \caption{Windy \texttt{FourRooms}: the single-goal regime is no longer sufficient for stochastic dynamics, but $|\mathcal{G}| = 4$ drives the WM-derived policy to within a few percent of the oracle on both unseen goals.}\label{fig:rooms-windy}
\end{figure}

\begin{figure}[t]
    \centering
    \begin{subfigure}{0.48\textwidth}
        \centering
        \includegraphics[width=\linewidth]{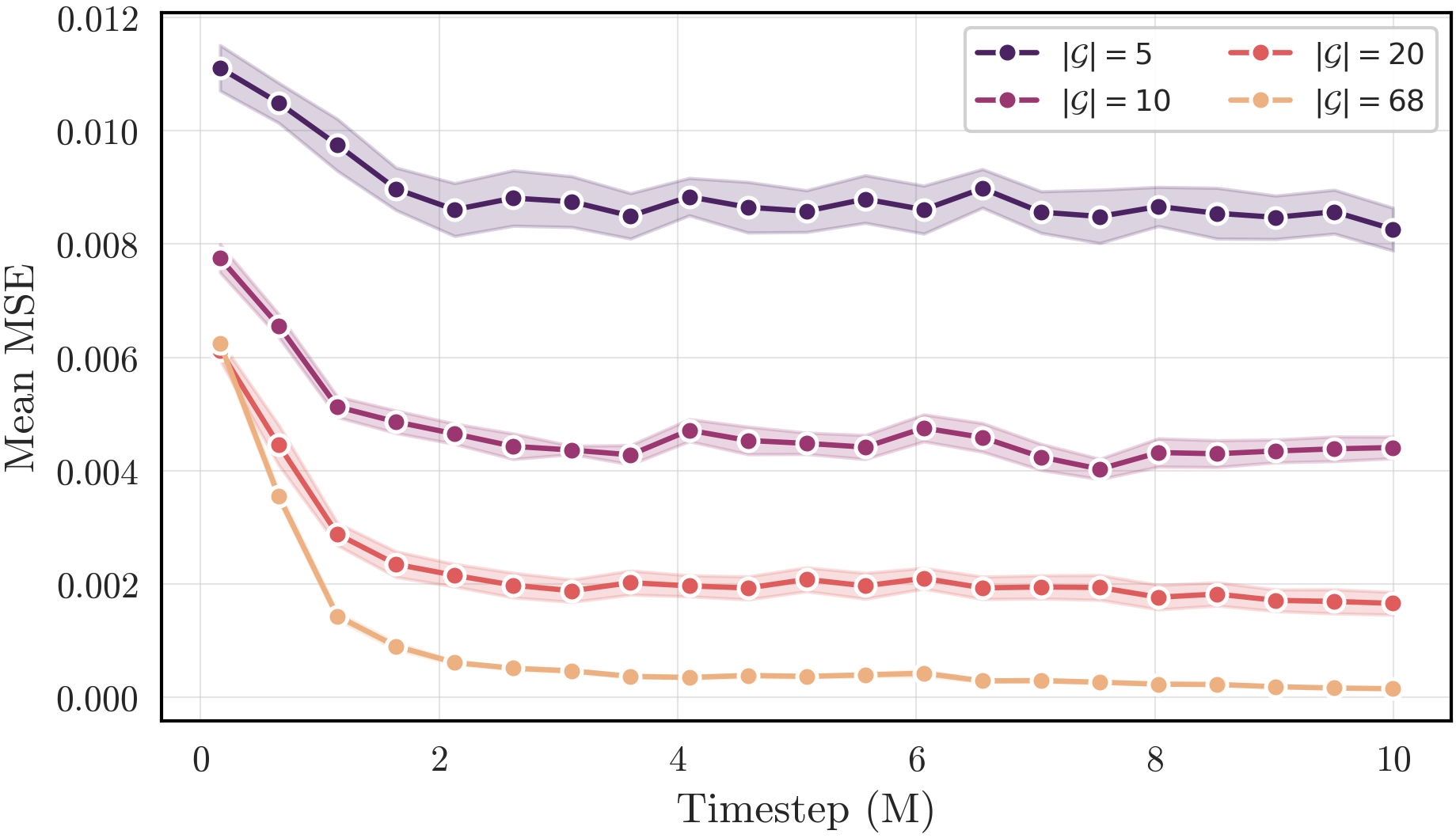}
        \caption{Transition MSE vs.\ training step.}
    \end{subfigure}\hfill
    \begin{subfigure}{0.48\textwidth}
        \centering
        \includegraphics[width=\linewidth]{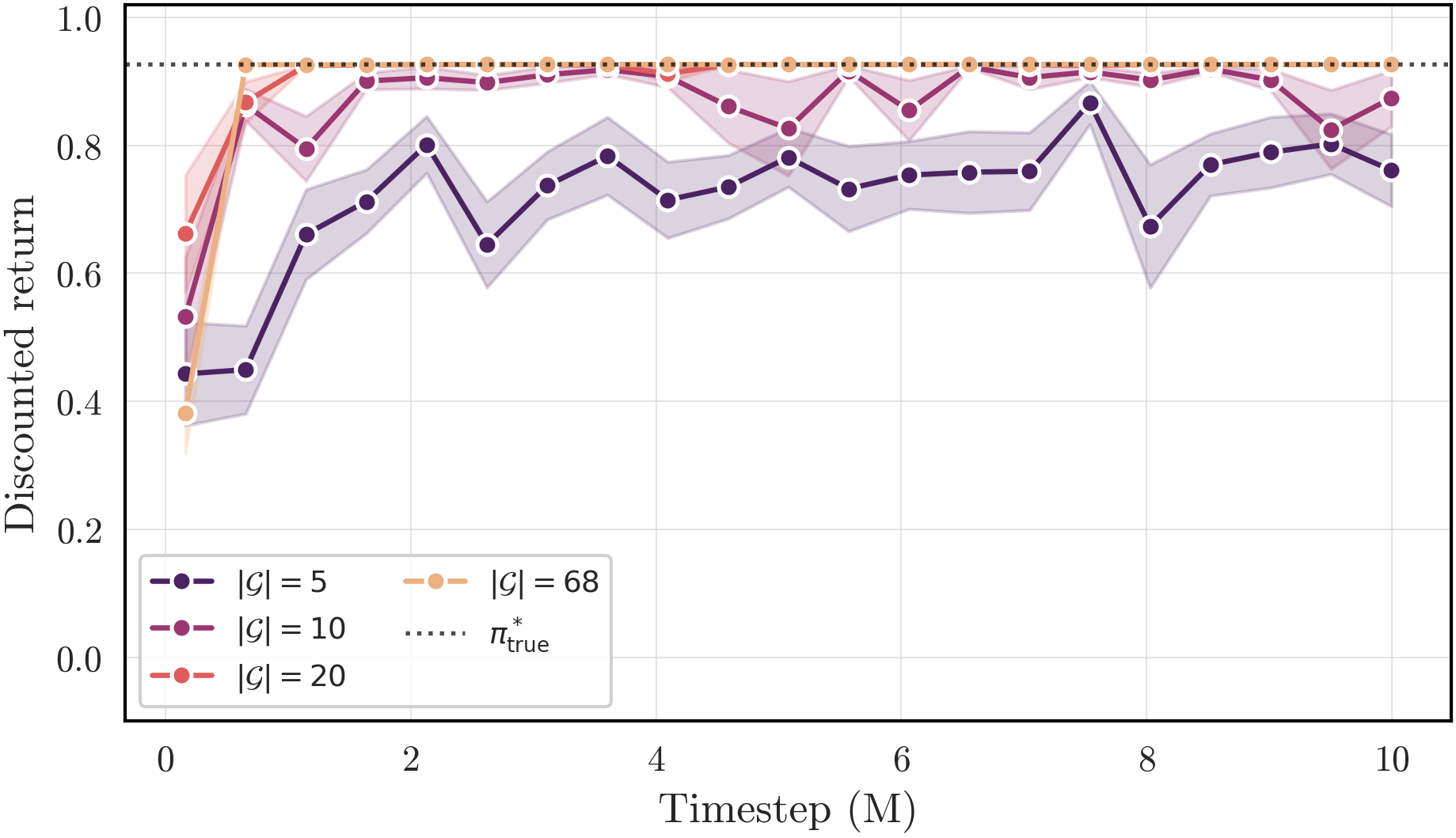}
        \caption{Discounted return on \textit{one-unsafe}.}
    \end{subfigure}\\[0.4em]
    \begin{subfigure}{0.48\textwidth}
        \centering
        \includegraphics[width=\linewidth]{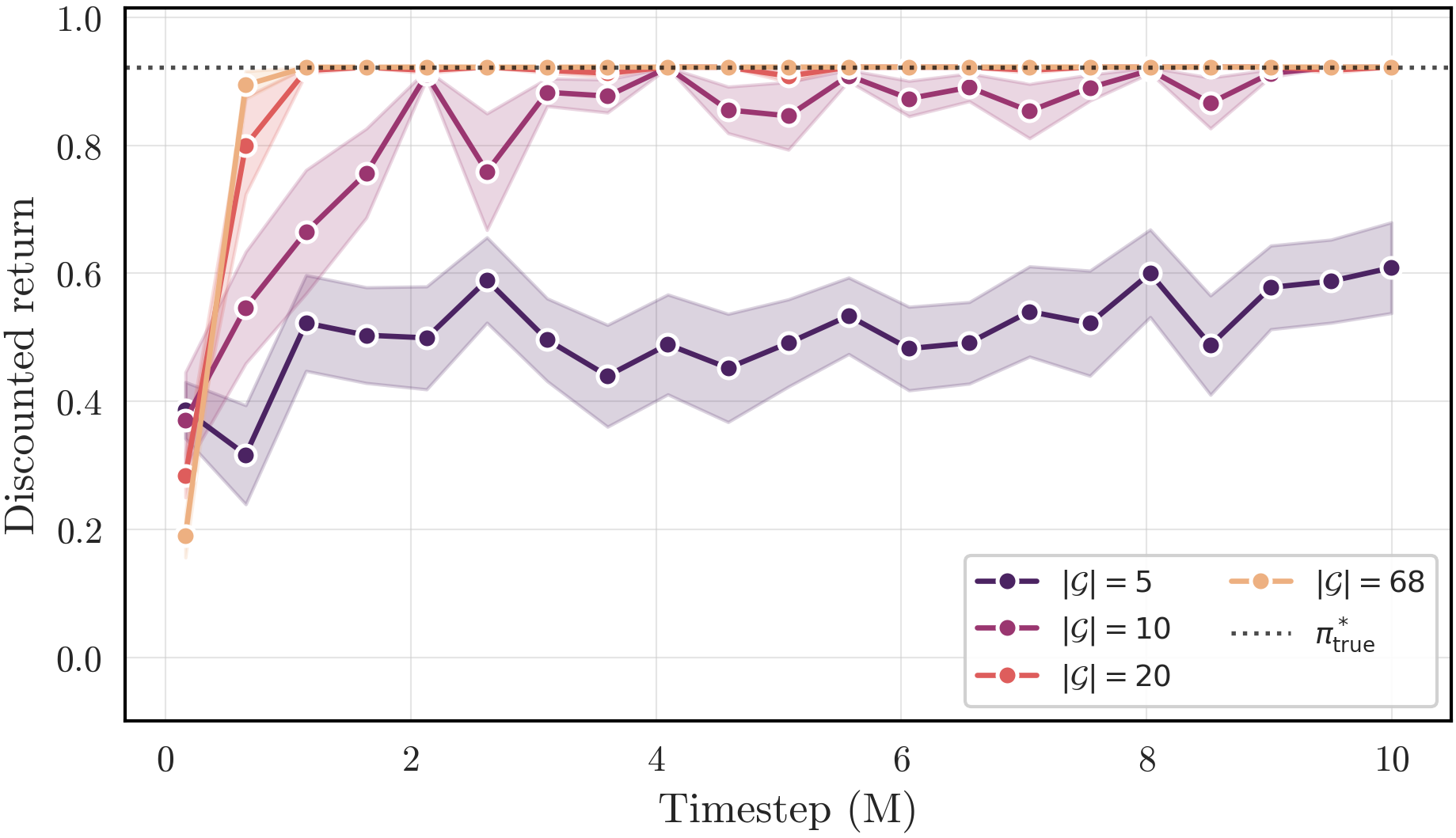}
        \caption{Discounted return on \textit{two-unsafe}.}
    \end{subfigure}\hfill
    \begin{subfigure}{0.5\textwidth}
        \centering
        \includegraphics[width=\linewidth]{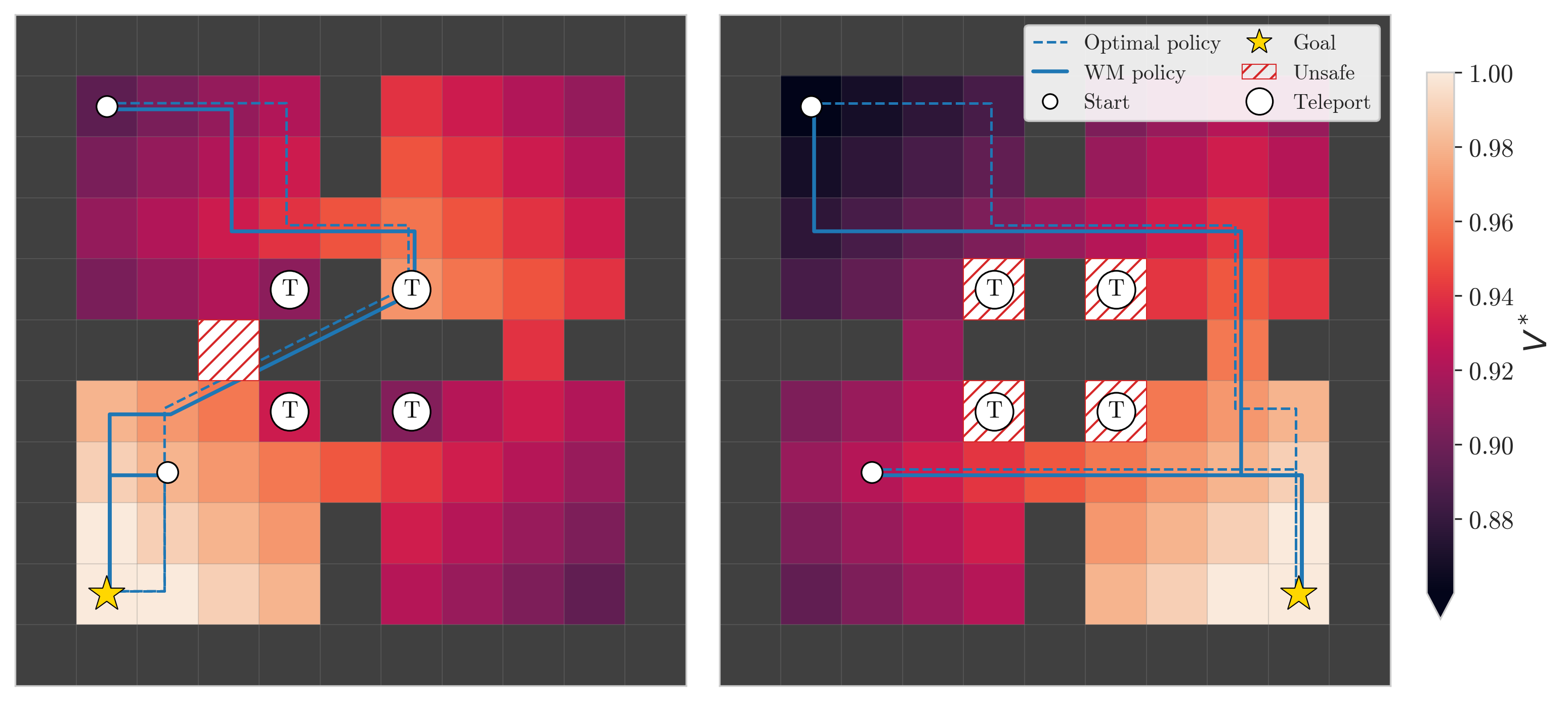}
        \caption{$V^\star$ heatmap with trajectories of optimal (dashed) and WM-derived (solid) policies on both unseen goals, for an agent trained with $|\cG| = 20$ goals.}
    \end{subfigure}
    \caption{Teleporting \texttt{FourRooms}: for $|\mathcal{G}| = 20$, the extracted WM produces quasi-optimal policies on OOD goals, well below our worst-case theoretical bound $|\mathcal{G}| \geq |\mathcal{S}| = 68$.}\label{fig:rooms-teleport}
\end{figure}

We complement the continuous-control experiments with a finite-state-space study in \texttt{FourRooms} \citep{Sutton1999}, a $11 \times 11$ gridworld divided into $4 \times 4$ rooms by interior walls and connected by single-cell corridors. After removing wall cells, the open state space has $|\cS| = 68$ cells, the action space is $\cA = \{\text{up}, \text{right}, \text{down}, \text{left}\}$ with $|\cA| = 4$, and the episode horizon is $T_\text{max} = 200$. To cover the regimes of our finite-MDP theory (deterministic in \cref{thm:deterministic}; local in \cref{prop:local}; stochastic in \cref{thm:finite}), we instantiate three variants of increasing stochasticity. Across all three, the WM-derived policy reaches near-optimal return on unseen goals from a \emph{small fraction} of the $|\cS| = 68$ candidate training goals: 1 goal for deterministic, 4 for windy, and roughly 20 for teleporting -- consistent with our identifiability theory in the first two cases, and exceeding our expectations in the third. To summarise the environments and results:

\begin{enumerate}
    \item \textbf{Deterministic} (\cref{fig:rooms-bar-plot,fig:rooms-det}). Each action moves the agent one cell in the chosen cardinal direction; walls keep the agent in place. \emph{A single training goal} ($|\cG| = 1$) already drives $\hat P$ to zero world-model MSE within 8M env steps and the WM-derived policy is exactly optimal on both unseen goals, in line with \cref{thm:deterministic}.
    \item \textbf{Windy} (\cref{fig:rooms-bar-plot,fig:rooms-windy}). The chosen action is realised w.p. $\tfrac{1}{2}$, perturbed $90^\circ$ counter-clockwise or clockwise each w.p.\ $\tfrac{1}{4}$. Here, $|\cG| = 4$ training goals already drive the WM-derived policy to within $\sim 1\%$ of optimal return on both unseen goals.
    \item \textbf{Teleporting} (\cref{fig:rooms-bar-plot,fig:rooms-teleport}). Four cells $(4,4), (4,6), (6,4), (6,6)$ teleport the agent \textit{uniformly} into the $16$ cells of the diagonally opposite $4 \times 4$ room. Here, $|\cG| = 20$ training goals -- well short of the $|\cG| = |\cS| = 68$ worst case of \cref{thm:finite} -- already drive the WM-derived policy to within $\sim 0.1\%$ of optimal return on both unseen goals.
\end{enumerate}

For each variant, the training goals are the first $|\cG|$ cells of a fixed seeded permutation of the $68$ states, so that the goal subsets nest as $|\cG|$ grows; all experiments use 10 seeds.

The reward function differs across variants in line with the regime each one targets: the windy and teleport variants use the indicator reward $r_g(s) = \delta_{gs}$, while the deterministic variant uses the noise-perturbed form $r_g(s) = \delta_{gs} - \xi_{gs}$ with each $\xi_{gs} \sim U[0, 1]$ drawn i.i.d. at initialisation, instantiating the noise-injection recipe stated as a corollary of \cref{thm:deterministic}. We take negative perturbation in order to guarantee that the optimal policy still seeks the goal, noting that $r_g(g) \geq 0 \geq r_g(s)$ if $s \neq g$.

We extract $\hat P$ from the trained PQN agent by $P$-learning via linear programming (LP), as defined in \cref{sec:appendix-lp-solver}, which we found to be fastest and most effective in this finite, one-hot-encoded setting, compared with function approximation. For the deterministic variant we use column matching, which is exact under the genericity condition of \cref{thm:deterministic}. As for \texttt{MountainCar} and \texttt{Reacher}, we evaluate $\hat P$ directly via the MSE, and on two unseen goals via value-iteration on $\hat P$. Final-step returns on unseen goals are summarised in \cref{fig:rooms-bar-plot}, and \cref{tab:hparams-fourrooms} lists the shared hyperparameters.

Our results for each variant are shown via a four-panel figure (\cref{fig:rooms-det,fig:rooms-windy,fig:rooms-teleport}): the transition MSE over training steps (one curve per $|\cG|$, with mean $\pm$ SE), the discounted return on each unseen goal over training steps, and a $V^\star$ heatmap overlaid with the optimal (dashed red) and WM-derived (solid blue) trajectories at the final checkpoint with $|\cG| = |\cG|_\text{max}$.

\subsection{\texorpdfstring{$P$-learning via linear programming}{P-learning via linear programming}}\label{sec:appendix-lp-solver}

Recall from \cref{eq:bellman-linear} that for finite MDPs, the $P$-learning fixed-point equation $\cT^\pi_P(Q) = Q$ decouples per $(s,a)$ into the linear system $M P(s,a) = Q(s,a)$, with $M \in \R^{|\cG|\times|\cS|}$, $M_{lk} = r(s'_k, g_l) + \gamma V(s'_k, g_l)$, and unknown row $P(s,a) \in \Delta^{|\cS|}$. For higher efficiency, we directly solve the per-$(s,a)$ system with constrained linear programming (LP) via SciPy, instead of using tabular $P$-learning. The full sweep over $|\cS||\cA| = 272$ pairs is computed on a single CPU in $< 1$ second.

\textbf{Local variant.} The windy variant is an $N$-local MDP as defined in \ref{sec:results}; we assume this local prior is known in order to illustrate our result for local MDPs, \cref{prop:local}, namely, that $N$ generic goals are sufficient to recover the world model. We therefore restrict the LP support to the cardinal-$4$ neighbours of $s$ plus $s$ itself, reducing each per-$(s,a)$ LP from $|\cS|=68$ to $5$ unknowns. This is a geometric prior derived from the gridworld topology, which injects no information about the probability masses in the true kernel.

\textbf{Deterministic variant.} For deterministic \texttt{FourRooms}, we use \emph{column matching}: $\hat P(s,a) = e_{k^\star}$ with $k^\star = \arg\min_k \norm{M_{\cdot,k} - q}_1$. This is the LP above with the simplex constraint replaced by a pure argmin; it is exact when $M$ is column-injective (\cref{thm:deterministic}) and requires no LP solver.

In all three cases, the LP solver is a numerical-stability and speed tradeoff over gradient-based $P$-learning, not a methodological one: both target the same Bellman residual $\norm{\cT^\pi_P(Q)-Q}^2$ and inherit the same identifiability guarantees of \cref{sec:proofs:finite}.

\subsection{Deterministic FourRooms}\label{sec:rooms-det}

\begin{figure}[ht]
    \centering
    \begin{subfigure}{0.48\textwidth}
        \centering
        \includegraphics[width=\linewidth]{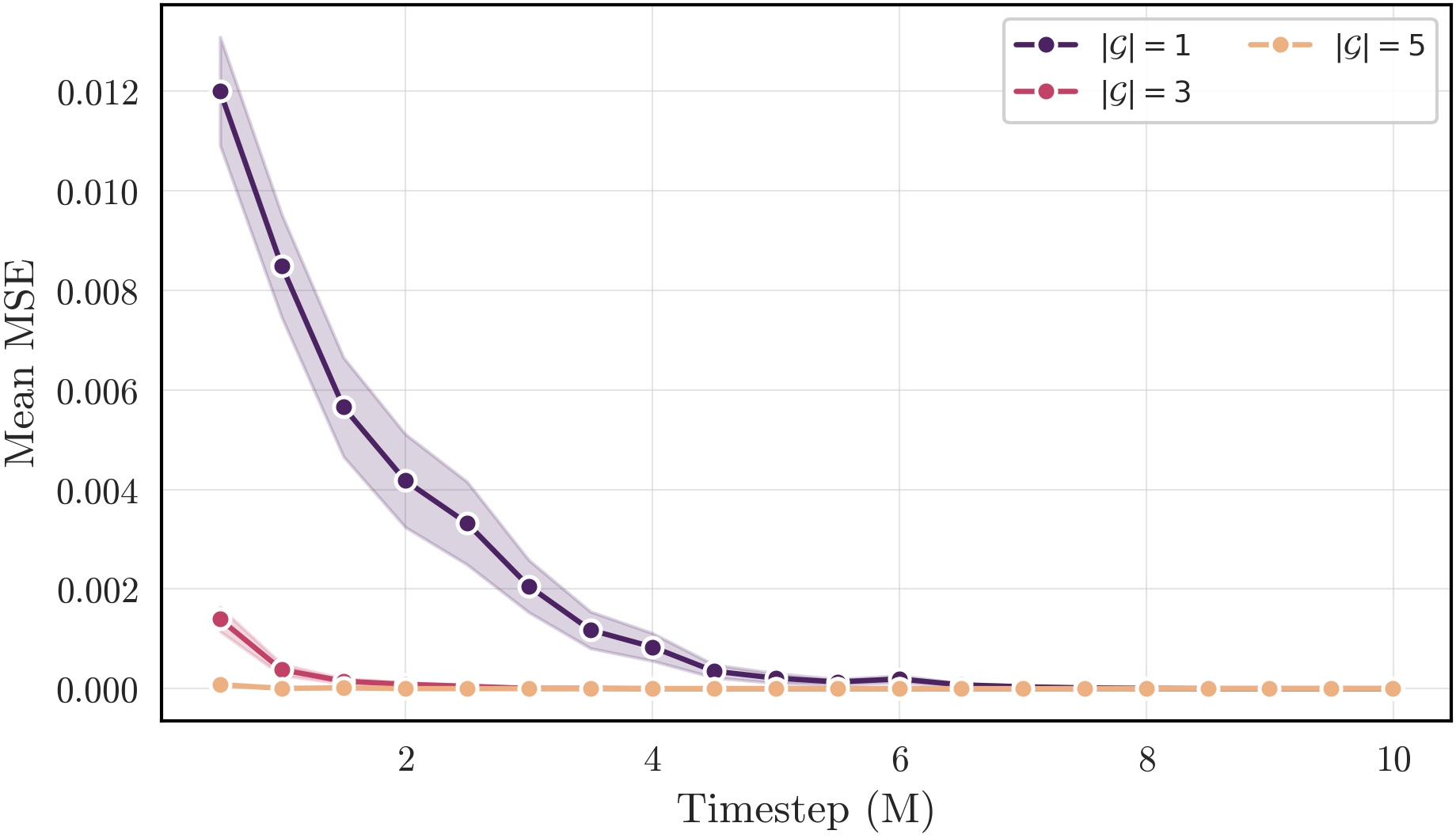}
        \caption{Transition MSE vs.\ training step.}
    \end{subfigure}\hfill
    \begin{subfigure}{0.48\textwidth}
        \centering
        \includegraphics[width=\linewidth]{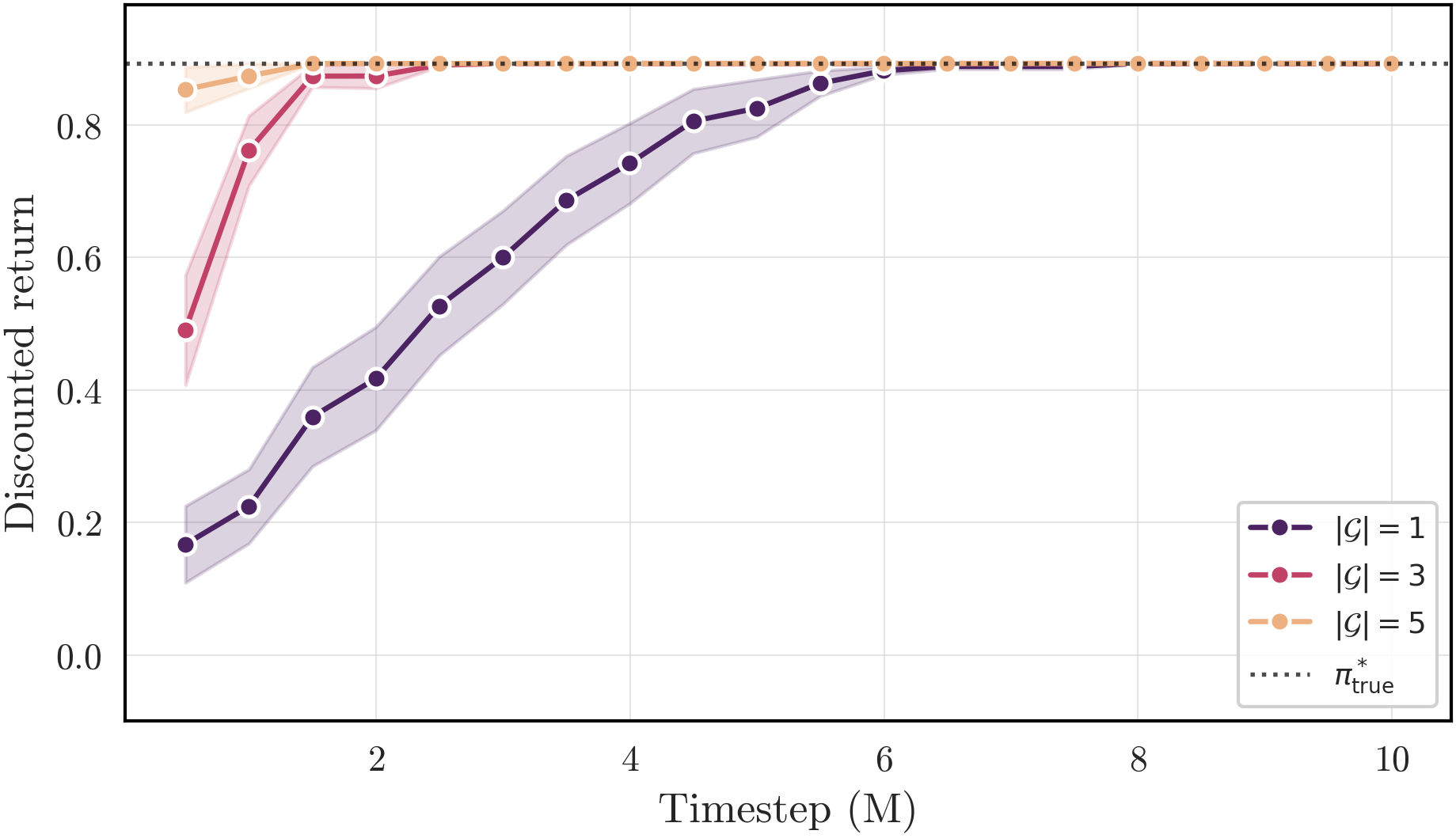}
        \caption{Discounted return on \textit{one-unsafe}.}
    \end{subfigure}\\[0.4em]
    \begin{subfigure}{0.48\textwidth}
        \centering
        \includegraphics[width=\linewidth]{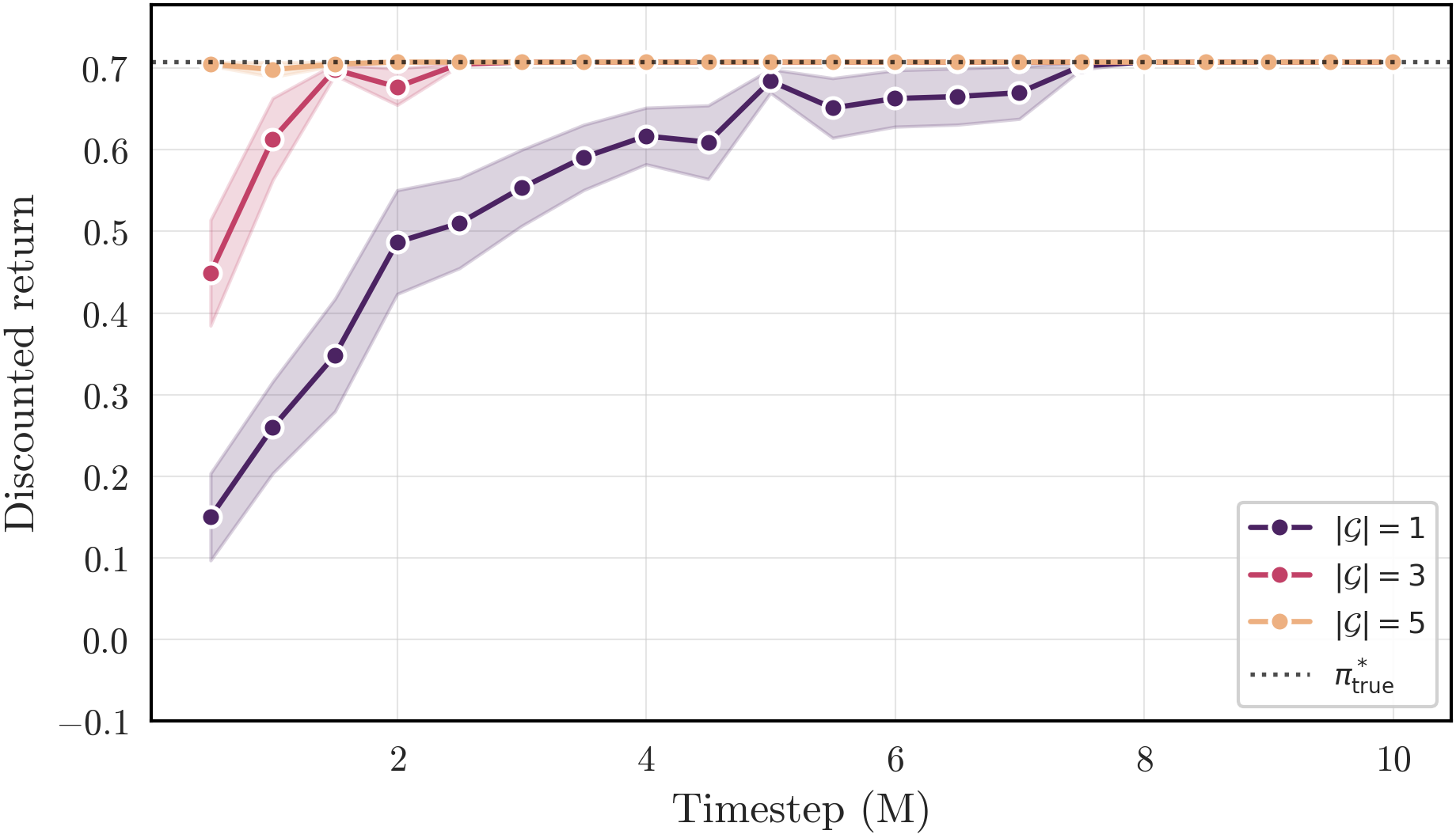}
        \caption{Discounted return on \textit{two-unsafe}.}
    \end{subfigure}\hfill
    \begin{subfigure}{0.5\textwidth}
        \centering
        \includegraphics[width=\linewidth]{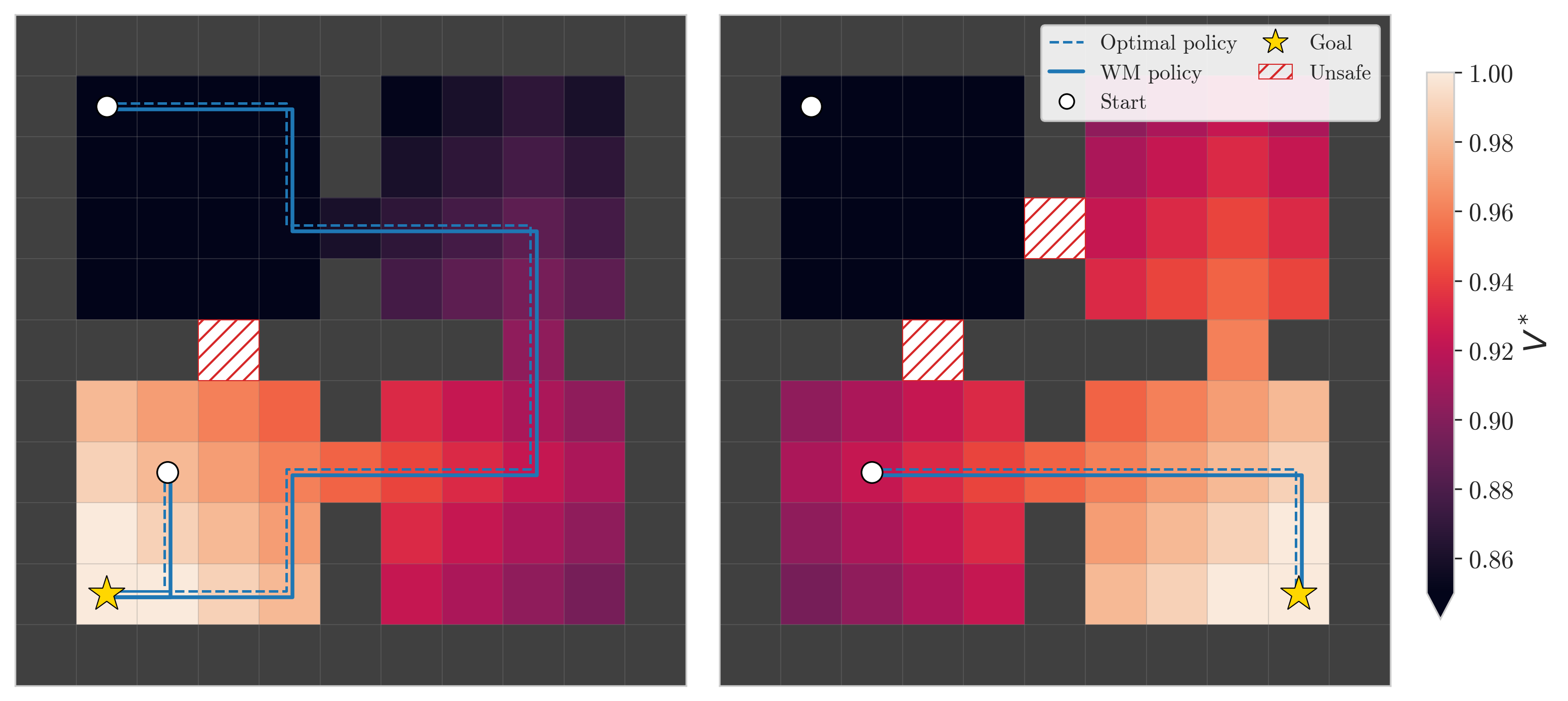}
        \caption{$V^\star$ heatmap with trajectories of optimal (dashed red) and WM-derived (solid blue) policies on both unseen goals, for an agent trained with $|\cG| = 1$ goal.}
    \end{subfigure}
    \caption{Deterministic \texttt{FourRooms}: a single generic goal drives the extracted WM to zero error during training, in line with \cref{thm:deterministic}, and training a policy inside of it produces a policy (solid) that is optimal (dashed) on both unseen goals.}\label{fig:rooms-det}
\end{figure}

\clearpage

In the deterministic variant, we use the noise-perturbed reward $r_g(s) = \delta_{gs} - \xi_{gs}$ with each $\xi_{gs} \sim U[0,1]$ drawn i.i.d. at initialisation, which guarantees column-injectivity of $M$ at any $|\cG| \geq 1$ via \cref{thm:deterministic}. We train PQN on $|\cG| \in \{1, 3, 5\}$ goals over $10$ seeds, recover $\hat P$ via column matching, and evaluate by training policies in the WM for two unseen goals: \textit{one-unsafe} (target $(9,1)$ with `unsafe' state $(5,3)$ that terminates the episode with zero reward) and \textit{two-unsafe} (target $(9,9)$ with two `unsafe' states $(3,5)$ and $(5,3)$). Consistent with \cref{thm:deterministic}, a single training goal already drives $\hat P$ to \textit{zero} world model MSE within 8M env steps (\cref{fig:rooms-det}), and the WM-derived policy is optimal on both unseen goals for any number of goals $|\cG| \geq 1$ (\cref{fig:rooms-bar-plot}).

\subsection{Windy FourRooms}\label{sec:rooms-windy}

The windy variant introduces local stochasticity: the chosen action is realised faithfully with probability $\tfrac{1}{2}$, and rotated $90^\circ$ counter-clockwise or clockwise each with probability $\tfrac{1}{4}$. This places the variant in the local-stochastic regime of \cref{prop:local}. We train PQN on $|\cG| \in \{1, 2, 3, 4\}$ training goals over $10$ seeds, recover $\hat P$ via the local LP, and evaluate on the same two unseen goals as the deterministic variant. As expected for stochastic dynamics, the single-goal regime is under-determined, but $|\cG| \geq 3$ already drives the WM-derived policy to within a few percent of the oracle on both goals (\cref{fig:rooms-bar-plot}).

\subsection{Teleporting FourRooms}\label{sec:rooms-teleport}

The teleporting variant injects fully-stochastic transitions at four \emph{teleport} cells $(4,4), (4,6), (6,4), (6,6)$, each of which re-emits the agent uniformly into the $16$ cells of the diagonally opposite $4 \times 4$ room; all other transitions are deterministic 4-connected moves. We train PQN on $|\cG| \in \{5, 10, 20, 68\}$ goals over $10$ seeds, and evaluate on \textit{one-unsafe} (as before) and \textit{teleport-unsafe} (all four teleport cells ``unsafe'' giving zero reward and terminating the environment). Surprisingly, $|\cG| \geq 20$ already induces a highly accurate WM, as further validated by a WM-derived policy which matches the optimal policy on both unseen goals to within $0.1\%$ -- well below the $|\cG| = 68$ goals that \cref{thm:finite} provides guarantees for.

\begin{table}[t]
    \centering
    \small
    \caption{Hyperparameters shared across the three \texttt{FourRooms} variants.}\label{tab:hparams-fourrooms}\vspace{3pt}
    \begin{tabular}{@{}l l p{0.55\linewidth}@{}}
        \toprule
        \textbf{Group} & \textbf{Hyperparameter} & \textbf{Value} \\
        \midrule
        \multirow{5}{*}{Env}
            & State / action dims & $68 \,/\, 4$ \\
            & Number of training goals $|\cG|$ & $\{1,3,5\}$ det., $\{1,2,3,4\}$ windy, $\{5,10,20,68\}$ teleport \\
            & Reward & terminate on arrival; indicator $r_g(s) = \delta_{gs}$ for windy/teleport; else perturbed $r_g(s) = \delta_{gs} - \xi$ with $\xi \sim U[0,1]$ \\
            & Discount $\gamma$ / Episode length & $0.99$ / $200$ steps \\
        \midrule
        \multirow{8}{*}{PQN}
            & Network & MLP, $4 \times 1024$, layer norm, sigmoid outputs \\
            & Backup & 1-step TD ($\lambda = 0$) \\
            & Optimiser & RAdam, global-norm gradient clipping at $10$ \\
            & Learning rate & $10^{-4}$, linear decay \\
            & Total env steps & $10^7$ ($256$ envs $\times\, 64$ rollout steps $\times\, 610$ updates) \\
            & Minibatch / epochs per update & $256$ / $1$ \\
            & $\varepsilon$-greedy schedule & $1.0 \to 0.1$ over first $50\%$ of training \\
            & HER / resets & $k=4$, ``future'' strategy; optimistic resets, ratio $16$ \\
        \midrule
        \multirow{4}{*}{WM}
            & Method (det.\ / windy / tele.) & column matching / local-prior LP / global LP \\
            & LP norm & $\ell_1$ (per-$(s,a)$, simplex-constrained) \\
            & Local prior (windy) & cardinal-$4$ neighbours $\cup\,\{s\}$ \\
            & Sampling & full sweep of the discrete grid $\cS \times \cA$ ($272$ pairs) \\
        \bottomrule
    \end{tabular}
\end{table}

\end{document}